\def\cleanversion{}

\documentclass[iicol,pdflatex,sn-mathphys-ay]{sn-jnl}%

\usepackage{graphicx}%
\usepackage{multirow}%
\usepackage{amsmath,amssymb,amsfonts}%
\usepackage{amsthm}%
\usepackage{mathrsfs}%
\usepackage[title]{appendix}%
\usepackage{xcolor}%
\usepackage{textcomp}%
\usepackage{manyfoot}%
\usepackage{booktabs}%
\usepackage{algorithm}%
\usepackage{algorithmicx}%
\usepackage{algpseudocode}%
\usepackage{listings}%
\usepackage{hyperref}
\usepackage{url}
\usepackage{bmpsize}
\usepackage{multirow}
\usepackage{subcaption}
\usepackage{amsmath}
\usepackage{amssymb}
\usepackage{booktabs}
\usepackage{svg}

\usepackage{nicematrix}
\usepackage{xcolor}

\newcommand{\parsection}[1]{\textbf{#1}}
\newcommand{\target}{\mathrm{target}}
\newcommand{\planner}{\mathrm{planner}}
\newcommand{\imitation}{\mathrm{imitation}}
\newcommand{\collision}{\mathrm{collision}}
\newcommand{\riskmonitor}{\mathrm{RiskMonitor}}
\newcommand{\loss}{\mathrm{loss}}
\newcommand{\motion}{\mathrm{motion}}
\newcommand{\plan}{\mathrm{plan}}
\newcommand{\IoU}{\mathrm{IoU}}

\renewcommand{\parsection}[1]{\noindent\textbf{#1}:}

\newcommand{\change}[1]{#1}%
\definecolor{roundtwo}{HTML}{D55E00}
\ifdefined\cleanversion\colorlet{roundtwo}{black}\fi%
\newcommand{\changeTwo}[1]{{\color{roundtwo}#1}}
\newcommand{\stk}[2]{\begin{tabular}[c]{@{}c@{}}#1\\#2\end{tabular}}%
\newcommand{\newmon}{RiskMonitor-G}
\newcommand{\tokmon}{RiskMonitor-T}
\newcommand{\newmonS}{RM-G}
\newcommand{\tokmonS}{RM-T}
\usepackage{xcolor}
\usepackage[normalem]{ulem}

\newcommand{\replaceTwo}[2]{\changeTwo{#2}}%
\usepackage{etoolbox}
\makeatletter
\patchcmd{\@maketitle}{\vskip36pt}{\vskip4pt}{}{\errmessage{title patch failed}}
\makeatother

\theoremstyle{thmstyleone}%

\theoremstyle{thmstyletwo}%

\theoremstyle{thmstylethree}%

\begin{document}
\hypersetup{citecolor=black,linkcolor=black,urlcolor=black}

\title[Collision Risk Estimation via Loss Prediction in End-to-End Autonomous Driving]{Collision Risk Estimation via Loss Prediction in End-to-End Autonomous Driving}

\author*[1]{\fnm{Ziliang} \sur{Xiong}}\email{ziliang.xiong@liu.se}

\author[1]{\fnm{Shipeng} \sur{Liu}}\email{shipeng.liu@liu.se}

\author[1]{\fnm{Nathaniel} \sur{Helgesen}}\email{nathaniel.helgesen@liu.se}

\author[3]{\fnm{Hongwei} \sur{Li}}\email{hongwei.bran.li@nus.edu.sg}

\author[1,2]{\fnm{Joakim} \sur{Johnander}}\email{joakim.johnander@zenseact.com}

\author[1]{\fnm{Per-Erik} \sur{Forssén}}\email{per-erik.forssen@liu.se}

\affil*[1]{\orgdiv{Department of Electrical Engineering}, \orgname{Linköping University}, \orgaddress{\city{Linköping}, \postcode{58183},  \country{Sweden}}}

\affil[2]{\orgname{Zenseact}, \orgaddress{\street{Lindholmspiren 2}, \city{Gothenburg}, \postcode{41756}, \country{Sweden}}}

\affil[3]{\orgname{Department of Biomedical Engineering and Department of Diagnostic Radiology, National University of Singapore}, \orgaddress{\street{21 Lower Kent Ridge Road}, \postcode{119077}, \country{Singapore}}}

\abstract{
Collision risk estimation and avoidance play central roles in the safety of autonomous driving (AD) systems.
Recently emerged end-to-end AD systems gain collision avoidance ability by minimizing losses to penalize planning trajectories that are too close to other objects. 
Despite a \replaceTwo{significant}{non-negligible} collision rate during testing, most end-to-end planners do not explicitly quantify the collision risk in their outputs.
To address this, we introduce RiskMonitor, an efficient \replaceTwo{plug-and-play module}{monitor} that interprets \replaceTwo{planning and motion tokens from state-of-the-art}{intermediate outputs of} end-to-end planners to estimate collision risk.
Inspired by loss prediction based uncertainty quantification, RiskMonitor predicts whether the collision loss is positive along planned waypoints, framing collision risk estimation as a binary classification task.
\changeTwo{We study two variants: \tokmon{} reads only the planner's tokens, and \newmon{} additionally reads the relative geometry between each agent's predicted trajectory and the ego plan.}
We evaluate \replaceTwo{RiskMonitor}{both variants} on the real-world nuScenes dataset (open-loop) and the neural-rendering based simulator NeuroNCAP (closed-loop).
\changeTwo{On nuScenes, \newmon{} is significantly better than \tokmon{} and all baselines, including Deep Ensembles, in AUROC, and in average precision on three of the four planners.}
When integrated with a simple braking policy, \replaceTwo{RiskMonitor improves collision avoidance ability by $66.5\%$ in a closed-loop test
on safety-critical scenarios}{\newmon{} reduces the collision rate of UniAD by $62.9\%$ in a closed-loop test on safety-critical scenarios}.
\replaceTwo{These results demonstrate that monitoring collision risk using plan and motion tokens enhances the safety of end-to-end AD without retraining it.}{These results show that the tokens and decoded predictions of end-to-end planners carry a collision-risk signal that a lightweight monitor can extract without retraining the planner.}
}


\maketitle

\section{Introduction}
\label{sec:intro}

\begin{figure*}[htbp!]
    \centering
    \includegraphics[width=\linewidth]{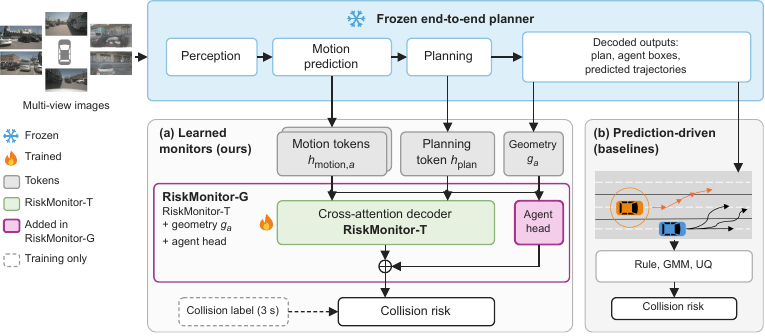}
    \caption{\replaceTwo{Two approaches for monitoring the collision risk in frozen end-to-end planners.  (a). Our proposed token-driven: Train RiskMonitor to decode road user motion tokens and ego planning tokens from E2e planners; (b). Prediction-driven: The monitor could be post-hoc, such as a rule, GMM, or MC Dropout, but these suffer from accumulated prediction error.}{Two approaches for monitoring the collision risk of a frozen end-to-end planner. (a) Learned monitors (ours): \tokmon{} reads the motion tokens and the planning token with a cross-attention decoder. \newmon{} (purple box) reads the same tokens and, in addition, the relative geometry $g_a$ between each agent's predicted trajectory and the ego plan; its per-agent risk head (magenta) adds the largest agent score to the decoder output. Both monitors are trained with the collision label, while the planner stays frozen. (b) Prediction-driven: a post-hoc monitor, such as a rule, a GMM or an uncertainty quantification (UQ) method, compares the planner's decoded predictions with the plan and suffers from accumulated prediction error.}}
    \label{fig:teaser}
\end{figure*}

For safety reasons, collision checking and avoidance are necessary parts of the perception and planning system for autonomous driving.
Recently, several works~\citep{hu2023planning,jiang2023vad,zheng2024genad,chen_eccv24_ppad, sun2025sparsedrive} have proposed end-to-end planners which map sensor input to planned trajectories for the ego-vehicle to follow. 
In these works, collision avoidance is \change{dealt with either during the planning stage by optimizing a collision constraint alongside a large set of other loss terms, or in post-processing techniques such as optimization and scoring plan candidates}.
\change{Such a} collision constraint penalizes planned trajectories that are too close to the trajectories of other objects~\citep{hu2023planning,jiang2023vad}.
However, due to the limited number of safety-critical behaviors in the training data, these methods suffer from a \replaceTwo{significant}{non-negligible} collision rate during both open-loop and closed-loop testing.
One way to improve the collision avoidance performance is to train end-to-end planners on more safety-critical data. 
This is time-consuming and expensive in the real world; thus, research such as \citep{ljungbergh2024neuroncap,jia2024bench2drive} proposes collecting data via simulation.
However, this introduces the risk of a domain shift between the simulated and real-world data, thus hampering the generalization. 
Here, we instead monitor
the collision risk at inference time, without retraining the planner. 

\citet{koopman2016challenges} proposed the decomposition of a monitor and actuator architecture.
The actuator performs the primary functions, and the monitor conducts an acceptance test or other forms of behavioral validation.
End-to-end planners do not explicitly monitor their behavior in this way, e.g., they do not estimate the collision risk of the planned trajectory while driving.
Aligned with Koopman's decomposition philosophy, we propose integrating a monitor module into the end-to-end planner to explicitly estimate the collision risk of the planned trajectory.
A straightforward approach (Fig.~\ref{fig:teaser}b) to such a monitor is to leverage physically interpretable predictions from end-to-end planners, specifically by comparing the planned trajectory for the ego-vehicle with the predicted trajectories of other road users.
This comparison can employ deterministic rules, such as whether intersection over union (IoU) exceeds a threshold, or probabilistic rules, such as the likelihood under a Gaussian mixture model (GMM)  \citep{hu2018gmm}.
However, our experimental results indicate that prediction errors in end-to-end planners render this approach unreliable \changeTwo{for most samples: a rule-based check is precise when the predicted agents overlap the plan, but cannot rank the remaining samples, which form the large majority}.

\replaceTwo{Instead, we propose implementing the monitor module using loss prediction.}{Instead, we implement the monitor with loss prediction.}
\emph{Loss prediction}~\citep{yoo2019learning, kirchhof2023url} is a post-hoc uncertainty quantification method that uses an auxiliary network to map deep features to estimated loss values, providing an error estimate for predictions.
Compared to previous loss prediction work, end-to-end planners are more complex models, supervised by a large number of loss functions, but only one of them (the collision constraint) is of interest.
\replaceTwo{The question we try to answer in this work is}{We ask} \emph{whether collision risk can be estimated through loss prediction}\changeTwo{\emph{, and which outputs of the planner the loss predictor needs}}.

As end-to-end planners are typically partially supervised by a collision loss during training, the collision risk monitor could be a small neural network that maps the end-to-end planner's tokens to an estimate of the collision loss value, which can then be interpreted as the collision risk.
Thus, we introduce a novel and efficient monitor module, RiskMonitor, which predicts the collision risk for the end-to-end plan outcome. 
As Fig.~\ref{fig:teaser} (a) indicates, our RiskMonitor is trained with motion and planning tokens from a given frozen end-to-end planner as inputs and historical collision loss values as targets.
RiskMonitor prediction can be used during inference as a binary collision prediction.
\change{
Different from single-input loss prediction, }\changeTwo{we use the pair of motion and planning tokens as input, since these tokens are the last representations before the agent and ego trajectories are decoded.
The planner also decodes these tokens into explicit agent boxes and trajectories.
We therefore study two variants of RiskMonitor: \tokmon{} reads the tokens only, and \newmon{} additionally reads the relative geometry between each agent's predicted trajectory and the ego plan.
This tests whether the planner's explicit predictions add information beyond its tokens.}

We evaluate \replaceTwo{RiskMonitor}{both monitors} on real-world \emph{open-loop} data \replaceTwo{and data collected
from \emph{closed-loop} simulation. We use nuScenes
\citep{caesar2020nuscenes} and NeuroNCAP \citep{ljungbergh2024neuroncap}.}{from nuScenes \citep{caesar2020nuscenes} and in \emph{closed-loop} simulation with NeuroNCAP \citep{ljungbergh2024neuroncap}.}
\changeTwo{We compare with a rule-based discriminator, a GMM, Monte Carlo Dropout \citep{gal2016dropout}, Deep Ensembles \citep{lakshminarayanan2017simple} and a loss-prediction FCN on four planners.}
We further \replaceTwo{validated the actual collision-avoidance performance}{test in NeuroNCAP whether the alarms of} \replaceTwo{\tokmon{}}{the monitors} \changeTwo{can produce a useful downstream reaction,}\replaceTwo{ in NeuroNCAP by}{} \replaceTwo{harshly braking}{by braking to a standstill} \replaceTwo{if}{when} the predicted collision risk is high.
\replaceTwo{Our results show that this simple strategy reduces the collision rate on NeuroNCAP by $66.5\%$.}{With either monitor, this simple strategy lowers the collision rate of UniAD with its test-time collision optimization.}
\\To summarize, our main \textbf{contributions} are:
\begin{enumerate}
\item We introduce collision risk estimation (CRE) for planned trajectories as a side task for end-to-end AD and propose \replaceTwo{a plug-and-play module}{a monitor}, RiskMonitor, to solve it \changeTwo{for a frozen planner}.

\item We show that loss prediction \replaceTwo{is applicable to end-to-end AD models}{can be used to construct such a monitor for end-to-end planners}\replaceTwo{, which are substantially more complex than models previously tested in loss prediction}{}. We also \replaceTwo{show that loss prediction can be treated as binary classification rather than regression}{extend loss prediction from regression to binary classification and show that its input matters: in addition to the deep features (tokens) used in conventional loss prediction, the planner's explicit outputs, i.e., the predicted agent trajectories relative to the ego plan, substantially improve the monitor}.

\item \changeTwo{On nuScenes with four end-to-end planners, \newmon{} is significantly better than all baselines, including Deep Ensembles, in AUROC, and in average precision on three of the four planners; it captures $50\%$ of all collisions with $46.5\%$--$63.1\%$ precision.}

\item \replaceTwo{In an extensive evaluation across end-to-end planners, RiskMonitor captures $50\%$ of all collisions with $67.5\%$ precision on NeuroNCAP. Moreover,
RiskMonitor reduces the collision rate on NeuroNCAP by $66.5\%$ on average.}{On NeuroNCAP, \newmon{} captures $50\%$ of all collisions with $95.0\%$ precision (UniAD, $50.4\%$ positive samples). To the best of our knowledge, we are the first to evaluate a collision-risk monitor for end-to-end planners in closed-loop simulation, and the first to use its prediction to trigger braking in closed loop, which lowers the collision rate of UniAD on NeuroNCAP from $63.9\%$ to $23.7\%$ ($26.5\%$ with \tokmon{}).}
\end{enumerate}

\section{Related Work}
\label{sec:related_work}

\parsection{End-to-End Autonomous Driving}
\citet{hu2023planning} aim to secure the advantages of both a modular stack and an end-to-end planner. Therefore, their modular method, UniAD, comprises a bird's-eye-view (BEV) encoder~\citep{li2022bevformerlearningbirdseyeviewrepresentation}, a combined detector and tracker~\citep{meinhardt2022trackformer}, a map encoder, a motion forecaster, an occupancy grid estimation module, and a planner, where the information flow between different modules is learned end-to-end. Several works follow this design.
VAD \citep{jiang2023vad} is more computationally efficient because it utilizes a vectorized representation. 
PPAD~\citep{chen_eccv24_ppad} iteratively couples prediction and planning to enable adaptive, interaction-aware planning.
\citet{zheng2024genad} employ a gated recurrent unit to progressively predict the next future in the latent space and use a decoder to obtain explicit trajectories.
{
SparseDrive~\citep{sun2025sparsedrive} replaces the BEV representation with a sparse representation and uses a two-stage parallel architecture to replace the cascaded modular design.}
These works have in common that the planning heads are all explicitly supervised by a collision loss term that punishes the planned trajectory for overlapping with obstacles. \changeTwo{This inspires us to use a loss predictor to construct the proposed RiskMonitor.} %
We use UniAD, VAD, GenAD and SparseDrive in our open-loop experiments. %
UniAD and VAD are shown to work with the closed-loop simulator NeuroNCAP, which we use to evaluate RiskMonitor in safety-critical scenarios.

\parsection{Collision Risk Modelling and Avoidance}
For safe planning in AD, one must quantify the collision risk associated with the planned trajectory. 
There are two main approaches to probabilistic CRE: collision state probability (CSP) and collision event probability (CEP) \citep{philipp2019analytic}. CSP estimates the probability density that two objects spatially overlap at a specific future time.
\citet{houenou_risk,lambert_mccollision,toit_probabilistic} use Monte Carlo to calculate CSP but suffer from high sample complexity and latency.
CEP extends this by estimating the probability density that a collision occurs at any point over a future time horizon~\citep{altendorfer_whatis,nordlund2008conflict}.
\citet{philipp2019analytic} use the geometric simplification of a collision octagon to enable efficient closed-form approximations under Gaussian uncertainty. 
However, these rule-based methods rely on simplified scene geometry and have not been evaluated on large driving datasets.
As risk modelling is a context-dependent task, it has been integrated into collision-avoidance planning \citep{9511299, 10598864} in the field of model predictive control (MPC).
\changeTwo{End-to-end planners lack such an explicit risk estimate, which RiskMonitor provides as a monitor.}

\parsection{Uncertainty Quantification} 
Trajectory forecasting methods typically model future object states probabilistically~\citep{hu2021fiery,akan2022stretchbev,deo2022multimodal}.
\citet{inproceedings} boost the generation of physically feasible trajectories by incorporating guidance from a UQ module, yet the UQ performance is not explicitly evaluated. 
In out-of-distribution (OOD) scenarios ~\citep{malinin_shifts_2021,bahari_vehicle_2021}, performance degradation and lack of predictive uncertainty can lead to severe consequences.
\citet{wiederer2023joint} use an encoder-decoder-like model to predict trajectory and then jointly estimate OOD score and in-distribution (ID) uncertainty score.
They treat ID uncertainty as a LP regression task \citep{yu2024discretization,yoo2019learning}, training a small MLP to predict future trajectory error.
Though similar, we instead frame loss prediction as a classification task, which improves interpretability and enables integration with more complex end-to-end planners.
In end-to-end planners \citep{hu2023planning, jiang2023vad}, the future trajectory of a road user is modeled as an ensemble of possible candidates along with probabilities.
The uncertainty in these methods quantifies the ambiguity of the trajectory and not the collision risk.
\changeTwo{Ensembles have also been applied to multimodal trajectory forecasting: \citet{lafage2025hierarchical} train lightweight Transformer ensembles with grouped attention layers to approach Deep Ensemble \citep{lakshminarayanan2017simple} performance at a fraction of its cost.
Like BatchEnsemble \citep{wenbatchensemble}, such efficient ensembles are evaluated on trajectory accuracy (e.g., ADE/FDE) and likelihood, and do not provide an estimate of whether the ego plan will cause a collision.}
\citet{cai2020probabilistic} introduce a probabilistic motion planning network which outputs a distribution of future motion based on the Gaussian mixture model, but without explicit collision risk awareness.
Concurrent work, Redoubt~\citep{wangredoubt}, proposes to predict the collision risk of object-level planners, assuming the availability of ground-truth objects and lanes. In contrast, our work focuses on end-to-end planning and presents results in safety-critical scenarios.

\changeTwo{
\parsection{Runtime Safety Monitoring}
\label{sec:rw-monitoring}
A runtime safety monitor observes a machine-learning component during operation and flags inputs or outputs on which the component is likely to fail.
\citet{ferreira2025safety} survey such monitors and argue that they should be evaluated by their ability to detect incorrect predictions, rather than by the detection of distribution shift as such: not every out-of-distribution input is dangerous, and not every in-distribution input is safe.
Following this criterion, we evaluate our monitors by how well they detect plans that lead to a collision on driving sequences unseen during training.
Failure detectors have also been proposed for vision-language models \citep{dong2026to} and multimodal classifiers \citep{Liu_2026_CVPR}; they are motivated by safety-critical applications such as autonomous driving, but are evaluated only on static test sets, and whether the detected failures can be acted upon is not tested.
In contrast, we evaluate our monitors in closed-loop simulation and use their output to trigger braking.
A complementary line of work verifies the intended trajectory online against predictions of the other road users and, if it is unsafe, executes a fail-safe trajectory to a standstill~\citep{pek2020using}.
Such verification relies on accurate predictions of the other road users, which an end-to-end planner does not guarantee; our monitors instead learn from the planner's outputs which plans are likely to cause a collision.
In both cases the monitor and the reaction are separate components~\citep{koopman2016challenges}: the monitor estimates the risk, and a downstream policy decides how to react, e.g.\ by invoking the braking policy of Sec.~\ref{sec:simulation} or a fail-safe trajectory.
}

\parsection{Closed-loop Evaluation}
End-to-end planning has been analysed both in the CARLA~\citep{dosovitskiy2017carla} closed-loop simulator~\citep{chen2021learning,codevilla2018end,prakash2021multi,wu2023policy} and on real data~\citep{hu2021safe,khurana2022differentiable,hu2022st}.
In NeuroNCAP \citep{ljungbergh2024neuroncap}, an end-to-end planner receives sensor input and outputs the planned trajectory.
The planned trajectory is then passed to a controller and vehicle kinematics model that moves the ego vehicle forward. Finally, the NeRF-based renderer NeuRAD~\citep{tonderski2024neurad} creates sensor data for the next frame of the 3D scene.
In this way, end-to-end AD models can be evaluated in a closed-loop manner.
NeuroNCAP generates three types of safety-critical scenarios, namely ``stationary", ``side", and ``frontal".
We adopt NeuroNCAP rather than CARLA since it concentrates on safety-critical scenarios. 

\section{Preliminaries}
Sec.~\ref {sec:formulation} first gives the task formulation of collision risk estimation.
Afterwards, Sec.~\ref {sec:loss_pred} reviews the theory of loss prediction.

\subsection{Task Formulation}
\label{sec:formulation}
Modern end-to-end planners typically predict trajectories over a short fixed time horizon.
Unlike previous methods that calculate CEP density as a time-evolving process \citep{philipp2019analytic}, 
we aim to directly predict the event-level collision probability over the entire horizon, i.e., 
the probability that a collision occurs at any time within the horizon.
More specifically, the planning outcome is a unimodal trajectory\footnote{Though some planners predict a planning vocabulary, the controller requires a \emph{unimodal}, or singular, trajectory.} %
consisting of waypoints,~$\hat{\tau}_\plan\in \mathbb{R}^{N_T\times2}$, where~$N_T$ is the number of predicted future waypoints in the time horizon $T$ and $2$ corresponds to 2D coordinates in the ego-vehicle centered BEV frame.

The collision risk over the time horizon $T$ is then formally expressed as 
\begin{multline}
    \mathrm{P}(\collision\mid \hat{\tau}_\plan) \\= \int^{T}_0 \mathrm{p}(\sum_i^{N_a}\IoU(\hat{\tau}_\plan(t), \tau_\motion^i(t))>0 ) dt\enspace,
\end{multline}
where the true future trajectory for the $i$th road user is $\tau_\motion^i \in \mathbb{R}^{N_T\times2}$ and the total number of road users is $N_a$. 
With a slight abuse of notations, both $\hat{\tau}_\plan$ and $\tau_\motion^i$ take the object geometry dimension into account during IoU calculation.

\subsection{Loss Prediction} 
\label{sec:loss_pred}
The core idea of loss prediction is to add an auxiliary module to a target model $\Theta_\target$~\citep{yoo2019learning}, for instance, a classifier, to predict its loss value as an indicator of prediction uncertainty.
Given a training sample pair $(\mathbf{x}, y)$, we obtain a target prediction $\hat{y}$ and intermediate features $h_i$ through the target model as in Eq.~(\ref{eq:target-pred}).
\begin{align}
    \hat{y}, h_1,..., h_n &= \Theta_\target(\mathbf{x})\enspace,\label{eq:target-pred}\\
    h &= \{h_1, h_2,..., h_n\}, n\geq1\enspace, \label{eq:feature-set}\\
    L_\target &= \mathcal{L}_\target(\hat{y}~,~y) \label{eq:target-loss}\enspace.
\end{align}

The target loss value $L_\target$ can be computed with the annotation~$y$ as in Eq.~(\ref{eq:target-loss}).
Meanwhile, an auxiliary module $\Theta_\loss$ predicts the loss value of the target model with a set of features $h$ as input,
\begin{align}
    \hat{L}_\target &= \Theta_\loss(h)\enspace,\label{eq:aux-pred}\\
    L_\loss &= \mathcal{L}_\loss(\hat{L}_\target, L_\target)\enspace.\label{eq:aux-loss}
\end{align}
The target loss value $L_\target$ serves as the ground truth for the loss predictor $\Theta_\loss$.
Formally, the total loss function used to train the target model $\Theta_\target$ and the loss prediction module $\Theta_\loss$ is
\begin{equation}
    \label{eq:loss_pred}
    \mathcal{L}_\target(\hat{y}~,~y) + \lambda \mathcal{L}_\loss(\hat{L}_\target, L_\target)\enspace,
\end{equation}
where~$\lambda$ is a predefined constant scaling factor.
Previous loss prediction work typically trains the auxiliary module as a regressor using $L_2$ loss \citep{yoo2019learning} or a ranking-based objective \citep{kirchhof2024pretrained} to detach from the scale of target loss $L_\target$. 

To avoid~$\Theta_\loss$ from interfering with~$\Theta_\target$ during training, \citet{kirchhof2024pretrained} proposed to pretrain $\Theta_\target$ and then freeze it. To speed up training,~$(h, L_\target)$ pairs are cached during the training of~$\Theta_\target$, and~$\Theta_\loss$ is then trained separately.
After training, the loss prediction module~$\Theta_\loss$ becomes an uncertainty quantifier, with a higher predicted value indicating that the target model is more uncertain~\citep{kirchhof2023url}. Next, we extend loss prediction to predict whether modern end-to-end planners will generate colliding plans.

\section{Method}
\label{sec:method}
We adapt loss prediction to \replaceTwo{collision risk estimation (CRE)}{CRE} for end-to-end planners in Sec.~\ref{sec:collision_prediction}.
Sec.~\ref{sec:riskmonitor} introduces our RiskMonitor architecture.
\replaceTwo{Finally, }{}Sec.~\ref{sec:imbalance} introduces techniques for handling class imbalance in real-world data.
\changeTwo{Finally, Sec.~\ref{sec:newmon} extends RiskMonitor with explicit per-agent geometry.}

\subsection{Collision risk estimation}
\label{sec:collision_prediction}
For driving safety, we are particularly interested in predicting whether the planned trajectory will cause a collision. 
To train RiskMonitor, we first need to determine which loss value in the end-to-end AD model should be used as the training label~$\hat{L}_\target$ and which features to include in the input set~$h$.

\parsection{Choice of target loss} 
Prior work on loss prediction \citep{yoo2019learning, kirchhof2023url} typically considers models supervised by a small number of loss terms;
in contrast, state-of-the-art end-to-end planners are more complex and supervised by hundreds of loss terms. 
They comprise multiple modules designed to handle different tasks, e.g.,  motion prediction and planning \citep{hu2023planning, jiang2023vad}.
There is no single loss as in standard loss prediction tasks; instead, different modules have distinct functions, and each can have multiple parts.
This results in a large and heterogeneous set of loss components, raising the question of which loss should be selected as the prediction target.

For training RiskMonitor, we focus on the collision loss component present in the planning module.
The planning module loss function has multiple parts, including an imitation loss~$\mathcal{L}_\imitation$ and a collision loss~$\mathcal{L}_\collision$. 
The imitation loss is usually the ~$L_2$ distance to the ground-truth trajectory so that the model learns to exhibit a human-like planning capability, while the collision loss penalizes plans that result in collisions with other road users on the road.
To apply loss prediction, we adopt the collision loss component from the last transformer layer, $\mathcal{L}_\collision$, and train $\Theta_\loss$ to predict its value. 

Our RiskMonitor frames loss prediction as a binary classification task. 
The scale of~$L_\collision$ does not reflect the real-world accident severity of collisions, as it is the sum of IoU and does not account for speed. 
Consequently, treating CRE as a regression task—as is standard in prior loss prediction work~\citep{yoo2019learning, kirchhof2024pretrained}—is suboptimal in this setting.
Instead, we note that whether the collision loss value is positive indicates if a collision is going to occur along the planned trajectory.\footnote{See Eq. (4) in VAD \citep{jiang2023vad} for an example.} 
Thus, it is natural to treat CRE as a binary classification task instead and predict the probability of a collision occurring along the planned trajectory~$\hat{\tau}_\plan$,

\begin{equation}
     \mathrm{P}(\collision\mid \hat{\tau}_\plan) = \Theta_\riskmonitor(h)\enspace.
     \label{eq:collision_prob}
\end{equation}
Then, RiskMonitor is trained using 
\begin{equation}
    L_\loss =  \mathcal{L}_\riskmonitor(\mathrm{P}(\collision\mid \hat{\tau}_\plan), L_\collision)\enspace,
\end{equation}
where $\mathcal{L}_\riskmonitor$ is the cross-entropy loss or the focal loss if there is class imbalance (See Sec. \ref{sec:imbalance}).

\parsection{Choice of features} 
An open question is which features~$h$ should be fed into~$\Theta_\riskmonitor$. 
Given a training sample~$\mathbf{x}$, each module of the end-to-end planner produces separate predictions using many consecutive intermediate features. 
Normally, rule-based methods, e.g.~in Sec.~\ref{sec:CP}, make use of the motion prediction~$\hat{\tau}_\motion$ and the planned ego trajectory~$\hat{\tau}_\plan$ to predict CEP. 
We note that this motion prediction is intentionally stochastic \citep{hu2023planning, jiang2023vad}, as end-to-end AD models typically predict multimodal trajectories to balance the unpredictability of other drivers' intentions.
Since both~$\hat{\tau}_\plan$ and~$\hat{\tau}_\motion$ are outputs from individual transformer modules, we can also output the respective planning token~$h_\plan$ and road user tokens~$h_\motion$. Because these tokens are the layers directly before~$\hat{\tau}_\plan$ and~$\hat{\tau}_\motion$, they conceivably contain the necessary agent-level spatial-temporal information; thus, we opt to use them as inputs. 
Formally, the outputs we cache from the end-to-end planner are 
\begin{equation}
    \hat{\tau}_\plan, \hat{\tau}_\motion, h_\plan, h_\motion = \Theta_\planner(\mathbf{x})\enspace.
     \label{eq:AD-pred}
\end{equation}

\subsection{RiskMonitor Architecture}
\label{sec:riskmonitor}
To model the interaction between the planned trajectory and surrounding agents, we employ a transformer decoder architecture. \changeTwo{We refer to the model of this section, which reads the tokens only, as \tokmon{}; Sec.~\ref{sec:newmon} extends it with explicit per-agent geometry.} The decoder takes the planning tokens $h_{\text{plan}} \in \mathbb{R}^{d}$ as input and attends to the set of agent-motion tokens $H_{\text{motion}} \in \mathbb{R}^{N_a \times d}$, where~$N_a$ is the number of road users. 
Specifically, original motion tokens are in the shape of $h_\motion \in \mathbb{R}^{N_a\times N_m \times d}$, where~$N_m$ is the number of modalities.
Motion tokens are passed through an MLP fuser to flatten out the second dimension.

The core operation is the multi-head cross-attention mechanism \citep{vaswani17}, which computes attention weights between the query and the key-value pairs derived from $H_{\text{motion}}$. For each head $i \in \{1,\dots,h\}$, we define
\begin{equation}
\text{head}_i = \mathrm{softmax}\!\left(
\frac{(Q W^Q_i)(K W^K_i)^\top}{\sqrt{d_k}}
\right)(V W^V_i)\enspace,
\end{equation}
where $Q = h_{\text{plan}}$, $K = H_{\text{motion}}$, $V = H_{\text{motion}}$. The matrices $W^Q_i, W^K_i, W^V_i$ are learned projections for the $i$-th head, and $d_k$ is the dimension per head.
The outputs of all heads are concatenated and projected to form the decoder output as
\begin{equation}
z = \mathrm{Concat}\big[\text{head}_1,\dots,\text{head}_h\big] W^O\enspace,
\end{equation}
where $W^O \in \mathbb{R}^{(h \cdot d_v) \times d}$ is the output projection matrix.
Finally, the collision risk is predicted by passing $z$ through a feed-forward network followed by a sigmoid activation as
\begin{equation}
\Theta_{\text{RiskMonitor}}(h) = \sigma(\mathrm{MLP}(z))\enspace.
\end{equation}
This formulation allows the decoder to capture complex interactions between the planned trajectory and multiple agents across modalities through multi-head attention.\\\\

\subsection{Handling Data Imbalance}
\label{sec:imbalance}
Real-world AD datasets are typically imbalanced as samples without collisions greatly outnumber those with collisions.
To alleviate this issue, we propose combining the bagging ensemble and undersampling \citep{westny2021vehicle} with the focal loss \mbox{\citep{lin2017focal}.}
Given the set of collision samples~$P$ and the set of noncollision class examples~$Q$,~$|P| < |Q|$.
In bagging,~$Q$ is split into~$N$ subsets,~$Q_1,~Q_2,~\cdots,~Q_N$, and each subset is combined with~$P$ to form~$N$ training sets.
Then,~$N$ models are trained independently with a focal loss. The final prediction is the average of the predictions of the~$N$ models.
We use a focal loss with the default~$\gamma~=~2$, and set the class weight~$\alpha$, to the inverse of the class ratio in each of the~$N$ bags.
\begin{equation}
    \alpha = \frac{|Q|}{N|P|+|Q|}\enspace.
\end{equation}
\replaceTwo{The optimal~$N$ is chosen via a validation set.}{$N$ is chosen on a held-out subset of the training scenes.}
\replaceTwo{Moreover, RiskMonitor suffers from a distribution shift between the training and evaluation sequences (NuScenes and NeuroNCAP, see Fig.~\mbox{\ref{fig:tsne}}). We alleviate this problem with Mixup~\mbox{\citep{zhang2017mixup}}, which improves model generalization on OOD data \mbox{\citep{thulasidasan2019mixup}}.}{}

\changeTwo{
\subsection{\newmon: RiskMonitor with Explicit Per-Agent Geometry}
\label{sec:newmon}
\begin{figure}[t]
    \centering
    \includegraphics[width=\columnwidth]{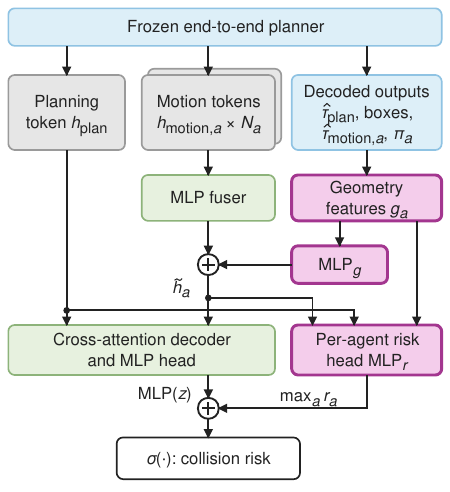}
    \caption{\changeTwo{Architecture of \newmon{}. The frozen planner provides the planning token $h_\plan$, the motion tokens $h_{\motion,a}$ and its decoded outputs. The relative geometry features $g_a$, computed with the same definition on all planners, are embedded and added to each fused agent token before the cross-attention decoder of \tokmon{}, and a per-agent risk head adds the largest agent score to the logit of the \tokmon{} head. Components added to \tokmon{} are highlighted.}}
    \label{fig:newmon}
\end{figure}

In \tokmon{}, the relation between each agent and the ego plan is encoded only implicitly, through the cross-attention of the planning token to the motion tokens.
The planner, however, also decodes the motion tokens into explicit agent boxes and trajectories, from which this relation can be computed directly.
\newmon{} therefore keeps the tokens and the cross-attention decoder of \tokmon{} and adds two components: the relative geometry between each agent's predicted trajectory and the ego plan, and a per-agent risk head (Fig.~\ref{fig:newmon}).

\parsection{Inputs}
In addition to the planning and motion tokens ($h_\plan$ and $h_\motion$), \newmon{} also reads the planner's decoded outputs: the planned ego waypoints $\hat{\tau}_\plan$, and, for each agent, its detected box and its predicted multimodal trajectory $\hat{\tau}_{\motion,a}$ with mode probabilities $\pi_a$.
From these it computes, for each agent, a vector of relative geometry features $g_a$: the displacement between the agent and the ego waypoint at each future step, for the most likely mode and for all modes, the mode probabilities, and the agent's box attributes (Appendix~\ref{sec:cache}).
These features have the same definition in all tested planners, and no further internal representation of the planner is accessed.

\parsection{Architecture}
Before the cross-attention, the geometry features are embedded and added to each fused agent token,
\begin{equation}
\tilde{h}_a = \mathrm{ReLU}\big(W h_{\motion,a}\big) + \mathrm{MLP}_g(g_a)\enspace,
\end{equation}
where $\mathrm{ReLU}(W\,\cdot)$ is the MLP fuser of \tokmon{} (Sec.~\ref{sec:riskmonitor}) and $\mathrm{MLP}_g$ has one hidden layer of 256 units.
As in \tokmon{}, the planning token cross-attends to $\{\tilde{h}_a\}$ in one decoder layer, which gives $z$.
Additionally, a per-agent risk head scores each agent from the planning token, its fused token and its geometry, and the largest score is added to the logit of the \tokmon{} head,
\begin{align}
r_a &= \mathrm{MLP}_r\big([h_\plan, \tilde{h}_a, g_a]\big)\enspace,\\
\Theta_{\text{\newmon}}(h) &= \sigma\Big(\mathrm{MLP}(z) + \max_{a} r_a\Big)\enspace.
\end{align}
We take the maximum because a single colliding agent suffices for a collision.
The output layers of $\mathrm{MLP}_g$ and $\mathrm{MLP}_r$ are initialized to zero, so that both branches output zero at initialization and \newmon{} starts as the \tokmon{} architecture.
\newmon{} is trained with the same bagging ensemble and focal loss as \tokmon{} (Sec.~\ref{sec:imbalance}) but without mixup; its components are ablated in Sec.~\ref{sec:ablation}.

}

\section{Experiments}
\label{sec:exp}
We first introduce the baselines to which \tokmon{} \changeTwo{and \newmon{} are} compared \replaceTwo{in Sec~\ref{sec:CP}; and evaluation metrics in Sec.~\ref{sec:metrics}}{(Sec.~\ref{sec:CP}) and the evaluation metrics (Sec.~\ref{sec:metrics})}. We then evaluate \changeTwo{the monitors} on real-world driving logs (Sec.~\ref{sec:realworld})\changeTwo{, ablate the components of \newmon{} (Sec.~\ref{sec:ablation}), measure false alarms at the event level (Sec.~\ref{sec:event}),} and \changeTwo{evaluate them} in neural-rendered closed-loop simulation (Sec.~\ref{sec:simulation}).\replaceTwo{ Finally, we ablate the components of \newmon{} in Sec.~\ref{sec:ablation}.}{}

\subsection{Collision Predictors} 
\label{sec:CP}
To the best of our knowledge, there is no existing work on collision risk estimation for end-to-end planners. We therefore compare to \changeTwo{four} prediction-driven baselines which map predictions to collision risk and one token-driven baseline (see Fig.~\ref{fig:teaser}).
\changeTwo{Implementation details of the baselines are given in Appendix~\ref{sec:CP-A}, and the training details of the learned monitors in Appendix~\ref{sec:train-detail}.}

\parsection{Rule-based Discriminator (RBD)}
We compare the planning trajectory with the future locations of other road users and \replaceTwo{see if they will intersect}{check whether they intersect}.
Since the true future locations are not available, motion prediction is used as a proxy. 
\replaceTwo{We take the collision loss function in UniAD and VAD and feed their own planning waypoints, detected bounding boxes, and motion trajectories.
With their own detected bounding boxes, the geometric dimensions of traffic road users are accounted for.}{}
\changeTwo{Specifically, RBD applies the rule that defines the collision labels for all four planners (Sec.~\ref{sec:realworld}), the UniAD collision loss with a $1.0$\,m safety distance and axis-aligned boxes, to the planner's own predictions, i.e., its detected boxes and the most likely predicted mode of each agent; the score is the overlap summed over agents and time steps.}

\parsection{GMM}
In end-to-end planners, each road user's multimodal trajectory,~$\hat{\tau}_\motion$, at each step is described as a GMM. 
Specifically, the trajectory of each road user consists of a set of possible future trajectories, each with corresponding probabilities.
We use this set to define a sequence of chained GMMs.
Collision risk is a combinatorial product of the collision probability in each GMM.
\replaceTwo{See Sec.~\ref{sec:GMM-A} for details about these predictors.}{}

\parsection{Monte Carlo Dropout}
\replaceTwo{MCD is a UQ technique that uses dropout during inference by performing multiple stochastic forward passes and obtaining an ensemble of predictions.}{MC Dropout is a UQ technique that keeps dropout active at inference and obtains an ensemble of predictions from multiple stochastic forward passes.}
We use the dropout in the \replaceTwo{last linear layer}{MLP fuser} of the planning head\replaceTwo{ (See Sec.~\ref{sec:drop})}{} to obtain an ensemble of planned trajectories $\{\hat{\tau}_\plan^i\}$.
The covariance of the endpoints of $\{\hat{\tau}_\plan^i\}$ is calculated as a generic uncertainty score, and we test whether it can serve as a collision risk estimation.

\changeTwo{
\parsection{Deep Ensemble}
Deep Ensemble \citep{lakshminarayanan2017simple} is a strong, widely used baseline for uncertainty quantification.
For UniAD and VAD, we train five ensemble members with different random seeds by re-initializing and retraining the modules of the planner's second training stage.
Within our compute budget, we build the Deep Ensemble for these two planners only.
Tab.~\ref{tab:nuscenes} reports DE-RBD, a vote of RBD over the members, which is the best of the four ensemble scores in AP (Appendix~\ref{sec:DE}).
}

\parsection{LP-FCN}
We adopt the model in previous loss prediction work \citep{kirchhof2024pretrained, kirchhof2023url}, a two-layer fully-connected network that maps the plan token to the loss value of an end-to-end planner.
\replaceTwo{As the plan token is obtained with BEV feature awareness, FCN has the potential to generalize.
But it is not aware of the surrounding road users.}{Since the plan token is computed with awareness of BEV features, LP-FCN may generalize, but it is not aware of the surrounding road users;}
\changeTwo{comparing it with \tokmon{} isolates the contribution of the motion tokens and the cross-attention.}

\begin{table*}[t]
\centering
\scriptsize
\setlength\tabcolsep{1.2mm}
{\color{roundtwo}
\begin{tabular}{@{}llccccc@{}}
\toprule
Planner & Method & AUROC & AP & $\mathrm{Pr}_{30}$ & $\mathrm{Pr}_{50}$ & $\mathrm{Pr}_{70}$ \\ \midrule
\multirow{7}{*}{UniAD} & MC Dropout & 63.98\,{\tiny[58.9, 68.7]} & 11.60\,{\tiny[7.9, 16.1]} & 13.59\,{\tiny[8.6, 20.2]} & 12.90\,{\tiny[8.6, 17.0]} & 11.07\,{\tiny[7.7, 15.4]} \\
 & GMM & 75.99\,{\tiny[69.6, 81.8]} & 17.31\,{\tiny[11.1, 25.7]} & 19.56\,{\tiny[11.6, 34.5]} & 19.11\,{\tiny[11.8, 28.4]} & 16.66\,{\tiny[10.8, 23.8]} \\
 & RBD & 78.57\,{\tiny[72.6, 84.2]} & 36.59\,{\tiny[23.6, 55.6]} & 57.03\,{\tiny[38.6, 82.7]} & 57.85\,{\tiny[39.2, 76.0]} & 19.27\,{\tiny[11.4, 63.9]} \\
 & LP-FCN & 80.99\,{\tiny[76.2, 85.2]} & 31.29\,{\tiny[22.5, 41.4]} & 37.63\,{\tiny[20.6, 55.4]} & 28.40\,{\tiny[19.7, 38.2]} & 23.02\,{\tiny[14.2, 33.8]} \\
 & Deep Ensemble & 81.53\,{\tiny[75.1, 86.9]} & \underline{42.74}\,{\tiny[30.5, 56.2]} & \underline{62.31}\,{\tiny[45.5, 79.0]} & \underline{58.98}\,{\tiny[40.5, 76.8]} & \underline{35.54}\,{\tiny[14.1, 58.1]} \\
 & \tokmonS{} & \underline{82.39}\,{\tiny[77.7, 86.4]} & 35.95\,{\tiny[25.1, 46.6]} & 42.76\,{\tiny[28.5, 57.2]} & 33.51\,{\tiny[20.8, 45.7]} & 23.94\,{\tiny[15.9, 32.8]} \\
 & \newmonS{} & \textbf{90.27}\,{\tiny[85.9, 94.0]} & \textbf{61.79}\,{\tiny[48.8, 72.9]} & \textbf{92.61}\,{\tiny[72.1, 99.1]} & \textbf{63.09}\,{\tiny[40.8, 87.9]} & \textbf{37.22}\,{\tiny[20.9, 57.4]} \\
\midrule
\multirow{6}{*}{VAD} & GMM & 70.90\,{\tiny[65.7, 75.7]} & 19.36\,{\tiny[12.9, 28.0]} & 18.33\,{\tiny[11.9, 33.4]} & 16.14\,{\tiny[11.2, 23.3]} & 14.21\,{\tiny[10.8, 18.7]} \\
 & RBD & 71.61\,{\tiny[66.7, 76.3]} & 37.87\,{\tiny[28.3, 47.0]} & \textbf{71.63}\,{\tiny[60.4, 80.5]} & 27.73\,{\tiny[13.7, 70.1]} & 11.14\,{\tiny[8.3, 15.3]} \\
 & LP-FCN & 77.03\,{\tiny[72.0, 81.5]} & 20.21\,{\tiny[15.4, 25.8]} & 26.56\,{\tiny[19.1, 32.5]} & 22.66\,{\tiny[16.3, 29.6]} & 18.28\,{\tiny[12.6, 24.0]} \\
 & Deep Ensemble & 70.01\,{\tiny[65.1, 74.6]} & 30.43\,{\tiny[22.5, 38.9]} & 61.71\,{\tiny[49.5, 71.3]} & 21.37\,{\tiny[12.0, 61.6]} & 10.60\,{\tiny[7.9, 14.2]} \\
 & \tokmonS{} & \underline{85.05}\,{\tiny[81.9, 88.1]} & \underline{40.23}\,{\tiny[30.3, 49.3]} & 47.26\,{\tiny[32.0, 66.0]} & \underline{35.37}\,{\tiny[23.5, 43.1]} & \underline{22.71}\,{\tiny[17.4, 29.6]} \\
 & \newmonS{} & \textbf{90.71}\,{\tiny[88.3, 92.7]} & \textbf{50.95}\,{\tiny[40.9, 59.8]} & \underline{63.45}\,{\tiny[44.6, 79.8]} & \textbf{46.53}\,{\tiny[34.0, 60.1]} & \textbf{37.42}\,{\tiny[28.0, 48.4]} \\
\midrule
\multirow{5}{*}{GenAD} & GMM & 73.47\,{\tiny[68.4, 78.5]} & 14.63\,{\tiny[10.5, 20.5]} & 16.46\,{\tiny[10.5, 26.0]} & 15.85\,{\tiny[11.5, 22.1]} & 15.09\,{\tiny[11.1, 20.2]} \\
 & RBD & 76.94\,{\tiny[71.9, 81.4]} & 44.18\,{\tiny[34.6, 52.7]} & \underline{71.37}\,{\tiny[61.5, 80.9]} & \textbf{65.95}\,{\tiny[29.3, 74.4]} & 14.71\,{\tiny[9.9, 27.1]} \\
 & LP-FCN & 81.09\,{\tiny[76.8, 85.0]} & 36.65\,{\tiny[28.3, 45.6]} & 54.60\,{\tiny[34.7, 68.6]} & 28.76\,{\tiny[19.0, 43.5]} & 18.46\,{\tiny[13.8, 26.5]} \\
 & \tokmonS{} & \underline{87.33}\,{\tiny[83.7, 90.4]} & \underline{46.10}\,{\tiny[36.9, 55.1]} & 66.20\,{\tiny[44.6, 77.4]} & 41.66\,{\tiny[30.2, 54.5]} & \underline{31.27}\,{\tiny[21.2, 40.9]} \\
 & \newmonS{} & \textbf{90.31}\,{\tiny[87.3, 92.8]} & \textbf{56.03}\,{\tiny[48.3, 63.8]} & \textbf{76.98}\,{\tiny[62.5, 87.5]} & \underline{59.27}\,{\tiny[47.5, 70.1]} & \textbf{42.83}\,{\tiny[30.5, 53.2]} \\
\midrule
\multirow{5}{*}{SparseDrive} & GMM & 76.18\,{\tiny[71.4, 80.9]} & 16.73\,{\tiny[11.7, 24.3]} & 17.69\,{\tiny[11.8, 31.3]} & 17.49\,{\tiny[11.9, 25.9]} & 15.64\,{\tiny[11.6, 20.5]} \\
 & RBD & 77.78\,{\tiny[72.4, 82.8]} & \underline{47.53}\,{\tiny[37.2, 57.9]} & \underline{76.95}\,{\tiny[66.3, 85.9]} & \textbf{72.83}\,{\tiny[32.2, 80.3]} & 15.21\,{\tiny[9.9, 36.1]} \\
 & LP-FCN & 80.00\,{\tiny[75.4, 83.8]} & 24.77\,{\tiny[17.9, 32.5]} & 29.57\,{\tiny[18.9, 40.7]} & 23.09\,{\tiny[15.1, 35.8]} & 18.28\,{\tiny[12.4, 25.6]} \\
 & \tokmonS{} & \underline{87.63}\,{\tiny[84.7, 90.3]} & 45.97\,{\tiny[36.2, 55.3]} & 58.68\,{\tiny[43.2, 74.3]} & 43.33\,{\tiny[28.8, 55.0]} & \underline{27.97}\,{\tiny[19.7, 38.0]} \\
 & \newmonS{} & \textbf{90.78}\,{\tiny[88.0, 93.1]} & \textbf{54.62}\,{\tiny[45.7, 63.3]} & \textbf{77.84}\,{\tiny[64.2, 86.5]} & \underline{54.33}\,{\tiny[38.4, 75.2]} & \textbf{36.76}\,{\tiny[25.5, 51.2]} \\
\bottomrule
\end{tabular}}
\caption{\changeTwo{Classification performance of the collision predictors on the open-loop benchmark nuScenes (validation split, \%), for UniAD, VAD, GenAD and SparseDrive. Bold: highest, underlined: second-highest point estimate per planner; significance of the differences is given in Tab.~\ref{tab:paired}. Brackets: $95\%$ CIs from a bootstrap over the 150 validation scenes. $\mathrm{Pr}_X$ is the precision at recall $X$ (Sec.~\ref{sec:metrics}). Deep Ensemble: the DE-RBD score (Appendix~\ref{sec:DE}). RM-T/RM-G: \tokmon{}/\newmon{}. MC Dropout was evaluated on UniAD only, and Deep Ensemble on UniAD and VAD.}}
\label{tab:nuscenes}
\end{table*}

\begin{table*}[t]
\centering
\scriptsize
\setlength\tabcolsep{0.8mm}
{\color{roundtwo}
\begin{tabular}{@{}lcccccccc@{}}
\toprule
 & \multicolumn{2}{c}{UniAD} & \multicolumn{2}{c}{VAD} & \multicolumn{2}{c}{GenAD} & \multicolumn{2}{c}{SparseDrive} \\
Difference & AUROC & AP & AUROC & AP & AUROC & AP & AUROC & AP \\ \midrule
\newmonS{} $-$ \tokmonS{} & \stk{\textbf{$+$7.9}}{{\tiny[3.6, 12.4]}} & \stk{\textbf{$+$25.8}}{{\tiny[16.2, 35.9]}} & \stk{\textbf{$+$5.7}}{{\tiny[2.9, 8.3]}} & \stk{\textbf{$+$10.7}}{{\tiny[4.5, 17.0]}} & \stk{\textbf{$+$3.0}}{{\tiny[1.1, 5.1]}} & \stk{\textbf{$+$9.9}}{{\tiny[3.9, 16.5]}} & \stk{\textbf{$+$3.1}}{{\tiny[1.2, 5.1]}} & \stk{\textbf{$+$8.6}}{{\tiny[2.1, 15.8]}} \\
\newmonS{} $-$ RBD & \stk{\textbf{$+$11.7}}{{\tiny[8.3, 15.1]}} & \stk{\textbf{$+$25.2}}{{\tiny[9.5, 36.5]}} & \stk{\textbf{$+$19.1}}{{\tiny[14.4, 24.0]}} & \stk{\textbf{$+$13.1}}{{\tiny[5.4, 21.2]}} & \stk{\textbf{$+$13.4}}{{\tiny[9.0, 18.5]}} & \stk{\textbf{$+$11.9}}{{\tiny[5.0, 20.3]}} & \stk{\textbf{$+$13.0}}{{\tiny[8.2, 17.8]}} & \stk{$+$7.1}{{\tiny[$-$0.9, 14.9]}} \\
\newmonS{} $-$ LP-FCN & \stk{\textbf{$+$9.3}}{{\tiny[4.0, 14.7]}} & \stk{\textbf{$+$30.5}}{{\tiny[18.0, 41.8]}} & \stk{\textbf{$+$13.7}}{{\tiny[9.3, 18.3]}} & \stk{\textbf{$+$30.7}}{{\tiny[22.6, 37.8]}} & \stk{\textbf{$+$9.2}}{{\tiny[6.0, 12.7]}} & \stk{\textbf{$+$19.4}}{{\tiny[13.8, 25.5]}} & \stk{\textbf{$+$10.8}}{{\tiny[7.7, 14.3]}} & \stk{\textbf{$+$29.8}}{{\tiny[22.0, 37.9]}} \\
\newmonS{} $-$ GMM & \stk{\textbf{$+$14.3}}{{\tiny[9.7, 19.3]}} & \stk{\textbf{$+$44.5}}{{\tiny[32.7, 54.7]}} & \stk{\textbf{$+$19.8}}{{\tiny[15.0, 24.8]}} & \stk{\textbf{$+$31.6}}{{\tiny[23.4, 38.1]}} & \stk{\textbf{$+$16.8}}{{\tiny[11.0, 22.4]}} & \stk{\textbf{$+$41.4}}{{\tiny[34.2, 48.7]}} & \stk{\textbf{$+$14.6}}{{\tiny[10.2, 18.7]}} & \stk{\textbf{$+$37.9}}{{\tiny[29.8, 45.4]}} \\
\newmonS{} $-$ DE-RBD & \stk{\textbf{$+$8.7}}{{\tiny[5.2, 12.3]}} & \stk{\textbf{$+$19.1}}{{\tiny[8.1, 29.4]}} & \stk{\textbf{$+$20.7}}{{\tiny[15.7, 25.8]}} & \stk{\textbf{$+$20.5}}{{\tiny[12.9, 27.9]}} & -- & -- & -- & -- \\
\newmonS{} $-$ MC Dropout & \stk{\textbf{$+$26.3}}{{\tiny[19.3, 32.8]}} & \stk{\textbf{$+$50.2}}{{\tiny[37.0, 61.4]}} & -- & -- & -- & -- & -- & -- \\
\midrule
\tokmonS{} $-$ RBD & \stk{$+$3.8}{{\tiny[$-$2.6, 10.0]}} & \stk{$-$0.6}{{\tiny[$-$19.0, 11.9]}} & \stk{\textbf{$+$13.4}}{{\tiny[9.4, 17.7]}} & \stk{$+$2.4}{{\tiny[$-$4.7, 9.3]}} & \stk{\textbf{$+$10.4}}{{\tiny[5.3, 15.5]}} & \stk{$+$1.9}{{\tiny[$-$6.6, 10.5]}} & \stk{\textbf{$+$9.8}}{{\tiny[5.2, 14.5]}} & \stk{$-$1.6}{{\tiny[$-$10.0, 6.6]}} \\
\tokmonS{} $-$ LP-FCN & \stk{$+$1.4}{{\tiny[$-$2.2, 4.8]}} & \stk{$+$4.7}{{\tiny[$-$4.8, 13.3]}} & \stk{\textbf{$+$8.0}}{{\tiny[3.6, 12.9]}} & \stk{\textbf{$+$20.0}}{{\tiny[10.4, 28.3]}} & \stk{\textbf{$+$6.2}}{{\tiny[3.1, 9.7]}} & \stk{\textbf{$+$9.5}}{{\tiny[3.9, 15.6]}} & \stk{\textbf{$+$7.6}}{{\tiny[4.7, 10.9]}} & \stk{\textbf{$+$21.2}}{{\tiny[13.0, 28.7]}} \\
\tokmonS{} $-$ GMM & \stk{\textbf{$+$6.4}}{{\tiny[0.1, 13.2]}} & \stk{\textbf{$+$18.6}}{{\tiny[9.4, 27.5]}} & \stk{\textbf{$+$14.1}}{{\tiny[8.8, 19.7]}} & \stk{\textbf{$+$20.9}}{{\tiny[12.5, 27.5]}} & \stk{\textbf{$+$13.9}}{{\tiny[7.8, 19.8]}} & \stk{\textbf{$+$31.5}}{{\tiny[22.7, 39.3]}} & \stk{\textbf{$+$11.5}}{{\tiny[7.1, 15.7]}} & \stk{\textbf{$+$29.2}}{{\tiny[21.4, 36.2]}} \\
\tokmonS{} $-$ DE-RBD & \stk{$+$0.9}{{\tiny[$-$5.3, 7.5]}} & \stk{$-$6.8}{{\tiny[$-$19.9, 4.4]}} & \stk{\textbf{$+$15.0}}{{\tiny[10.5, 20.0]}} & \stk{\textbf{$+$9.8}}{{\tiny[1.6, 18.5]}} & -- & -- & -- & -- \\
\tokmonS{} $-$ MC Dropout & \stk{\textbf{$+$18.4}}{{\tiny[12.5, 24.1]}} & \stk{\textbf{$+$24.3}}{{\tiny[14.6, 34.0]}} & -- & -- & -- & -- & -- & -- \\
\bottomrule
\end{tabular}}
\caption{\changeTwo{Paired differences between collision predictors on nuScenes (percentage points), with $95\%$ CIs from the same 2000 scene-bootstrap resamples. Bold: significant (the interval excludes zero). RM-T/RM-G: \tokmon{}/\newmon{}.}}
\label{tab:paired}
\end{table*}

\begin{figure*}[t]
\centering
\includegraphics[width=\linewidth]{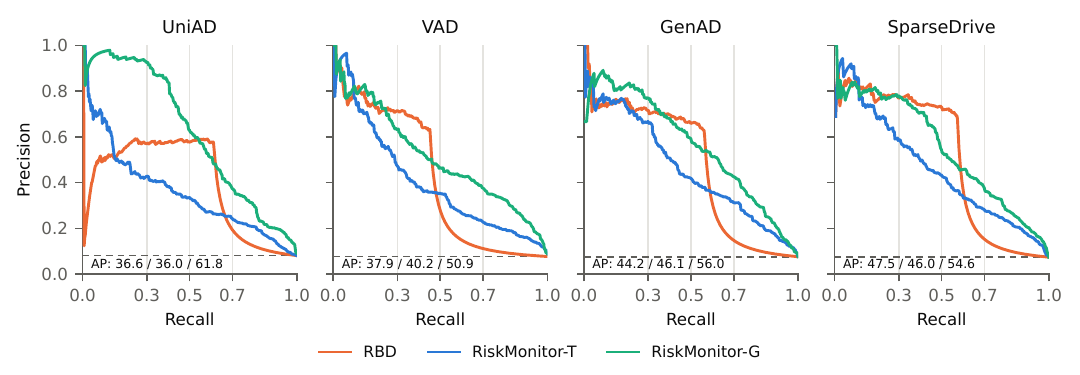}
\caption{\changeTwo{Precision--recall curves of RBD, \tokmon{} and \newmon{} on nuScenes (validation split). Precision is the expected precision with tied scores in random order, as for $\mathrm{Pr}_X$ (Sec.~\ref{sec:metrics}). AP (\%): RBD / \tokmon{} / \newmon{}. Dashed line: positive rate.}}
\label{fig:pr}
\end{figure*}

\subsection{Evaluation Metrics}
\label{sec:metrics}
Since we treat CRE as a binary classification problem, we use area under the ROC curve (AUROC) and average precision (AP) to evaluate \replaceTwo{the performance of RiskMonitor}{the collision predictors}.
\replaceTwo{Since AP is known to be more sensitive to data imbalance, it is more suitable for real-world evaluation in Sec.~\ref{sec:realworld}.}{AP is more sensitive to data imbalance and thus better suited to the real-world evaluation in Sec.~\ref{sec:realworld}.}
We also report precision at recall levels of $30\%$, $50\%$, and $70\%$, which indicate how many false positives (FP) we must accept to find a certain percentage of collisions\changeTwo{; $\mathrm{Pr}_{100}$ is omitted, since it is close to the positive rate for every method}.
\changeTwo{Some predictors assign the same score to many samples; RBD, for instance, gives a score of zero to all samples in which no predicted agent overlaps the ego plan.
A recall level can then fall inside a large block of tied samples, where the precision depends on the order in which these samples are flagged.
We therefore report the \textbf{expected precision \boldmath$\mathrm{Pr}_X$}: the samples are flagged in order of decreasing score, samples of equal score in random order, until exactly $XP/100$ of the $P$ positives are found.
If the cut falls into a block of $n_b$ tied samples containing $p_b$ positives, preceded by $F$ flagged samples with $T$ positives, then
\begin{equation}
\mathrm{Pr}_X = \frac{XP/100}{F + (XP/100 - T)\, n_b/p_b}\enspace.
\end{equation}
Without ties at the cut ($n_b=p_b=1$), $\mathrm{Pr}_X$ reduces to the ordinary precision at recall $X\%$.

\parsection{Confidence intervals}
Frames of the same scene are correlated, so all metrics are reported with $95\%$ confidence intervals (CIs) from a bootstrap over scenes, in which all frames of a scene are resampled together (2000 resamples).
On nuScenes, the bootstrap is over the 150 validation scenes; on NeuroNCAP, it is over the 14 scenes, each with all its runs and scenario types, and the scores of the five test folds are pooled (Sec.~\ref{sec:simulation}).
Comparisons between two methods use the paired difference computed on the same resamples, and we call a difference significant only when its interval excludes zero (Appendix~\ref{sec:bootstrap}).
}

For the closed-loop evaluation, the collision rate \citep{ljungbergh2024neuroncap} is used to \replaceTwo{reflect RiskMonitor's collision avoidance ability integrated with end-to-end planners}{measure whether the predictions of the monitor, when used to trigger a braking policy (Sec.~\ref{sec:simulation}), reduce the collisions of the end-to-end planner}.

\subsection{Open-Loop Classification}
\label{sec:realworld}
\replaceTwo{Firstly, our RiskMonitor is evaluated on the realworld benchmark, nuScenes.
UniAD\changeTwo{, VAD, GenAD and SparseDrive are all} trained and evaluated on the real-world dataset nuScenes.}{We first evaluate the monitors on the real-world benchmark nuScenes, on which UniAD, VAD, GenAD and SparseDrive are trained.}
\replaceTwo{Note that nuScenes training and validation splits contain different sequences.
We cache the~$h_\plan$,~$h_\motion$, and~$L_\collision$ of \changeTwo{each planner} on each split to build the training and test sets for RiskMonitor.
\changeTwo{The nuScenes validation split serves as the test set.}}{The monitors are trained on the training split and tested on the validation split, which contains different sequences.}
\replaceTwo{In UniAD, various levels of safety distance are used for the collision loss, meaning~$L_\collision > 0$ if a road user's distance to the ego-vehicle is within the safety distance.
In RiskMonitor, we use ~$L_\collision$ with 1 meter as the safety distance.
There is no such setting in \citep{jiang2023vad}, so we reimplement UniAD's collision loss in VAD and use the same safety distance for both models.}{}

\changeTwo{
\parsection{Setup}
All nuScenes results are produced with a single protocol for all four planners and all methods.
Tokens, decoded outputs and labels are exported from the released checkpoints of UniAD, VAD-Base, GenAD and SparseDrive with each planner's official inference pipeline; we reproduce the official planning L2 errors of all four planners within $0.03$\,m (Appendix~\ref{sec:repro}).
For collision labeling we use the UniAD collision loss, which is positive when a road user comes within the safety distance of the ego vehicle, here $1.0$\,m; it uses axis-aligned boxes and all ground-truth objects and is applied to the plan before any test-time collision optimization (UniAD) or collision-aware re-scoring (SparseDrive).
The positive rates on the validation split are $8.09\%$ for UniAD, $7.69\%$ for VAD, $7.59\%$ for GenAD and $7.39\%$ for SparseDrive.
All learned monitors are trained on all 700 training scenes with the bagging ensemble of Sec.~\ref{sec:imbalance}, for a fixed number of 200 epochs (Appendix~\ref{sec:train-detail}); nothing is selected on the validation split, which is evaluated once.
}
\replaceTwo{The bagging ensemble described in Sec.~\ref{sec:imbalance} is used in training to help alleviate this imbalance.
See more details on the training in the Sec.~\ref{sec:exp-A}.}{}

\subsubsection{\changeTwo{Comparison with Baselines}}
\label{sec:comparison}
\replaceTwo{\parsection{Results}}{}
\changeTwo{Tab.~\ref{tab:nuscenes} compares the collision predictors on the four planners, with $95\%$ CIs, and Tab.~\ref{tab:paired} gives the paired differences on which our conclusions rest.
Since the inputs (tokens and decoded trajectories) come from each planner and the label depends on its own plan, the inputs, the positive samples and the positive rates differ between planners; we therefore compare methods only within a planner.}

\changeTwo{\parsection{Baselines}
RBD is a strong baseline in AP.
It gives a non-zero score only when a predicted agent box overlaps the planned ego box; these samples are mostly collisions ($\mathrm{Pr}_{30}$ of $57.03\%$--$76.95\%$), but they cover only $45\%$--$61\%$ of the collisions.
Beyond this recall, all remaining samples share a score of zero, and the precision of RBD falls toward the positive rate (Fig.~\ref{fig:pr}), to $11.14\%$--$19.27\%$ at $70\%$ recall.
GMM, which integrates the predicted multimodal motion over the ego footprint, has an AUROC of $70.90\%$--$76.18\%$, close to that of RBD, but its AP of $14.63\%$--$19.36\%$ is around half that of RBD.
The generic UQ scores are weaker: on UniAD, MC Dropout reaches $63.98\%$ AUROC and $11.60\%$ AP, and among the Deep Ensemble scores only DE-RBD is competitive (Appendix~\ref{sec:DE}); it requires five planners and the hand-written collision rule.}
\replaceTwo{Fig.~D3 visualizes examples of false-positive prediction from the rule-based discriminator based on UniAD prediction. 
The reasons for false-positive predictions include accumulated error in object detection, noisy ego-vehicle plan, and incorrect motion predictions.
Fig.~D3 (a) and (b) show accumulated error from object detection.
The front camera view shows that the ego-vehicle is driving straight forward, and there is no road user coming towards it. 
However, the BEV says that the road user (red!80!black boxes) charges straight and collides head-on.
Fig.~D3 (d) shows wrong motion predictions.
Front camera views indicate the white car on the left is driving in the same direction as the ego-vehicle; however, the corresponding blue boxes indicate the opposite.
Fig.~D3 (e) shows a noisy ego plan.
Front left camera views suggest the ego-vehicle is moving slowly, while the ego plan becomes noisy in BEV (red boxes).}{}

\changeTwo{\parsection{\tokmon{}}
\tokmon{} ranks samples better than RBD on VAD, GenAD and SparseDrive (AUROC $+9.8$ to $+13.4$ points, significant), but not on UniAD ($+3.8$ [$-2.6$, $10.0$]), and its AP is not significantly different from that of RBD on any of the four planners.
The two reach a similar AP in different ways: RBD is more precise up to the end of its non-zero scores, and \tokmon{} beyond it (Fig.~\ref{fig:pr}).
Compared with LP-FCN, which reads only the planning token, \tokmon{} is significantly better in AUROC and AP on VAD, GenAD and SparseDrive, but not on UniAD (Tab.~\ref{tab:paired}).
The advantage of the cross-attention over the planning token alone is therefore significant on three planners.
On UniAD, \tokmon{} is also not significantly different from DE-RBD, while on VAD it is significantly better.
\tokmon{} is significantly better than GMM and MC Dropout on every planner where they are evaluated.

\begin{figure*}[t]\centering
\includegraphics[width=\linewidth]{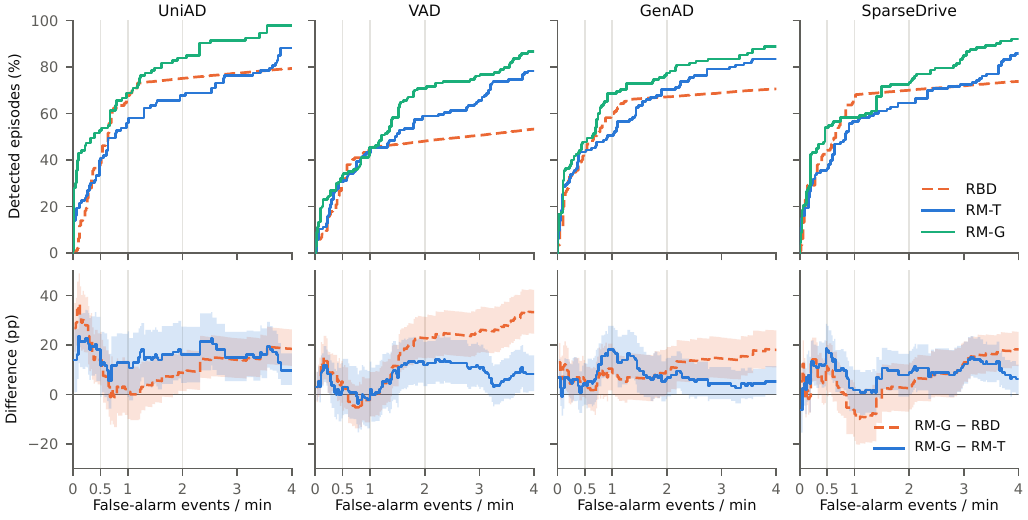}
\caption{\changeTwo{Event-level detection on nuScenes (validation split). Top: share of detected collision episodes against the rate of false-alarm events, sweeping the threshold of each method; the x-axis is the highest rate reached down to the threshold (Sec.~\ref{sec:event}). Bottom: paired differences RM-G $-$ RBD and RM-G $-$ RM-T in detected episodes (percentage points), each method at its operating point for the rate on the x-axis; shaded: pointwise $95\%$ CIs from the scene bootstrap. Vertical lines: the operating points of Tab.~\ref{tab:event} (Appendix~\ref{sec:event-A}). RM-G: \newmon{}; RM-T: \tokmon{}.}}
\label{fig:froc}
\end{figure*}

\parsection{\newmon}
Adding the explicit per-agent geometry to the tokens improves both AUROC and AP.
\newmon{} reaches $90.27\%$--$90.78\%$ AUROC and $50.95\%$--$61.79\%$ AP on the four planners.
It is significantly better than \tokmon{} on all four planners in both AUROC and AP, and better than RBD in AUROC on all four planners and in AP on all but SparseDrive (Tab.~\ref{tab:paired}).
At low recall, the point estimates of RBD are sometimes higher ($\mathrm{Pr}_{30}$ on VAD, $\mathrm{Pr}_{50}$ on GenAD and SparseDrive), but none of these differences is significant, whereas at $70\%$ recall \newmon{} is significantly more precise on VAD, GenAD and SparseDrive.
In Fig.~\ref{fig:pr}, \newmon{} lies above \tokmon{} over almost the whole recall range; on VAD, GenAD and SparseDrive, RBD is more precise in the middle of the range, up to the end of its non-zero scores, after which \newmon{} is clearly more precise.
\newmon{} is significantly better than LP-FCN, GMM, DE-RBD and MC Dropout on every planner where they are evaluated, with a single forward pass of one planner.
Each monitor is an ensemble of four bagging members, trained with different random initializations on different subsets of the negatives, so their spread reflects the variability of training.
On UniAD, the members of \newmon{} reach $88.7\%$--$90.1\%$ AUROC and $47.6\%$--$57.4\%$ AP, and those of \tokmon{} $78.6\%$--$81.6\%$ AUROC and $30.2\%$--$33.3\%$ AP, so the ranking does not depend on a particular training run.}

\changeTwo{\parsection{Objects missed by the planner}
On the four planners, $10.6\%$--$20.7\%$ of the collisions involve an object that the planner did not detect (case~1; Appendix~\ref{sec:reliance}).
On these collisions, \tokmon{} reaches a per-case AUROC of $81.8\%$--$89.3\%$, similar to that on collisions with detected objects ($82.7\%$--$88.6\%$), but at a fixed false-positive rate of $10\%$ it detects fewer of them on all four planners, most clearly on UniAD.
Its performance is therefore affected by, but not strictly bounded by, the perception of the planner.}

\subsubsection{Ablation}
\label{sec:ablation}

\begin{table*}[t]
\centering
\scriptsize
\setlength\tabcolsep{1.3mm}
\resizebox{\textwidth}{!}{\color{roundtwo}
\begin{tabular}{@{}lcccccccccc@{}}
\toprule
Variant & Agents & Geometry & Layers & Agent head & Mixup & AUROC & AP & $\mathrm{Pr}_{30}$ & $\mathrm{Pr}_{50}$ & $\mathrm{Pr}_{70}$ \\ \midrule
\tokmonS{} & vehicles & -- & 1 & -- & $\checkmark$ & \stk{82.39}{{\scriptsize[77.7, 86.4]}} & \stk{35.95}{{\scriptsize[25.1, 46.6]}} & 42.76 & 33.51 & 23.94 \\
(a) & all & -- & 1 & -- & -- & \stk{82.80}{{\scriptsize[78.6, 86.5]}} & \stk{34.28}{{\scriptsize[24.7, 43.9]}} & 41.05 & 33.51 & 22.23 \\
(b) & all & $\checkmark$ & 1 & -- & -- & \stk{88.61}{{\scriptsize[84.4, 92.5]}} & \stk{53.11}{{\scriptsize[41.6, 63.5]}} & 73.82 & 51.51 & 36.14 \\
(c) & all & $\checkmark$ & 3 & -- & -- & \stk{88.42}{{\scriptsize[84.4, 92.1]}} & \stk{49.06}{{\scriptsize[37.6, 59.5]}} & 69.72 & 47.12 & 30.21 \\
(d) \newmonS{} & all & $\checkmark$ & 1 & $\checkmark$ & -- & \stk{\textbf{90.27}}{{\scriptsize[85.9, 94.0]}} & \stk{\textbf{61.79}}{{\scriptsize[48.8, 72.9]}} & \textbf{92.61} & 63.09 & \textbf{37.22} \\
(e) & all & $\checkmark$\,+\,overlap & 1 & $\checkmark$ & -- & \stk{90.14}{{\scriptsize[86.2, 93.6]}} & \stk{60.63}{{\scriptsize[48.0, 71.1]}} & 87.58 & \textbf{65.73} & 35.27 \\
(f) & all & $\checkmark$ & 1 & $\checkmark$ & $\checkmark$ & \stk{83.99}{{\scriptsize[79.5, 88.0]}} & \stk{39.19}{{\scriptsize[29.3, 49.2]}} & 44.80 & 38.37 & 23.62 \\
\bottomrule
\end{tabular}}
\caption{\changeTwo{Ablation of \newmon{} on UniAD, nuScenes validation split (\%), with $95\%$ scene-bootstrap CIs. Variants (a)--(f) are described in Sec.~\ref{sec:ablation}. RM-T/RM-G: \tokmon{}/\newmon{}. Agents: vehicle queries only, or all tracked agents with their class. Geometry: per-agent relative geometry features $g_a$ (Sec.~\ref{sec:newmon}); overlap: the per-step overlap of the agent and ego boxes as additional features.}}
\label{tab:ablation}
\end{table*}

\changeTwo{
We ablate the components of \newmon{} on UniAD, under the protocol of Tab.~\ref{tab:nuscenes}.
The design of \newmon{} was chosen on a held-out $30\%$ of the UniAD training scenes before the validation split was read; the validation results below are reported for reference and were not used for any choice.
Tab.~\ref{tab:ablation} adds the components one at a time, and the paired differences below have $95\%$ scene-bootstrap CIs.

\parsection{Agents and mixup (a)}
Reading all tracked agents with their class and training without mixup does not change \tokmon{} significantly (AUROC $+0.4$ [$-2.7$, $3.8$], AP $-1.7$ [$-8.1$, $4.9$]).

\parsection{Geometry (b)}
The explicit per-agent geometry accounts for most of the gain: compared with \tokmon{}, AUROC $+6.2$ [$2.0$, $10.9$] and AP $+17.2$ [$7.0$, $28.0$] points.

\parsection{Depth (c)}
Using three decoder layers instead of one does not improve on (b) (point estimate $-4.1$ AP) and is significantly worse than \newmon{} in AP ($-12.7$ [$-19.1$, $-5.7$]).

\parsection{Per-agent head (d)}
The per-agent risk head adds $+1.7$ [$0.1$, $3.3$] AUROC and $+8.7$ [$2.9$, $14.1$] AP points over (b), which gives \newmon.

\parsection{Explicit overlap (e)}
Adding the per-step overlap of the agent and ego boxes, i.e., the quantity tested by the label rule, does not help (AUROC $-0.1$ [$-1.2$, $0.9$], AP $-1.2$ [$-4.5$, $2.1$] relative to \newmon): the network derives the overlap from the raw relative coordinates.

\parsection{Mixup (f)}
Training \newmon{} with the mixup of \tokmon{} ($\alpha_m=3$, Appendix~\ref{sec:train-detail}) removes most of the gain (AUROC $-6.3$ [$-10.2$, $-2.3$], AP $-22.6$ [$-32.2$, $-12.9$]), and the result is not significantly different from \tokmon{}.
A plausible reason is that mixup interpolates the geometry features of two unrelated samples, so that the mixed features no longer describe the relation between an agent and the ego plan on which the label depends.

}

\changeTwo{
\subsubsection{Event-Level False Alarms}
\label{sec:event}
The metrics of Sec.~\ref{sec:metrics} score each keyframe independently.
However, the three-second prediction windows of consecutive keyframes overlap, so one hazardous situation produces several consecutive positive keyframes, and one false-alarm situation produces several consecutive alarms.
In operation, what matters is how often a monitor raises a false alarm per unit of driving time, and whether it warns before each hazardous situation.
We therefore measure false alarms at the level of events.

All keyframes of a scene are ordered in time.
A predicted \emph{alarm event} is a run of consecutively predicted positive keyframes in a scene; it is \emph{false} when no ground-truth positive keyframe occurs during it or within the following $3$\,s, the horizon of planners.
A ground-truth \emph{collision episode} is a run of consecutive ground-truth positive keyframes; it is detected when an alarm occurs between $3$\,s before its start and its end.
We compare the monitors at equal false-alarm rates: for each monitor, the threshold is set so that the rate of false-alarm events is $F = 0.5$, $1$ or $2$ per minute over the validation split (one every 120, 60 or 30\,s), and we report the share of detected collision episodes.
This is the event-level counterpart of the recall at a fixed false-positive rate; Appendix~\ref{sec:event-A} details the choice of thresholds, the handling of tied scores and the CIs.

Fig.~\ref{fig:froc} reports the results, and Tab.~\ref{tab:event} lists the values at the three operating points.
At equal false-alarm rates, \newmon{} detects the largest share of collision episodes at 11 of the 12 operating points (at one tied with RBD) and never significantly fewer than \tokmon{} or RBD; against each of the two, the difference is significant at four operating points.
RBD is competitive only up to about one false-alarm event per minute, where its non-zero scores are exhausted and its curve flattens.
Even for \newmon{}, a false-alarm event every 2\,min leaves about half of the collision episodes undetected ($33.6\%$--$54.0\%$ detected), and detecting $70.9\%$--$83.9\%$ requires a false-alarm event every 30\,s.
These false-alarm rates are of the same order as the rate of collision episodes itself: the plans of the four planners enter a new collision episode every 26--45\,s.
Rates relevant to deployment, such as one false alarm per hour, cannot be measured on the 50 minutes of nuScenes validation driving and require much longer driving logs; these open-loop numbers therefore do not establish deployability, and we report them as a reference for future monitors.
The evaluation is also limited to 20-second scenes, so events cannot span scenes, and the 67--115 collision episodes per planner make the CIs wide.

}

\subsection{Closed-Loop Validation}
\label{sec:simulation}
We evaluate our RiskMonitor on NeuroNCAP \citep{ljungbergh2024neuroncap}, which is a closed-loop neural simulator containing safety-critical scenarios.
\replaceTwo{We first test the closed-loop classification performance as in the open-loop case, then validate its collision-avoidance ability by thresholding on the risk.}{We first evaluate classification as on nuScenes, then use the alarms of the monitors to trigger a downstream braking policy.}

\subsubsection{Closed-Loop Classification}
\begin{figure}[!htbp]

    \centering
    \begin{subfigure}{1.0\linewidth} 
        \centering
        \begin{minipage}[c]{0.12\linewidth}
            \centering \tiny
            \rotatebox{90}{Prediction} \\[1.2cm]
            \rotatebox{90}{Simulation results}
        \end{minipage}
        \begin{minipage}[c]{0.85\linewidth}
            \includegraphics[width=\linewidth, bb=0 0 1224 1103.04]{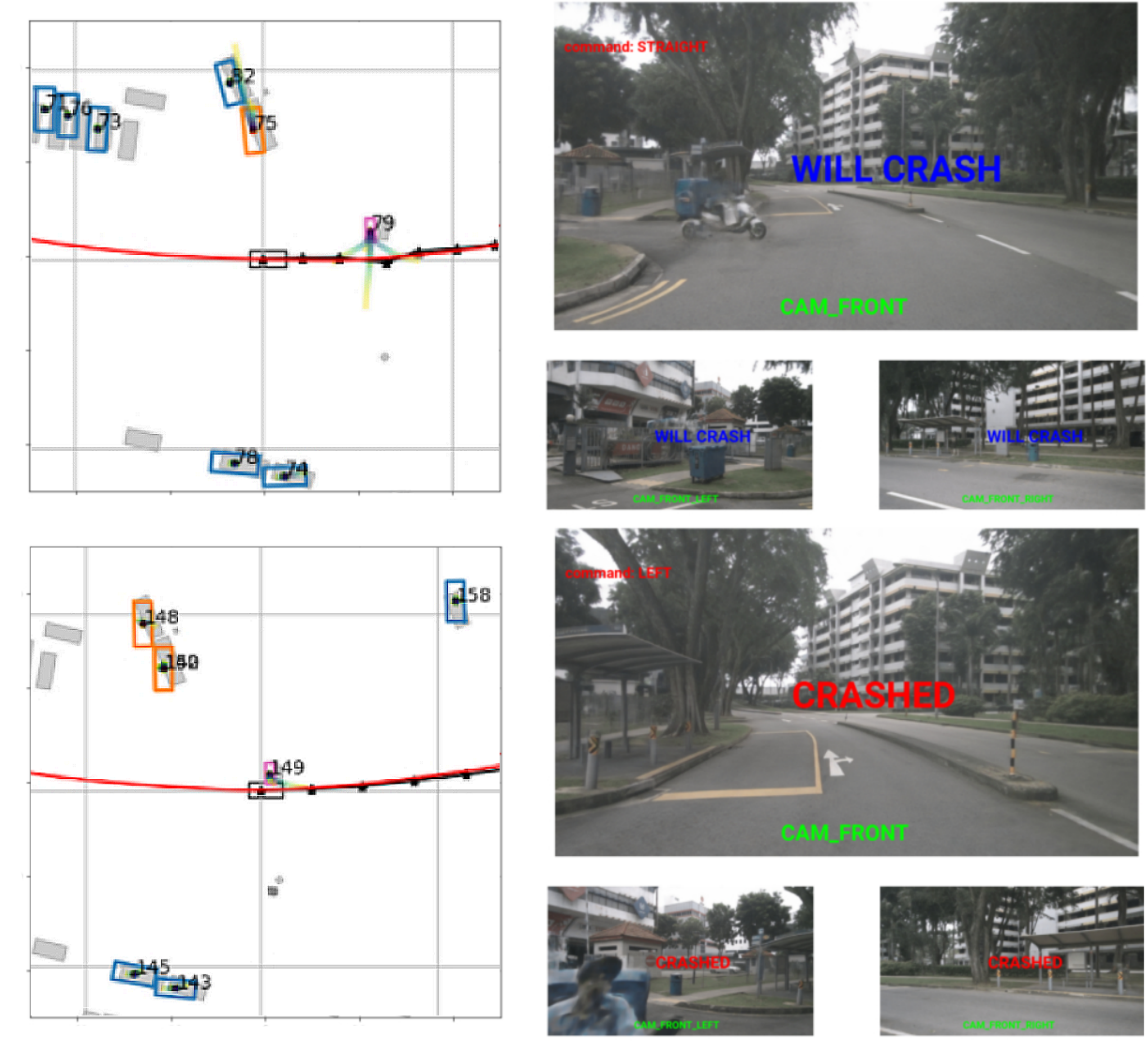}
        \end{minipage}
        \caption{Example of true positive prediction. Top: \tokmon{} predicts``will crash" at time $t$; Bottom: Simulation result shows ``crash" after a few steps.}
        \label{fig:tp1}
    \end{subfigure}
    \begin{subfigure}{1.0\linewidth}  
        \centering
        \begin{minipage}[c]{0.12\linewidth}
            \centering \tiny
            \rotatebox{90}{Prediction} \\[1.2cm]
            \rotatebox{90}{Simulation results}
        \end{minipage}
        \begin{minipage}[c]{0.85\linewidth}
            \includegraphics[width=\linewidth]{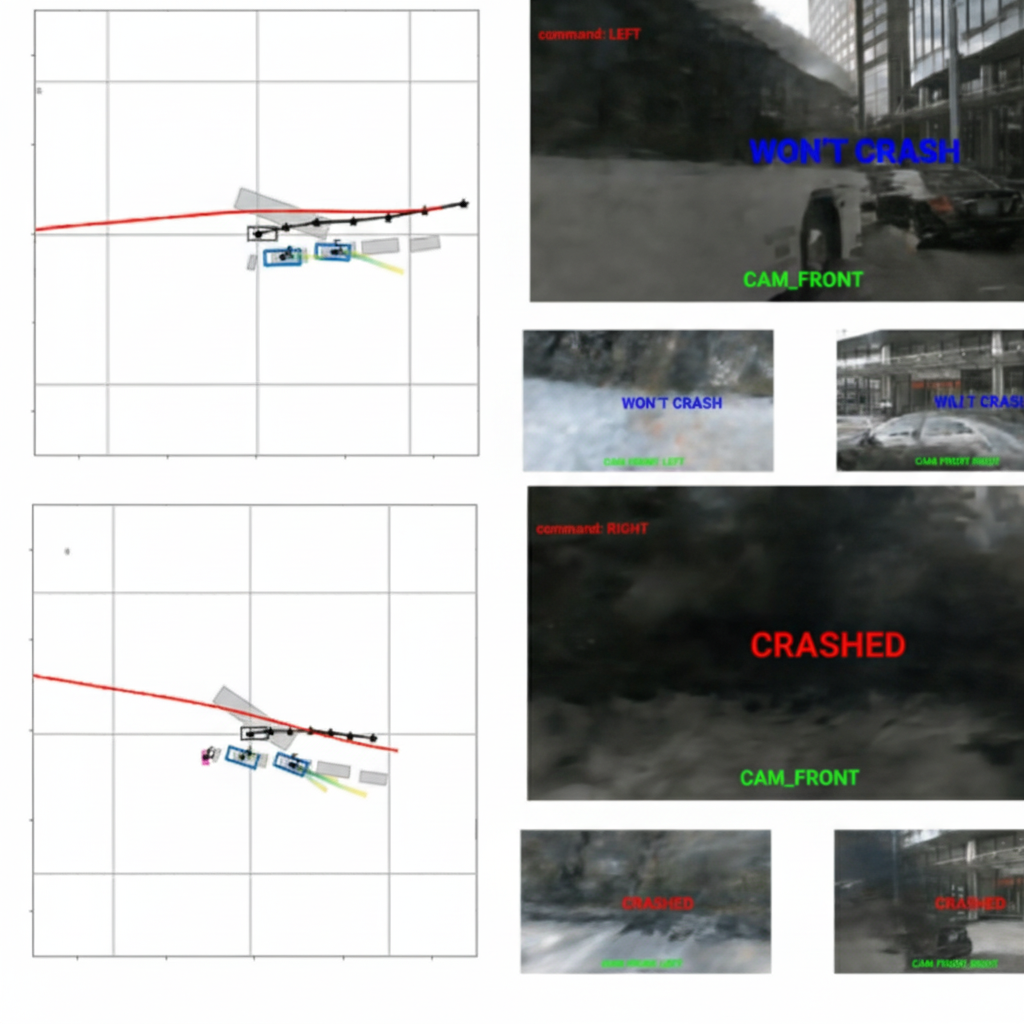}
        \end{minipage}
        \caption{Example of false negative prediction. Top: \tokmon{} predicts ``won't crash" at time $t$; Bottom: Simulation result shows ``crash" after a few steps.}
        \label{fig:fn1}
    \end{subfigure}
    
    \caption{Qualitative Examples on NeuroNCAP Simulation with UniAD as the end-to-end planner. The left figure presents ground-truth objects (gray) alongside predicted objects (color-coded by category) and their forecasted trajectories (blue-to-yellow gradient line) in BEV. It also depicts the ego-vehicle (black), the planned trajectory (black), and the reference trajectory (red), indicating its intended turn. The right figures are the inputs of the 3 front cameras. Best viewed in color.}
    \label{fig:quantitative-examples-ncap}
    
\end{figure}

\replaceTwo{We use the same settings as in NeuroNCAP, where UniAD and VAD are only pretrained on nuScenes, to cache tokens in the end-to-end AD planners.}{As in NeuroNCAP, UniAD is trained on nuScenes only, and we cache its tokens in the simulator.}
\replaceTwo{We split them into training, validation, and test sets with a 6:1:3 ratio.
The fraction of collision samples in the test set is~$46\%$ (see Sec.~\ref{sec:exp-A} for details).
Each set contains all three types of scenarios but strictly avoids using the same sequence in both sets to reduce the risk that the model learns the sequences rather than the task.}{The planner runs with its test-time collision optimization, and the tokens and decoded predictions are recorded at every planning step.
The same labeling as on nuScenes is used: the collision loss of the plan before the collision optimization (Appendix~\ref{sec:cache}), computed against the future boxes of the simulated actors.
The dataset of UniAD contains $2439$ runs over the 20 scene-sequence pairs, with $28025$ samples, $50.4\%$ of them positive (Appendix~\ref{sec:rationale}).
The 20 pairs share 14 nuScenes scenes; scene 0103, for example, appears in all three scenario types.
We therefore evaluate by five-fold cross-validation over scenes: each fold holds out all runs and scenario types of its scenes as the test set, a second fold is the validation set, and the remaining three folds are the training set.
Every sample is thus scored once, by monitors that have not seen its scene.
}
\replaceTwo{The models are trained for 20 epochs with a learning rate set as 0.001, and the results are shown in Tab.~\ref{tab:neuroncap}. For training \tokmon{} and LP-FCN, we use focal loss with  $\alpha=0.5$ and $\gamma=2$.}{\tokmon{}, \newmon{} and LP-FCN are trained on the training folds as described in Appendix~\ref{sec:train-detail}. RBD, GMM and MC Dropout are computed as on nuScenes.}\replaceTwo{ The qualitative examples of \tokmon{} are shown in Fig.~\ref{fig:quantitative-examples-ncap}.}{}

\begin{table*}[t]
\centering
\scriptsize
\setlength\tabcolsep{1.2mm}
{\color{roundtwo}
\begin{tabular}{@{}lccccc@{}}
\toprule
Method & AUROC & AP & $\mathrm{Pr}_{30}$ & $\mathrm{Pr}_{50}$ & $\mathrm{Pr}_{70}$ \\ \midrule
MC Dropout & 50.35\,{\tiny[40.5, 62.6]} & 50.46\,{\tiny[36.7, 70.7]} & 50.40\,{\tiny[35.4, 73.5]} & 50.82\,{\tiny[37.1, 70.6]} & 50.80\,{\tiny[37.9, 68.4]} \\
GMM & 78.24\,{\tiny[72.2, 86.2]} & \underline{77.69}\,{\tiny[72.9, 87.5]} & \underline{85.37}\,{\tiny[78.5, 93.8]} & \underline{83.74}\,{\tiny[76.4, 92.4]} & 72.49\,{\tiny[66.7, 89.1]} \\
RBD & \underline{79.80}\,{\tiny[69.9, 88.6]} & 72.25\,{\tiny[59.4, 88.7]} & 77.03\,{\tiny[58.8, 96.9]} & 81.13\,{\tiny[66.7, 96.7]} & \underline{82.73}\,{\tiny[70.4, 94.7]} \\
LP-FCN & 64.05\,{\tiny[41.1, 79.2]} & 59.58\,{\tiny[40.9, 79.4]} & 57.35\,{\tiny[35.1, 84.8]} & 59.92\,{\tiny[39.4, 82.0]} & 58.49\,{\tiny[43.8, 76.0]} \\
\tokmonS{} & 77.57\,{\tiny[62.5, 87.7]} & 74.23\,{\tiny[57.7, 84.9]} & 73.47\,{\tiny[50.7, 90.2]} & 72.95\,{\tiny[54.2, 87.0]} & 70.87\,{\tiny[57.8, 83.2]} \\
\newmonS{} & \textbf{87.21}\,{\tiny[80.8, 92.9]} & \textbf{88.49}\,{\tiny[82.3, 93.8]} & \textbf{96.36}\,{\tiny[93.0, 99.3]} & \textbf{95.02}\,{\tiny[91.3, 97.8]} & \textbf{89.75}\,{\tiny[73.5, 94.1]} \\
\midrule
Difference & \multicolumn{5}{c}{} \\ \midrule
\newmonS{} $-$ \tokmonS{} & \textbf{$+$9.6}\,{\tiny[0.2, 24.5]} & \textbf{$+$14.3}\,{\tiny[4.7, 28.7]} & \textbf{$+$22.9}\,{\tiny[7.6, 42.8]} & \textbf{$+$22.1}\,{\tiny[7.6, 40.0]} & \textbf{$+$18.9}\,{\tiny[1.0, 33.9]} \\
\newmonS{} $-$ RBD & $+$7.4\,{\tiny[$-$0.9, 14.4]} & \textbf{$+$16.2}\,{\tiny[1.3, 26.5]} & $+$19.3\,{\tiny[$-$0.8, 38.3]} & $+$13.9\,{\tiny[$-$1.0, 25.3]} & $+$7.0\,{\tiny[$-$3.3, 10.6]} \\
\newmonS{} $-$ GMM & \textbf{$+$9.0}\,{\tiny[2.5, 13.5]} & \textbf{$+$10.8}\,{\tiny[4.2, 14.0]} & \textbf{$+$11.0}\,{\tiny[2.2, 19.1]} & \textbf{$+$11.3}\,{\tiny[3.0, 16.5]} & \textbf{$+$17.3}\,{\tiny[0.6, 23.9]} \\
\newmonS{} $-$ LP-FCN & \textbf{$+$23.2}\,{\tiny[7.6, 46.1]} & \textbf{$+$28.9}\,{\tiny[11.8, 45.6]} & \textbf{$+$39.0}\,{\tiny[13.6, 58.3]} & \textbf{$+$35.1}\,{\tiny[13.8, 54.5]} & \textbf{$+$31.3}\,{\tiny[11.0, 47.8]} \\
\newmonS{} $-$ MC Dropout & \textbf{$+$36.9}\,{\tiny[23.5, 47.7]} & \textbf{$+$38.0}\,{\tiny[21.1, 51.0]} & \textbf{$+$46.0}\,{\tiny[23.1, 61.4]} & \textbf{$+$44.2}\,{\tiny[25.8, 57.4]} & \textbf{$+$38.9}\,{\tiny[22.4, 50.5]} \\
\midrule
\tokmonS{} $-$ RBD & $-$2.2\,{\tiny[$-$22.2, 11.5]} & $+$2.0\,{\tiny[$-$23.6, 21.0]} & $-$3.6\,{\tiny[$-$39.6, 26.5]} & $-$8.2\,{\tiny[$-$36.9, 15.4]} & $-$11.9\,{\tiny[$-$33.1, 5.4]} \\
\tokmonS{} $-$ GMM & $-$0.7\,{\tiny[$-$13.5, 7.0]} & $-$3.5\,{\tiny[$-$20.1, 7.2]} & $-$11.9\,{\tiny[$-$38.6, 9.4]} & $-$10.8\,{\tiny[$-$33.2, 6.9]} & $-$1.6\,{\tiny[$-$16.8, 6.1]} \\
\tokmonS{} $-$ LP-FCN & \textbf{$+$13.5}\,{\tiny[4.4, 23.1]} & \textbf{$+$14.7}\,{\tiny[3.8, 20.8]} & \textbf{$+$16.1}\,{\tiny[3.3, 23.5]} & \textbf{$+$13.0}\,{\tiny[1.8, 21.2]} & \textbf{$+$12.4}\,{\tiny[5.0, 19.6]} \\
\tokmonS{} $-$ MC Dropout & \textbf{$+$27.2}\,{\tiny[3.2, 43.8]} & \textbf{$+$23.8}\,{\tiny[1.8, 42.3]} & $+$23.1\,{\tiny[$-$8.6, 50.7]} & $+$22.1\,{\tiny[$-$2.1, 44.3]} & \textbf{$+$20.1}\,{\tiny[2.9, 38.1]} \\
\bottomrule
\end{tabular}}
\caption{\changeTwo{Classification performance on the closed-loop benchmark NeuroNCAP with UniAD (\%), from five-fold cross-validation over the 14 scenes (2439 runs, 28025 samples, $50.4\%$ positive). Top: bold: highest, underlined: second-highest point estimate. Bottom: paired differences (percentage points); bold: significant (the interval excludes zero). Brackets: $95\%$ CIs from a bootstrap over the 14 scenes, in which all runs and scenario types of a scene are resampled together. RM-T/RM-G: \tokmon{}/\newmon{}.}
}
\label{tab:neuroncap}
\end{table*}

\replaceTwo{\parsection{Analysis}}{\parsection{Results}}
\replaceTwo{The results presented in Tab.~\ref{tab:neuroncap} demonstrate the effectiveness of \tokmon{} in identifying potential collisions across safety-critical scenarios.
GMM achieves an AUROC of 49.4\% and AP of 43.1\%, indicating that its performance is close to random guessing. This suggests that it is not sufficient to estimate collision risk from an object-level interface.
MC Dropout also performs close to a random guess.
It shows that the standard UQ baseline fails in end-to-end planners trained with a large amount of losses.}{
\newmon{} has the highest point estimate in every metric of Tab.~\ref{tab:neuroncap}.
It is significantly better than \tokmon{}, GMM, LP-FCN and MC Dropout in all five metrics, and better than RBD in AP ($+16.2$ [$1.3$, $26.5$]); its differences to RBD in AUROC ($+7.4$ [$-0.9$, $14.4$]) and in precision at fixed recall are not significant.
\tokmon{} is significantly better than LP-FCN and MC Dropout in AUROC and AP, but not significantly different from RBD and GMM.
MC Dropout is close to random guessing (AUROC $50.35\%$).
With 14 scenes, the CIs are wider than on nuScenes, so only large differences are significant; smaller ones, such as those between \newmon{} and RBD in AUROC, would require more scenes to resolve. The comparison between methods therefore rests mainly on nuScenes.
Since the dataset is balanced, AP and $\mathrm{Pr}_X$ are higher than on nuScenes and not comparable across benchmarks \changeTwo{(PR curves in Fig.~\ref{fig:pr-ncap}, Appendix~\ref{sec:ncap-pr})}; AUROC does not depend on the positive rate, and \newmon{} reaches a similar AUROC on both benchmarks with UniAD ($87.21\%$ here, $90.27\%$ on nuScenes).}

\replaceTwo{By contrast, the LP-based model significantly improves detection performance, suggesting that learning a direct mapping from~$h_{plan}$ token to collision likelihood is beneficial.
It is more efficient than MC Dropout, which requires several forward passes of the planners.
\tokmon{} outperforms all baselines, achieving an AUROC of 70.6\% and AP of 66.7\%, indicating its better capability to capture collision-critical factors.
This suggests that by cross-attending ego planning and road user dynamics, \tokmon{} improves collision detection for diverse simulated safety-critical scenarios.}{}

\parsection{Qualitative Results} \replaceTwo{We present qualitative results for true positive and false negative predictions in Fig.~\ref{fig:tp1} and Fig.~\ref{fig:fn1}, respectively.
On the left of Fig.~\ref{fig:fn1}, the BEV display reveals that the UniAD model fails to detect the bus (the huge grey rectangular without a blue box indicates the wrong detection result) in front of the ego vehicle.
Consequently, the motion token lacks information about the bus, leading \tokmon{} to misclassify the scenario.
This suggests that the accuracy of the perception modules in the AD model directly influences the predictive capability of \tokmon{}.

In contrast, in the True Positive (TP) example shown in Fig. \ref{fig:tp1}, UniAD provides more accurate coordinate estimations for a nearby object (indicated by the red!80!black box), allowing \tokmon{} to correctly predict the collision.
Interestingly, the plan from UniAD is still going straight even if there is a detected road user coming from the side. 
This highlights the necessity of having collision prediction when deploying end-to-end planners for AD.}{Fig.~\ref{fig:quantitative-examples-ncap} shows a true positive and a false negative of \tokmon{}; in the false negative, UniAD does not detect the bus in front of the ego vehicle (Appendix~\ref{sec:qual-ncap}).}
\replaceTwo{To further investigate the performance bound for RiskMonitor, we analyze different causes for collisions on nuScenes. See results in Appendix~\ref{sec:reliance}.}{}

\subsubsection{\replaceTwo{Collision Avoidance}{Monitor-Triggered Intervention}}
\replaceTwo{To further validate the impact of \tokmon{}, it is integrated with UniAD into NeuroNCAP pipeline to affect the control signal.
The controller is adapted so that it slams on the brake with the maximum deceleration if the predicted collision risk is above a predetermined threshold.
We keep the same settings and compare with baseline, and UniAD with or without test-time collision optimization in NeuroNCAP.}{To test whether the alarms of a monitor are useful for triggering downstream reactions, we attach \tokmon{}, \newmon{} and, as a reference, RBD to UniAD with its test-time collision optimization (U-opt) in the NeuroNCAP pipeline.
Each planning step that NeuroNCAP executes is scored by the monitor whose cross-validation fold holds out the scene, so the monitor has not seen the scene in training.
The threshold of each fold maximizes Youden's J statistic~\citep{youden1950index} on its validation fold.
From the first step whose risk reaches it, the planner outputs a stop, and NeuroNCAP's stopping controller brakes the car to a standstill (deceleration proportional to the speed, at most $6\,\mathrm{m/s^2}$), as for NeuroNCAP's own baseline; the brake stays engaged until the end of the run (Appendix~\ref{sec:brake-A}).
We follow the official setting of NeuroNCAP, with 50 runs for each of the 20 scene-sequence pairs, and compare with the baseline of NeuroNCAP (Base-U) and with UniAD without and with the collision optimization, as reported by \citet{ljungbergh2024neuroncap}, and with our re-run of U-opt with the same seeds.}
The baseline in NeuroNCAP utilizes the perception outputs from UniAD: \replaceTwo{a braking maneuver is applied when objects detected in a corridor in front of the ego-vehicle.}{a braking maneuver is applied when objects are detected in a corridor in front of the ego-vehicle.}
The collision rate is reported for \replaceTwo{three simulation scenarios}{the three scenario types}.

\parsection{Results} \replaceTwo{Tab.~\ref{tab:brake} shows that using our UniAD-\tokmon{} (U-R) significantly reduces collision rate for side and frontal sequences compared with the baseline while maintaining the rate on stationary scenes.}{Tab.~\ref{tab:brake} shows that braking triggered by each monitor lowers the collision rate of U-opt from $63.9\%$ in our re-run to $23.7\%$ with \newmon{}, $26.5\%$ with \tokmon{} and $33.3\%$ with RBD.
\newmon{} gives the lowest rate on stationary ($0.6\%$) and side ($14.8\%$) scenarios.
On frontal scenarios, where the actor approaches fast, \tokmon{} is lower ($43.6\%$) than \newmon{} ($55.6\%$), and RBD barely lowers the rate ($81.6\%$).}
\replaceTwo{As the baseline only considers road users running into a corridor in front of the ego-vehicle, it could ignore road users that are approaching from a side out of range.
In frontal scenarios, road users coming towards a high speed could be detected too late and thus there is no sufficient time to brake for the baseline.}{As the baseline only considers road users entering a corridor in front of the ego vehicle, it can miss road users approaching from the side.
In frontal scenarios, road users approaching at high speed can be detected too late for the baseline to brake.}
\replaceTwo{\tokmon{} takes the road user motion tokens as the input, thus it can detect in a further range and brake in advance.}{The monitors take the motion tokens and, for \newmon{}, the predicted trajectories of the road users as input, and may therefore raise an alarm before the actor enters the corridor.}
\replaceTwo{UniAD, with or without post-processing, improves over the baseline but still reaches high collision rates.
It shows that the planning module fails to adapt according to upstream detections.
Through this experiment, the necessity of an auxiliary collision monitor is demonstrated.}{UniAD, with or without its collision optimization, collides less often than the baseline in frontal and side scenarios but more often in stationary ones, and still reaches high collision rates.
Its planning module thus fails to adapt to the upstream detections, which motivates an auxiliary collision monitor.}

\begin{table}[t]
\centering
    \setlength\tabcolsep{1.2mm}
    \footnotesize
{\color{roundtwo}
\resizebox{\columnwidth}{!}{
\begin{tabular}{@{}lccc|cccc@{}}
\toprule
 & \multicolumn{3}{c|}{\citet{ljungbergh2024neuroncap}} & \multicolumn{4}{c}{Ours (seeds 0--49)} \\
Scene & Base-U & UniAD & U-opt & U-opt & $+$\tokmonS{} & $+$\newmonS{} & $+$RBD \\ \midrule
Stationary & 9.6 & 87.8 & 34.8 & 36.6 & 6.0 & \textbf{0.6} & 1.4 \\
Frontal & 100.0 & 98.4 & 92.4 & 85.2 & \textbf{43.6} & 55.6 & 81.6 \\
Side & 100.0 & 79.6 & 78.8 & 70.0 & 30.0 & \textbf{14.8} & 16.8 \\
Avg & 69.9 & 88.6 & 68.7 & 63.9 & 26.5 & \textbf{23.7} & 33.3 \\ \bottomrule
\end{tabular}}}
\caption{\replaceTwo{Closed-loop validation of \tokmon{}'s collision avoidance capability.}{Collision rates (\%; lower is better) on NeuroNCAP with braking triggered by a monitor. Base-U (NeuroNCAP's rule-based baseline), UniAD and U-opt (UniAD with its test-time collision optimization) are taken from \citet{ljungbergh2024neuroncap}. The other columns are our runs in the official setting (20 scene-sequence pairs $\times$ 50 runs, seeds 0--49): the re-run of U-opt, and U-opt with braking triggered by \tokmon{} (RM-T), \newmon{} (RM-G) or RBD (Appendix~\ref{sec:brake-A}). Avg: mean of the three scenario types; bold: lowest among our runs.}}
\label{tab:brake}
\end{table}

\section{Discussion, Limitations, and Future Work}
\label{sec:discussions}
{
\changeTwo{\citet{sun2025sparsedrive} assess the collision risk of each planning trajectory proposal at inference, but with motion prediction, similar to our GMM baseline.}
Instead, we have shown that collision risk can be estimated by applying loss prediction to frozen planners, \replaceTwo{while traditional, powerful UQ methods fail to do so}{whereas generic uncertainty scores, such as the endpoint variance under MC Dropout or a Deep Ensemble, rank collisions poorly}.
}\replaceTwo{Beyond applying loss prediction, we prove that ego-planning tokens and motion tokens contain the spatial-temporal information required for CRE. }{The comparison of the two monitors with the rule-based discriminator shows which planner outputs a collision monitor needs.
The planner's own decoded predictions carry much of the collision information: a rule applied to them reaches an AP that is not significantly different from that of \tokmon{}, which reads the tokens only.
The tokens and the explicit geometry are complementary, and \newmon{}, which combines them, is significantly better than both \tokmon{} and RBD in AUROC on all four planners.}{
\changeTwo{We have also shown in closed-loop simulation that the estimated risk can trigger a braking reaction that reduces collisions.}

\parsection{From Simulation to Real-world Applicability} \replaceTwo{There are around $7.7\%$ collisions in nuScenes for UniAD, while we make NeuroNCAP a balanced dataset.}{The NeuroNCAP dataset is balanced by construction (Appendix~\ref{sec:rationale}), and collisions are the majority outcome of NeuroNCAP's safety-critical scenarios, so the braking policy cannot be evaluated there under the rare-collision distribution of real driving.} 
}\replaceTwo{For UniAD, we can see that $Pr_{50}$ is $31.3\%$ on nuScenes, lower than $67.5\%$ on NeuroNCAP. However, the False Positive Rate (FPR) on nuScenes is $3.18\%$, lower than $7.55\%$ on NeuroNCAP too. Together with the braking policy, what actually matters for the driving smoothness is FPR. Lower FPR means more smoothness. }{On nuScenes, the monitors are evaluated at the natural collision prevalence of each planner ($7.4\%$--$8.1\%$).
False-alarm rates relevant to deployment require much longer driving logs than nuScenes provides (Sec.~\ref{sec:event}).
The braking policy is a proof of concept that shows that the monitor's output can trigger a downstream reaction; its effect on driving smoothness cannot be measured on NeuroNCAP, whose scenarios are all safety-critical, and we do not claim real-world deployability.}

\parsection{Limitations}
\replaceTwo{First, we reveal that there is a distribution shift of different driving video sequences (see Sec.\ \ref{sec:tsne}). This is the main reason why the test performance drops on the test set compared to on the validation set in Sec.~\ref{sec:ab-A}.}{The planner tokens carry sequence-specific structure (Appendix~\ref{sec:tsne}), so a monitor may generalize worse to unseen sequences than to the sequences seen in training. All results are therefore reported on sequences that do not occur in training; the monitors do not remove this shift. Whether more diverse training sequences narrow this gap remains open.}
{
}\changeTwo{Collisions with objects that the planner did not detect make up $10.6\%$--$20.7\%$ of the collisions on the four planners (Appendix~\ref{sec:reliance}).
Both monitors still rank most of them above normal driving, but at a false-positive rate of $10\%$ \tokmon{} detects fewer of them than of the collisions with detected objects on all four planners, so the planner's perception affects, without strictly bounding, the monitor's performance.
On UniAD, which passes only the agents it tracks to its downstream modules, both monitors detect few of these collisions at this false-positive rate.
The monitors are compatible with heterogeneous token-based planners, but attaching them to a new planner requires locating its tokens and decoded outputs, and determining the conventions of these outputs, in the planner's code (Appendix~\ref{sec:cache}).
On NeuroNCAP, closed-loop classification and braking cover UniAD only; VAD, SparseDrive and GenAD are evaluated open-loop only, and their closed-loop evaluation is left to future work.}{

\parsection{Future Work}
A more advanced policy, e.g., using MPC, \replaceTwo{has the potential to further improve safety}{could react more smoothly than braking to a standstill}. 
\changeTwo{A simple extension is to calibrate the predicted risk with temperature scaling \citep{guo2017calibration} and to decelerate less for less confident predictions.}
}
\section{Conclusion}
\replaceTwo{Ensuring that self-driving vehicles do not cause road collisions is of paramount importance.}{End-to-end planners still produce colliding plans, but most do not estimate the collision risk of these plans.}
\replaceTwo{We have presented RiskMonitor, which is an efficient and novel plug-and-play module for predicting collision risk using tokens from modern end-to-end AD models.}{We have formulated collision risk estimation for end-to-end planners as loss prediction and presented two efficient monitors: \tokmon{}, which reads the planner's tokens, and \newmon, which additionally reads the explicit per-agent geometry of the planner's decoded predictions.}
\replaceTwo{RiskMonitor predicts}{Both monitors predict} whether the collision loss, which is commonly used to train end-to-end planners, is positive given the planned waypoints. 
\replaceTwo{This prediction corresponds nicely to a probability of a collision happening, allowing us to threshold it and reject decisions with high risk.}{Thresholding this prediction raises an alarm for high-risk plans.}
\changeTwo{Under a single evaluation protocol for four planners, \tokmon{} has a higher AUROC than the rule-based and probabilistic baselines on most planners, while its average precision is not significantly different from that of a rule-based check of the planner's own predictions (RBD).
\newmon{} is significantly better than \tokmon{} and all baselines in AUROC on all four planners and in average precision on three of them.
At equal false-alarm rates, it never detects significantly fewer collision episodes than \tokmon{} or RBD, and at some operating points significantly more.}
\replaceTwo{When integrated with a simple braking policy, \tokmon{}
improves collision avoidance ability by $66.5\%$ on average, when tested closed-loop on safety-critical scenarios.}{When their alarms trigger braking in closed-loop safety-critical scenarios, the collision rate of UniAD falls from $63.9\%$ to $23.7\%$ with \newmon{} and to $26.5\%$ with \tokmon{}.}
\replaceTwo{Despite limitations, we are confident that our RiskMonitor approach to collision risk estimation is a major step towards safer AD vehicles.}{These results show that the intermediate outputs of end-to-end planners carry a collision-risk signal that a lightweight monitor can extract without retraining the planner.}\\\\
\textbf{\large Declarations}\\\\
\textbf{Conflict of interest statement} The authors declare that they have no known competing financial interests or personal relationships that could have appeared to
influence the work reported in this paper.\\\\ 	   
\textbf{Funding Information} This work was funded by Swedish national strategic research environment ELLIIT, grant C08. Computations for this work were enabled
by the National Academic Infrastructure for Supercomputing in Sweden
(NAISS) and the \replaceTwo{Berzerlius}{Berzelius} resource provided by National Supercomputing Centre at Linköping University and the Knut and Alice Wallenberg foundation, partially funded by the Swedish Research Council
through grant agreement no. 2022-06725. This project is also supported
by the Wallenberg AI, Autonomous Systems, and Software Program
(WASP) funded by the Knut and Alice Wallenberg Foundation.\\\\
\textbf{Data Availability} This work does not propose any new dataset. The
dataset \citep{caesar2020nuscenes} that support the findings of this study is openly
available at the URL: https://www.nuscenes.org/\\\\

\setcounter{page}{1}

\begin{appendices}
\makeatletter
\def\sectionfont{\reset@font\fontfamily{\rmdefault}\fontsize{12.5bp}{15bp}\bfseries\selectfont\raggedright\boldmath}%
\let\@orighangfrom\@hangfrom
\def\@hangfrom#1{\ifdim\f@size pt>12.3pt \noindent#1\par\noindent\else\@orighangfrom{#1}\fi}
\makeatother

\section{Details about Collision Predictors}
\label{sec:CP-A}
\changeTwo{This appendix details the collision predictors compared in Sec.~\ref{sec:CP}.}

\changeTwo{
\subsection{Rule-based Discriminator}
\label{sec:rbd-A}
RBD applies the rule that defines the collision label to the planner's own predictions: the UniAD collision loss with axis-aligned box hulls and a $1.0$\,m safety distance, with the ego box of $1.85\times4.084$\,m oriented along the plan.
For each agent, the most likely predicted mode is placed by adding the predicted offsets to the agent's detected center, and the score is the overlap summed over agents and time steps.
The frame and the time base follow each planner's output conventions, so that the agent position and the plan waypoint of the same step refer to the same time ($0.5$\,s to $3$\,s), and when a plan step is shorter than $0.1$\,m, the ego heading is kept from the previous step, because the direction of such a short step is dominated by noise and would make the ego box rotate erratically.
We verified the conventions against the ground truth: with the ground-truth future boxes in place of the predicted ones and the ground-truth ego heading, the rule reproduces the collision label on every validation sample.
The agents are those that the planner passes to its motion forecaster (UniAD) or whose detection confidence is at least $0.5$ (VAD, GenAD, SparseDrive).
When the ego vehicle is stationary, UniAD's tracker sometimes detects the ego vehicle itself as a car (e.g., in frames 16--34 of scene \texttt{6d872d14}); this gives a non-zero RBD score to 22 negative samples.
We do not filter these detections, since learned monitors would receive the same outputs.
}

\subsection{\changeTwo{GMM}}
\label{sec:GMM-A}

\replaceTwo{The End2End planner of most AD systems provides}{Most end-to-end planners provide} a motion prediction~$\hat{\tau}_\motion\in~\mathbb{R}^{N_a\times N_m\times~N_T\times2}$ for each road user, in the form of an ensemble of~$N_m$ possible future trajectories with \replaceTwo{corresponding probabilities~$\left\{\pi_m\right\}_1^M$, for each individual trajectory}{probabilities~$\{\pi_m\}_{m=1}^{N_m}$} \citep{hu2023planning}.
We let these define a chain of GMMs, one for each time step~$k$~ and road user~$a$:
\begin{equation}
p(x_k | t_{a,k})=\sum_{m=1}^{N_m} \pi_m \mathcal{N}(x_k-\hat{\tau}_{\motion_{a,k,m}},\Sigma_k)
\end{equation}
\changeTwo{Here, $\pi_m$ is the softmax of the predicted mode scores.}
The covariance~$\Sigma_k$ \replaceTwo{is set to be the size of ego-vehicle when~$k=1$, and linearly increasing}{equals the size of the ego vehicle at~$k=1$ and increases linearly} with the prediction time, i.e.~$\Sigma_k=k\sigma_0{\bf I}$\changeTwo{, with $\sigma_0=(w/2)^2=0.856\,\mathrm{m}^2$, where $w=1.85\,$m is the ego width used in UniAD's collision loss}.

As in \changeTwo{the} model-based collision avoidance literature \citep{zhou2023interaction}, we model the chance of crashing into a specific road user at a given time step as the integral over the area of the ego-vehicle bounding box, denoted by $D$.
\begin{equation}
P(\collision |a, k) = \iint_{D} p(x_k | t_{a,k})
\end{equation}
\changeTwo{Here, $D$ is the ego footprint ($4.084\,\mathrm{m}\times1.85\,\mathrm{m}$) centered at the planned waypoint of step~$k$ and oriented along the direction between consecutive waypoints.
The predicted positions are placed in the same ego frame by adding each road user's detected center to its predicted offsets; as for RBD (Appendix~\ref{sec:rbd-A}), the frame and the time base follow each planner's conventions, so that the predicted position and the planned waypoint of step~$k$ both refer to the same time.
Because $\Sigma_k$ is isotropic, the integral has a closed form: in the frame of $D$, it is the product of two differences of one-dimensional normal CDFs.}

The chance of crashing into at least one road user at a given time step is
\begin{equation}
P(\collision | t_{k}) = 1 - \prod_{a=1}^{N_a} (1-P(\collision |a, k))
\end{equation}
The chance of getting at least one crash within the time horizon is
\begin{equation}
P(\collision) = 1 - \prod_{\changeTwo{k}=1}^{T} (1-P(\collision |t_{k}))
\end{equation}

\changeTwo{\parsection{Sensitivity}
The scale $\sigma_0$ is set from the ego width and is not tuned. Tab.~\ref{tab:gmm-sens} varies it over a 16-fold range and also replaces the isotropic covariance with the anisotropic, axis-aligned covariance $\Sigma_k=\mathrm{diag}(1.5,3)\,(1+\frac{k-1}{6})\,\mathrm{m}^2$, for which the integral is computed on a $20\times20$ grid.
AUROC stays between $73.6\%$ and $78.4\%$.
We also report the max-density score $R=\max_{a,k}p(x_k|t_{a,k})$ with the anisotropic covariance, i.e., the mixture density at the ego waypoint, which ignores the ego footprint and is not a probability.
It ranks samples better than the integrated probability, but all variants remain below \tokmon{} ($82.39\%$ AUROC, $35.95\%$ AP) in both AUROC and AP.
The GMM needs only the planner's single forward pass and is cheaper than \tokmon{}, but it ranks collisions worse (Tab.~\ref{tab:nuscenes}).}

\begin{table}[h!]
\centering
\footnotesize
\setlength\tabcolsep{1.2mm}
{\color{roundtwo}
\begin{tabular}{@{}lccccc}
\toprule
GMM variant & AUROC & AP & $\mathrm{Pr}_{30}$ & $\mathrm{Pr}_{50}$ & $\mathrm{Pr}_{70}$ \\ \midrule
$0.25\,\sigma_0$ & 76.68 & 19.08 & 26.81 & 19.62 & 15.88 \\
$0.5\,\sigma_0$ & 76.39 & 18.21 & 22.26 & 19.73 & 16.12 \\
$\sigma_0$ (Tab.~\ref{tab:nuscenes}) & 75.99 & 17.31 & 19.56 & 19.11 & 16.66 \\
$2\,\sigma_0$ & 75.25 & 16.71 & 19.80 & 17.30 & 15.65 \\
$4\,\sigma_0$ & 73.58 & 16.23 & 17.02 & 15.32 & 14.79 \\
Anisotropic $\Sigma_k$ & 78.37 & 18.98 & 24.59 & 19.92 & 16.96 \\ \midrule
Max-density score & 79.29 & 22.61 & 30.84 & 23.59 & 19.18 \\
\bottomrule
\end{tabular}}
\caption{\changeTwo{Sensitivity of the GMM baseline on UniAD, nuScenes validation split, under the protocol of Tab.~\ref{tab:nuscenes}. $\sigma_0=0.856\,\mathrm{m}^2$. All values are percentages.}}
\label{tab:gmm-sens}
\end{table}

\replaceTwo{It also takes one inference forward process to evaluate. Though more efficient compared with \tokmon{}, the performance is limited.}{}

\subsection{Monte Carlo Dropout}
\label{sec:drop}
\replaceTwo{Dropout Rate used during experiment is 0.1.}{The dropout rate is $0.1$.}
\changeTwo{The dropout layer is placed in the MLP fuser of the UniAD planning head, which fuses the ego queries into the planning query; it has no parameters, so the released checkpoint is used unchanged.
All other modules run in evaluation mode.
For each sample, one deterministic forward pass gives the plan that is labeled, and the planning head is then evaluated four more times with dropout active and without the test-time collision optimization.
The score is the trace of the covariance of the four plan endpoints.
With ten passes instead of four, the result changes little (AUROC $66.66\%$, AP $12.08\%$, against $63.98\%$ and $11.60\%$).
Although MC Dropout requires no additional caching or training, it needs repeated passes through the planning head, whereas \tokmon{} needs one pass of a small network on top of the planner's single forward pass.
Since only the planning head is repeated, ten stochastic passes for the 6019 validation samples took 19 minutes on four GH200 GPUs, including the deterministic forward pass.}

\changeTwo{
\subsection{Deep Ensemble}
\label{sec:DE}
We follow one rule for both planners: the modules produced by the planner's first training stage are taken from the released checkpoint and kept frozen in evaluation mode, and the modules that only the second stage trains are re-initialized with a different random seed for each member and trained with the official second-stage schedule.
For UniAD, the backbone, BEV encoder, tracking and mapping modules are frozen, and the motion, occupancy and planning heads are re-initialized and trained for 20 epochs.
For VAD, whose first stage already trains the detection, mapping and motion modules, only the planning modules are re-initialized and trained for 12 epochs with the official optimizer and losses.
Consequently, the UniAD members differ in their motion predictions and plans, and the VAD members in their plans only.
Tab.~\ref{tab:DE-gate} compares the members with the released checkpoints.
On UniAD, the members' L2 error is on average $8\%$ higher than that of the released checkpoint, while their collision rate is around that of the released checkpoint; on VAD, the members are at least as good in L2 and have a similar collision rate.
The members therefore plan comparably to the released checkpoints.

\begin{table*}[t]
\centering
\scriptsize
\setlength\tabcolsep{1mm}
{\color{roundtwo}
\begin{tabular}{@{}lcccc@{}}
\toprule
 & \multicolumn{2}{c}{L2 (m)} & \multicolumn{2}{c}{Col.\ rate (\%)} \\
Model & 1 / 2 / 3\,s & Avg. & 3\,s & Avg. \\ \midrule
UniAD, released & 0.51 / 0.98 / 1.65 & 1.05 & 0.61 & 0.29 \\
UniAD, members 1--5 & 0.53--0.55 / 1.03--1.09 / 1.74--1.87 & 1.10--1.17 & 0.51--0.70 & 0.21--0.30 \\ \midrule
VAD, released & 0.41 / 0.70 / 1.05 & 0.72 & 0.35 & 0.16 \\
VAD, members 1--5 & 0.38--0.39 / 0.66--0.68 / 1.01--1.03 & 0.68--0.70 & 0.28--0.31 & 0.13--0.17 \\
\bottomrule
\end{tabular}}
\caption{\changeTwo{Planning performance of the Deep Ensemble members on the nuScenes validation split, compared with the reproduced released checkpoints (Appendix~\ref{sec:repro}). Col.\ rate: each planner's own box collision rate; Avg.: mean over 1, 2 and 3\,s.}}
\label{tab:DE-gate}
\end{table*}

\parsection{Decision variables}
Let $m=1,\dots,M$ ($M=5$) index the members, $\hat{\tau}^m_\plan$ be the plan of member $m$ before any collision optimization, and $\hat{\tau}^m_\plan(T)$ its endpoint at $3$\,s. We evaluate four scores.
\begin{enumerate}
\item \emph{Endpoint variance}: $s=\mathrm{Var}_m[\hat{x}^m(T)]+\mathrm{Var}_m[\hat{y}^m(T)]$, the variance of the plan endpoint over the members summed over $x$ and $y$.
\item \emph{DE-GMM}: $s=\frac{1}{M}\sum_m P_m(\collision)$, where $P_m(\collision)$ is the GMM collision probability of Appendix~\ref{sec:GMM-A}, computed with the plan and the predicted agents of member $m$.
\item \emph{Clearance-LCB}: the clearance of member $m$,
\begin{equation*}
\begin{split}
d_m=\min_{t,a,k}\mathrm{dist}\big(&B_\mathrm{ego}(\hat{\tau}^m_\plan(t)),\\
&B_a(\hat{\tau}^{m,a,k}_\motion(t))\big)-1\,\mathrm{m},
\end{split}
\end{equation*}
is the smallest distance between the ego box along the plan and any predicted agent box, over all time steps $t$, agents $a$ and modes $k$, minus the $1$\,m margin of the label and capped at $50$\,m.
The score is the negative lower confidence bound of the clearance over the members, $s=-(\bar{d}-\kappa\,\sigma_d)$, with $\kappa=1$ (not tuned).
\item \emph{DE-RBD}: $s=\frac{1}{M}\sum_m \mathbf{1}[\mathrm{RBD}_m>0]$, the fraction of members for which RBD flags a collision between the member's own plan and its own predicted agents, i.e., an ensemble of RBD (Sec.~\ref{sec:CP}, Appendix~\ref{sec:rbd-A}).
\end{enumerate}
The ego box is $1.85\times4.084$\,m, as in the label. The agents are those used by RBD (UniAD: all tracked agents; VAD: agents with confidence $\geq0.5$).

\parsection{Results}
Tab.~\ref{tab:DE} reports the four scores.
The endpoint variance, a generic uncertainty measure, is only slightly better than chance, while the collision-relevant scores are much better.
Ensembling does not improve the GMM: on both planners, DE-GMM performs on par with the GMM of the released checkpoint (Tab.~\ref{tab:nuscenes}), and on VAD its AP is slightly lower.
DE-RBD is the best ensemble score in AP on both planners and in AUROC on UniAD; it is the Deep Ensemble entry of Tab.~\ref{tab:nuscenes}.
\newmon{} is significantly better than each of the four ensemble scores on both planners.

\begin{table}[h!]
\centering
\scriptsize
\setlength\tabcolsep{0.9mm}
{\color{roundtwo}
\begin{tabular}{@{}lcccc@{}}
\toprule
 & \multicolumn{2}{c}{UniAD} & \multicolumn{2}{c}{VAD} \\
\cmidrule(lr){2-3}\cmidrule(l){4-5}
Score & AUROC & AP & AUROC & AP \\ \midrule
Endpoint var. & \stk{59.87}{{\tiny[56.1, 63.8]}} & \stk{10.74}{{\tiny[7.8, 14.1]}} & \stk{62.69}{{\tiny[56.4, 68.8]}} & \stk{12.70}{{\tiny[10.0, 16.5]}} \\ \addlinespace[2pt]
DE-GMM & \stk{75.42}{{\tiny[69.0, 81.3]}} & \stk{16.78}{{\tiny[10.9, 24.7]}} & \stk{70.89}{{\tiny[65.6, 75.6]}} & \stk{18.60}{{\tiny[12.5, 26.6]}} \\ \addlinespace[2pt]
Clearance-LCB & \stk{78.01}{{\tiny[72.7, 82.6]}} & \stk{17.13}{{\tiny[11.7, 23.9]}} & \stk{\textbf{74.75}}{{\tiny[70.3, 78.5]}} & \stk{20.57}{{\tiny[14.3, 27.8]}} \\ \addlinespace[2pt]
DE-RBD & \stk{\textbf{81.53}}{{\tiny[75.1, 86.9]}} & \stk{\textbf{42.74}}{{\tiny[30.5, 56.2]}} & \stk{70.01}{{\tiny[65.1, 74.6]}} & \stk{\textbf{30.43}}{{\tiny[22.5, 38.9]}} \\
\bottomrule
\end{tabular}}
\caption{\changeTwo{Deep Ensemble scores on nuScenes (\%), with $95\%$ scene-bootstrap CIs. Bold: highest per column.}}
\label{tab:DE}
\end{table}
}

\section{Experimental Details}
\label{sec:exp-A}
\changeTwo{This appendix details the data, training and statistics behind Sec.~\ref{sec:realworld} and Sec.~\ref{sec:simulation}.
\subsection{Reproduction of the Planners}
\label{sec:repro}
All nuScenes results use the released checkpoints, run with each planner's official evaluation code.
Before exporting tokens, we checked that each planner reproduces its published planning results (Tab.~\ref{tab:repro}).
The planning L2 errors of all four planners agree with the published values within $0.03$\,m.
The collision rates of UniAD, VAD and SparseDrive are equal to or slightly lower than the published values; that of GenAD is slightly higher (about seven more samples at $3$\,s).
The released GenAD checkpoint does not predict the classes truck, bus and trailer: it labels these objects as cars, which lowers its detection mAP ($0.21$) under class-aware matching.

\begin{table}[h!]
\centering
\scriptsize
\setlength\tabcolsep{0.9mm}
\resizebox{\columnwidth}{!}{\color{roundtwo}
\begin{tabular}{@{}llcccc@{}}
\toprule
 & & \multicolumn{2}{c}{L2 (m)} & \multicolumn{2}{c}{Col.\ rate (\%)} \\
Planner & & 1 / 2 / 3\,s & Avg. & 3\,s & Avg. \\ \midrule
UniAD & ours & 0.51 / 0.98 / 1.65 & 1.05 & 0.61 & 0.29 \\
 & published & 0.48 / 0.96 / 1.65 & 1.03 & 0.71 & 0.31 \\ \midrule
VAD-Base & ours & 0.41 / 0.70 / 1.05 & 0.72 & 0.35 & 0.16 \\
 & published & 0.41 / 0.70 / 1.05 & 0.72 & 0.41 & 0.22 \\ \midrule
GenAD & ours & 0.25 / 0.46 / 0.76 & 0.49 & 0.45 & 0.25 \\
 & published & 0.28 / 0.49 / 0.78 & 0.52 & 0.34 & 0.19 \\ \midrule
SparseDrive & ours & 0.30 / 0.58 / 0.95 & 0.61 & 0.22 & 0.09 \\
 & published & 0.30 / 0.58 / 0.95 & 0.61 & 0.23 & 0.10 \\
\bottomrule
\end{tabular}}
\caption{\changeTwo{Planning performance of the released checkpoints on the nuScenes validation split, reproduced with each planner's own evaluation code, against the published values (for SparseDrive, those reported for the released SparseDrive-S checkpoint in its official repository, which corrects the collision-rate computation of the paper).}}
\label{tab:repro}
\end{table}

}

\subsection{\replaceTwo{Cache tokens}{Cached Tokens and Decoded Outputs}}
\label{sec:cache}
\changeTwo{This section describes how the tokens read by RiskMonitor (Sec.~\ref{sec:riskmonitor}) and the decoded outputs used by RBD, GMM and \newmon{} (Secs.~\ref{sec:CP} and~\ref{sec:newmon}) are exported from each planner.}
\replaceTwo{We use the pretrained checkpoints in the official GitHub repositories of UniAD, VAD, GenAD and SparseDrive during all the experiments.}{} %
\changeTwo{The tokens are exported with each planner's official inference pipeline: the model is in evaluation mode, the temporal state (history BEV features, tracks or instance memory) is carried over within each scene as in the official test script, and the planner's weights are not modified.
The plan is recorded before UniAD's test-time collision optimization and before SparseDrive's collision-aware re-scoring, and the label is computed on this plan.
Consequently, our positive rates are higher than the collision rates reported in \citep{hu2023planning,sun2025sparsedrive}.
For UniAD, the planning token is the planning query at the input of the planning head's BEV attention module, and the motion tokens are the last-layer trajectory queries of MotionFormer.}
\changeTwo{On UniAD, which forecasts motion only for tracked agents, keyframes without a tracked vehicle cannot be scored and are excluded ($1369$ of $6019$ keyframes; $4650$ remain).}
For VAD, \replaceTwo{we cache the tokens as the input to the final regression branch \texttt{self.ego\_fut\_decoder}.}{the planning token is the input to the final planning regression branch \texttt{self.ego\_fut\_decoder}, and the motion tokens are the $100$ most confident agent queries.}
\replaceTwo{See the code in the original UniAD repository:}{See the UniAD code:}
\url{https://github.com/OpenDriveLab/UniAD/blob/v2.0/projects/mmdet3d_plugin/uniad/dense_heads/planning_head.py#L190}
\replaceTwo{See VAD code in the original repository:}{See the VAD code:}
\url{https://github.com/hustvl/VAD/blob/36047b6b5985e01832d8a2ecb0355d7f3c753ee1/projects/mmdet3d_plugin/VAD/VAD_head.py#L792}

{
In the parallel motion planner, SparseDrive concatenates the ego instance with surrounding agents to get agent-level instances, and then three types of \replaceTwo{attentions are performed}{attention are applied}.
We cache the tokens output by the attention modules and slice them into ego-planning tokens and motion tokens.
See the SparseDrive code:
\url{https://github.com/swc-17/SparseDrive/blob/main/projects/mmdet3d_plugin/models/motion/motion_planning_head.py#L315-L316}.

GenAD \replaceTwo{extracts instance-centric scene representation first}{first extracts an instance-centric scene representation} and passes it to the future trajectory generator. We cache the instance-centric scene representation and slice \replaceTwo{them}{it} into ego-planning tokens and motion tokens.
See the GenAD code:
\url{https://github.com/wzzheng/GenAD/blob/main/projects/mmdet3d_plugin/GenAD/GenAD_head.py#L756}.
}

\changeTwo{
For each planner, the export thus locates the existing tensors that hold the planning token and the agent tokens; for SparseDrive and GenAD, where ego and agent instances are stored in one tensor, it splits this tensor by index into the ego-planning and motion tokens.
The monitors are therefore compatible with heterogeneous token-based planners, but attaching them to a new planner requires this planner-specific access to its intermediate representations.

\parsection{Decoded outputs for RBD, GMM and \newmon}
Besides the tokens, we export the planner's decoded outputs: the plan waypoints, and for each agent its detected box (center, size, yaw, class, score) and its predicted multimodal trajectories with mode scores.
For UniAD, these are the agents that the tracker passes to MotionFormer; for VAD, GenAD and SparseDrive, the 100 most confident detection queries, of which RBD and GMM use those with confidence at least $0.5$.
The planners differ in how they express the predicted agent displacements (as offsets from the agent's center or as per-step increments, and in which axes) and in the time base of the agent and ego waypoints.
For each planner, we determined these conventions from its code and verified them against the ground truth: the decoded agent positions are compared with the ground-truth future positions of the matched objects, and the label rule reproduces the collision label (Appendix~\ref{sec:rbd-A}).

\parsection{Geometry features of \newmon}
For each agent $a$, the predicted trajectory is $\hat{\tau}_{\motion,a}\in\mathbb{R}^{N_m\times N_T\times 2}$ with mode probabilities $\pi_a$, and the geometry feature vector $g_a\in\mathbb{R}^{119}$ of \newmon{} (Sec.~\ref{sec:newmon}), computed identically for all planners, consists of
(i) the displacement between the agent's most likely predicted position and the ego waypoint at each of the $N_T=6$ future steps ($0.5$ to $3$\,s), expressed both in the planner's ego frame and in the frame of the ego box oriented along the plan ($2\times12$ values);
(ii) the same displacement for all $N_m=6$ modes in the ego frame ($72$ values);
(iii) the mode probabilities ($6$ values);
(iv) the agent's box width and length, the sine and cosine of its yaw, its class (one-hot over the ten nuScenes detection classes), its detection score, and its current center ($17$ values).
Distances are divided by $10$\,m.
}

\subsection{A Balanced NeuroNCAP Dataset}

\label{sec:rationale}

\replaceTwo{AD models in NeuroNCAP are used for inference only, i.e., in the offical Github repository there is no collision loss implementation since it only requires inference process in close-loop evaluation. Therefore, we transplant the collision training loss to the End2End AD models.}{NeuroNCAP runs the planners for inference only, so its official GitHub repository does not implement the collision loss; we therefore compute the label as in Sec.~\ref{sec:simulation} against the future boxes of the simulated actors.}
\replaceTwo{To construct a balanced simulated dataset using NeuroNCAP, we}{We} first generate safety-critical scenarios using the NeuRAD simulator \citep{tonderski2024neurad}, which \replaceTwo{is capable of simulating}{simulates} three types of collision scenarios: \textit{frontal}, \textit{side}, and \textit{stationary}. Each scenario type consists of multiple scene-sequence pairs: five for frontal (0103, 0106, 0110, 0346, 0923), five for side (0103, 0108, 0110, 0278, 0921), and ten for stationary (0099, 0101, 0103, 0106, 0108, 0278, 0331, 0783, 0796, 0966). 

For each scene-sequence pair, we initially conduct \replaceTwo{100}{10} simulation runs, recording the number of timestamps and the number of collision timestamps. \replaceTwo{This provides a preliminary statistical analysis of the dataset.}{} Given these statistics, we assume that for any scene-sequence pair \( i \), the number of timestamps per run (\( t_i \)) and the collision-to-timestamp ratio remain constant. That is, each run \( r_i \) of a particular scene-sequence pair produces \( c_i \) collisions and \( t_i \) timestamps.

\changeTwo{Under this label, the plans before the collision optimization collide in most samples of most pairs: with the same number of runs for every pair, as in the official NeuroNCAP evaluation, $60.5\%$ of the UniAD samples would be positive, so the share of collision samples needs to be lowered.}
\replaceTwo{To ensure a balanced dataset while maintaining a realistic overall collision rate}{To this end}, we formulate an Integer Linear Programming (ILP) problem. The objective is to minimize the largest difference \( D \) in the number of runs among all scene-sequence pairs:

\begin{equation}
\min_{r_i} D
\label{eq:ilp}
\end{equation}

subject to the following constraints:

\begin{equation}
|r_i - r_j| \leq D, \quad \forall i,j
\end{equation}

\begin{equation}
r_i \geq 45, \quad \forall i
\end{equation}

\begin{equation}
\sum_i c_i \cdot r_i \geq 0.46 \sum_i t_i \cdot r_i
\end{equation}

\begin{equation}
\sum_i c_i \cdot r_i \leq 0.54 \sum_i t_i \cdot r_i
\end{equation}

\begin{equation}
\sum_i r_i \leq 1200
\end{equation}

These constraints ensure that: (1) the variation in the number of runs among scene-sequence pairs is minimized, (2) every scene-sequence pair is used at least 45 times to maintain diversity, (3) the total collision rate remains within \replaceTwo{49\%–51\%}{46\%–54\%}, and (4) the total number of runs does not exceed 1200. Solving the ILP in Eq.~\replaceTwo{\ref{eq:ilp}}{(\ref{eq:ilp})} yields the optimal number of runs \( r_i \) for each scene-sequence pair.
Once the optimal values of \( r_i \) are determined, we execute the corresponding simulation runs in NeuroNCAP\replaceTwo{, generating a well-balanced dataset that effectively captures safety-critical driving scenarios. This dataset serves as the foundation for evaluating the performance of RiskMonitor and other models in diverse and challenging collision scenarios.}{.}

\changeTwo{
For UniAD, the ILP is feasible only at the upper bound of the collision-rate constraint, with an expected collision fraction of $54\%$.
The solution has $45$ runs for each pair, and $83$--$84$ runs for the four pairs with the lowest collision fractions (side-0108, side-0921, stationary-0099 and stationary-0106).
We then ran the planned number of runs a second time with new random seeds, and a third time for the four low pairs only.

The final dataset has $2439$ runs ($90$ per pair, and $249$--$252$ for the four low pairs) and $28025$ labeled samples, of which $50.4\%$ are positive.
}

\subsection{Training Details for Collision Predictors}
\label{sec:train-detail}
\changeTwo{\parsection{nuScenes}}
\replaceTwo{nuScenes: Specifically, we use an ensemble of size 4 and~$\alpha_f=0.75$~ in the focal loss, and the mixup parameter~$\alpha_m=3.0$.
Both \tokmon{} and LP-FCN use the same batch size, optimizer, bagging ensemble, and focal loss.}{All learned monitors (\tokmon{}, \newmon{} and LP-FCN) are trained on all 700 nuScenes training scenes with the same bagging ensemble of four members, each of which contains all positive samples and one quarter of the negative samples, and with the same focal loss ($\alpha_f=0.75$, $\gamma=2$), optimizer (AdamW with an initial learning rate of $10^{-3}$ and weight decay $0.01$; for \tokmon{} and \newmon{}, the learning rate rises linearly over the first 660 iterations and then decays in proportion to the inverse of the iteration count) and batch size of $32$.
They are evaluated after a fixed 200 epochs; no hyperparameter is selected on the validation split.
The design of \newmon{} was chosen on held-out UniAD training scenes before the validation split was read (Sec.~\ref{sec:ablation}) and used unchanged for all four planners, where only the token dimension differs (256 for UniAD and SparseDrive, 512 for VAD and GenAD). The hidden widths of the geometry encoder and the per-agent head were set to 256 a priori and not tuned.
\tokmon{} is trained with mixup ($\alpha_m=3.0$), whereas \newmon{} and LP-FCN are trained without it (Sec.~\ref{sec:ablation}).
For LP-FCN, the learning rate starts at $1.5\times10^{-6}$ and follows a per-epoch cosine decay; with the learning-rate schedule of \tokmon{}, its AUROC was lower on all four planners.}

\changeTwo{\parsection{NeuroNCAP}
On NeuroNCAP, \tokmon{}, \newmon{} and LP-FCN are trained on the training folds of each cross-validation split (Sec.~\ref{sec:simulation}) with the same optimizer and batch size as on nuScenes, but for 20 epochs, with focal loss ($\alpha_f=0.5$, $\gamma=2$) and, for \tokmon{} only, mixup ($\alpha_m=1.0$); the epoch is chosen by the AUROC on the validation fold.
Each monitor is an ensemble of four bagging members, each of which leaves out a quarter of the training runs.}

\changeTwo{\parsection{Mixup}
\tokmon{} suffers from a distribution shift between the training and evaluation sequences (nuScenes and NeuroNCAP, see Fig.~\ref{fig:tsne}).
Mixup~\citep{zhang2017mixup} has been reported to improve generalization on OOD data \citep{thulasidasan2019mixup}.
In our experiments, however, it does not improve performance under this shift: training \tokmon{} without it (variant (a) in Sec.~\ref{sec:ablation}) does not change the result significantly, and training \newmon{} with it removes most of its gain (Sec.~\ref{sec:ablation}).}

\changeTwo{
\subsection{Statistical Analysis}
\label{sec:bootstrap}
The 6019 nuScenes validation keyframes come from 150 scenes of about 20\,s, and frames within a scene are strongly correlated.
We therefore resample scenes rather than frames: each of the 2000 bootstrap resamples draws 150 scenes with replacement and keeps all evaluated frames of each drawn scene.
A metric's $95\%$ CI is given by the $2.5\%$ and $97.5\%$ percentiles of its values over the resamples.
For a comparison between two methods (Tab.~\ref{tab:paired}), both are evaluated on the same resamples and the interval is computed for the difference; we call a difference significant when this interval excludes zero.
The same resamples are used for Tab.~\ref{tab:nuscenes}, Tab.~\ref{tab:paired}, the Deep Ensemble and MC Dropout results, the ablation (Tab.~\ref{tab:ablation}), the stratified evaluation (Appendix~\ref{sec:reliance}) and the event-level false alarms (Tab.~\ref{tab:event}).
For the event-level measures, the thresholds are fixed on the full validation split and only the counts are resampled.
We do not train each monitor with several seeds; as an indication of training variability, we report the range of the four bagging members, which are trained on different subsets with different initializations (Sec.~\ref{sec:realworld}).
On NeuroNCAP, the bootstrap resamples the 14 scenes, and all runs and scenario types of a scene are drawn together (Tab.~\ref{tab:neuroncap}).
}

\section{Additional Results}
\label{sec:result-A}
\subsection{\changeTwo{Event-Level Results at Fixed False-Alarm Rates}}
\label{sec:event-A}
\changeTwo{Tab.~\ref{tab:event} lists the values of Fig.~\ref{fig:froc} at the three operating points of Sec.~\ref{sec:event}.
The rate of false-alarm events is not monotonic in the threshold, because lowering the threshold can join adjacent alarm events into one run of consecutive alarms (in the limit, flagging every keyframe leaves one event per scene); we therefore use the lowest threshold at which the rate has not yet exceeded $F$.
Tied scores are flagged in random order, and we report the expected values over 200 orders, as for $\mathrm{Pr}_X$.
CIs come from the scene bootstrap with the thresholds fixed.}

\begin{table*}[t]
\centering
\footnotesize
\setlength\tabcolsep{1.5mm}
{\color{roundtwo}
\begin{tabular}{@{}llrlrlrlrlrl@{}}
\toprule
 & & \multicolumn{6}{c}{Detected collision episodes (\%) $\uparrow$} & \multicolumn{4}{c}{Paired differences (percentage points)} \\
\cmidrule(lr){3-8}\cmidrule(l){9-12}
Planner & False alarms & \multicolumn{2}{c}{\newmonS{}} & \multicolumn{2}{c}{\tokmonS{}} & \multicolumn{2}{c}{RBD} & \multicolumn{2}{c}{\newmonS{} $-$ \tokmonS{}} & \multicolumn{2}{c}{\newmonS{} $-$ RBD} \\
\midrule
\multirow{3}{*}{UniAD} & 0.5/min (120\,s) & 52.7 & {\tiny[40.6, 65.1]} & 40.9 & {\tiny[29.0, 52.8]} & 38.7 & {\tiny[28.3, 48.6]} & $+$11.8 & {\tiny[$-$0.9, 24.4]} & \textbf{$+$14.0} & {\tiny[3.7, 25.3]} \\
 & 1/min (60\,s) & 68.8 & {\tiny[58.3, 79.5]} & 55.9 & {\tiny[44.0, 68.4]} & 67.7 & {\tiny[56.8, 77.5]} & $+$12.9 & {\tiny[$-$1.1, 25.5]} & $+$1.1 & {\tiny[$-$9.8, 12.7]} \\
 & 2/min (30\,s) & 83.9 & {\tiny[75.0, 92.0]} & 67.7 & {\tiny[57.0, 78.1]} & 75.1 & {\tiny[66.0, 83.5]} & \textbf{$+$16.1} & {\tiny[4.7, 27.5]} & \textbf{$+$8.7} & {\tiny[1.4, 17.0]} \\
\midrule
\multirow{3}{*}{VAD} & 0.5/min (120\,s) & 33.6 & {\tiny[24.3, 43.4]} & 30.6 & {\tiny[21.7, 39.9]} & 31.3 & {\tiny[23.1, 40.3]} & $+$3.0 & {\tiny[$-$8.4, 14.4]} & $+$2.2 & {\tiny[$-$6.5, 11.2]} \\
 & 1/min (60\,s) & 45.5 & {\tiny[35.7, 55.2]} & 43.3 & {\tiny[34.5, 52.6]} & 45.5 & {\tiny[36.4, 55.0]} & $+$2.2 & {\tiny[$-$8.9, 13.3]} & $+$0.0 & {\tiny[$-$8.9, 9.1]} \\
 & 2/min (30\,s) & 70.9 & {\tiny[63.7, 78.6]} & 59.0 & {\tiny[49.2, 68.7]} & 48.1 & {\tiny[39.4, 57.2]} & \textbf{$+$11.9} & {\tiny[3.4, 21.3]} & \textbf{$+$22.8} & {\tiny[14.7, 31.4]} \\
\midrule
\multirow{3}{*}{GenAD} & 0.5/min (120\,s) & 47.8 & {\tiny[37.9, 58.0]} & 43.5 & {\tiny[33.1, 53.6]} & 42.6 & {\tiny[33.1, 52.1]} & $+$4.3 & {\tiny[$-$4.1, 12.7]} & $+$5.2 & {\tiny[$-$1.6, 12.3]} \\
 & 1/min (60\,s) & 68.7 & {\tiny[58.7, 78.0]} & 50.4 & {\tiny[40.0, 61.2]} & 60.0 & {\tiny[50.4, 69.2]} & \textbf{$+$18.3} & {\tiny[8.8, 27.8]} & $+$8.7 & {\tiny[0.0, 19.0]} \\
 & 2/min (30\,s) & 77.4 & {\tiny[68.2, 85.1]} & 70.4 & {\tiny[61.1, 79.4]} & 67.2 & {\tiny[57.8, 75.7]} & $+$7.0 & {\tiny[$-$0.7, 14.8]} & \textbf{$+$10.2} & {\tiny[1.0, 19.2]} \\
\midrule
\multirow{3}{*}{SparseDrive} & 0.5/min (120\,s) & 54.0 & {\tiny[42.5, 64.5]} & 36.3 & {\tiny[26.6, 46.0]} & 44.2 & {\tiny[33.9, 54.5]} & \textbf{$+$17.7} & {\tiny[8.1, 27.7]} & $+$9.7 & {\tiny[$-$0.8, 20.0]} \\
 & 1/min (60\,s) & 58.4 & {\tiny[47.7, 68.8]} & 56.6 & {\tiny[47.3, 66.1]} & 66.4 & {\tiny[57.0, 75.0]} & $+$1.8 & {\tiny[$-$7.4, 10.0]} & $-$8.0 & {\tiny[$-$18.8, 3.0]} \\
 & 2/min (30\,s) & 73.5 & {\tiny[64.1, 82.5]} & 64.6 & {\tiny[55.5, 73.1]} & 70.0 & {\tiny[61.0, 78.2]} & $+$8.8 & {\tiny[0.0, 17.7]} & $+$3.4 & {\tiny[$-$6.0, 12.9]} \\
\bottomrule
\end{tabular}}
\caption{\changeTwo{Detected collision episodes at fixed rates of false-alarm events on the nuScenes validation split (150 scenes, 50 minutes). RM-G: \newmon{}; RM-T: \tokmon{}. The threshold of each method is set so that the rate of false-alarm events equals the target (achieved rates within $0.02$ per minute of it; Sec.~\ref{sec:event}). Brackets: $95\%$ scene-bootstrap CIs; bold: significant paired difference.}}
\label{tab:event}
\end{table*}

{
\subsection{\replaceTwo{RiskMontior's reliance on E2E planner perception analysis}{RiskMonitor's Reliance on the Planner's Perception}}
\label{sec:reliance}
\replaceTwo{In the sequential design of E2E models, like UniAD, three cases could lead to a collision:}{To analyze the collisions with objects missed by the planner (Sec.~\ref{sec:comparison}), we distinguish three causes of a collision in sequential end-to-end planners such as UniAD:}
\begin{enumerate}
    \item The multi-object tracking (MOT) module misdetects, i.e., a FN;
    \item MOT works, but \replaceTwo{the motion forecasting module fails}{motion forecasting fails};
    \item Both the MOT and the forecasting modules work, but the planning module fails.
\end{enumerate}

\replaceTwo{For the first case, we count how many collided objects (in the future) are detected by E2E planners at the frame when the planning action is made.
More specifically, on nuScenes, \texttt{sample\_annotation\_token} is the unique identifier for each object, \texttt{sample\_token} is the frame identifier, and \texttt{instance\_token} is the tracked object's identifier.
We first find the collided object's \texttt{sample\_annotation\_token} at the frame of collision, then its \texttt{instance\_token}.
We then use its \texttt{instance\_token} to find the \texttt{sample\_annotation\_token} in the frame when the planning action is made.
Finally, we check the detection evaluation to see if this ground truth object is matched with a prediction.}{}

}
\changeTwo{
\parsection{Matching}
For every collision sample, the collided object is the ground-truth object that triggers the label at the earliest colliding step (with the largest overlap at that step); it is located in the planning frame through its nuScenes \texttt{instance\_token}.
It counts as detected if the planner's detections at the planning frame match it under the nuScenes detection evaluation with the vehicle classes (car, truck, bus, trailer and construction vehicle) merged, since the collision label does not depend on them: detections with a score of at least $0.2$ are matched, in decreasing order of score, to the nearest unmatched object within a center distance of $2$\,m.
Objects of classes that the nuScenes detection benchmark does not evaluate (e.g., debris or pushable objects) are not assigned to a case.

\parsection{Failure causes on four planners}
Tab.~\ref{tab:causes} lists the failure causes: collisions with an undetected object (case~1) account for $10.6\%$--$20.7\%$ of the collisions, and collisions with detected objects (cases~2 and~3) for $71\%$--$85\%$.

\begin{table}[h!]
\centering
\scriptsize
\setlength\tabcolsep{0.7mm}
{\color{roundtwo}
\begin{tabular}{@{}lcccc@{}}
\toprule
Planner & Total & Case 1 & Cases 2+3 & Other \\ \midrule
UniAD & 376 & 40 (10.6\%) & 320 (85.1\%) & 16 (4.3\%) \\
VAD & 463 & 96 (20.7\%) & 328 (70.8\%) & 39 (8.4\%) \\
GenAD & 457 & 58 (12.7\%) & 374 (81.8\%) & 25 (5.5\%) \\
SparseDrive & 445 & 58 (13.0\%) & 358 (80.4\%) & 29 (6.5\%) \\
\bottomrule
\end{tabular}}
\caption{\changeTwo{Failure causes of the collision samples on nuScenes (validation split): number of collision samples and share (\%). Case~1: the collided object is not detected by the planner; cases~2+3: it is detected; other: objects of classes that the nuScenes detection benchmark does not evaluate. The released GenAD checkpoint labels trucks, buses and trailers as cars (Appendix~\ref{sec:repro}); with class-aware matching, its case-1 share would be $32.4\%$.}}
\label{tab:causes}
\end{table}

\parsection{Stratified evaluation}
Tab.~\ref{tab:stratified} reports two measures for each case.
The per-case AUROC ranks the collisions of one case against all non-collision samples; it is independent of the threshold.
The per-case Recall@FPR10\% is the recall within one case at the threshold where the overall false-positive rate is $10\%$.
We fix the false-positive rate rather than the overall recall so that both monitors are compared at the same false-alarm cost; at a fixed overall recall, a monitor that detects more collisions of one case necessarily detects fewer of the other.

\parsection{Cases 2 and 3}
At this operating point, \tokmon{} detects $56.7\%$--$65.8\%$ and \newmon{} $69.8\%$--$78.8\%$ of the collisions with detected objects.
\newmon{} detects significantly more of them than \tokmon{} on all four planners (from $+8.4$ [$0.3$, $16.1$] points on SparseDrive to $+21.9$ [$12.9$, $33.9$] on UniAD), and its per-case AUROC is also significantly higher on all four.

\parsection{Case 1 and the perception bound}
Since an undetected object has no motion token, one might expect the efficacy of \tokmon{} to be bounded by the share of case~1.
Tab.~\ref{tab:stratified} shows that this bound does not hold strictly.
The case-1 AUROC of \tokmon{} ($81.8\%$--$89.3\%$) is similar to that on cases~2 and~3 ($82.7\%$--$88.6\%$), while at the same operating point it detects $12.5\%$--$51.7\%$ of the case-1 collisions, fewer than in cases~2 and~3 on all four planners.
The effect is strongest on UniAD, where \tokmon{} detects only $12.5\%$ of them.
The two measures differ because the AUROC averages over all thresholds, whereas Recall@FPR10\% counts only the collisions ranked among the highest-risk samples: case-1 collisions are rarely ranked as low as normal driving, but fewer of them reach the top of the ranking.
An object missed by the planner thus makes a collision harder to detect at a low false-alarm rate, but not invisible to the monitor.
The tokens therefore carry information about objects that the planner did not detect, consistent with the observation that UniAD still attends to agents missed by its tracker (Fig.~12 of \citet{hu2023planning}).
The planner's perception thus affects, but does not strictly bound, the performance of \tokmon{}.

On case~1, \newmon{} is on par with or better than \tokmon{} on all four planners: its per-case AUROC is significantly higher on VAD, GenAD and SparseDrive.
At the operating point, it detects a share of the case-1 collisions similar to that of cases~2 and~3 on VAD and SparseDrive, but a smaller one on GenAD and, in particular, on UniAD ($17.5\%$ versus $78.8\%$).
UniAD passes only the agents it tracks to downstream modules, so the geometry of an untracked object is not available to \newmon{}; on UniAD, \newmon{} is not significantly different from \tokmon{} on case~1 (AUROC $-7.3$ [$-14.2$, $16.8$]).

\begin{table*}[t]
\centering
\scriptsize
\setlength\tabcolsep{1.5mm}
{\color{roundtwo}
\begin{tabular}{@{}lcccc@{}}
\toprule
 & \multicolumn{2}{c}{Per-case AUROC} & \multicolumn{2}{c}{Per-case Recall@FPR10\%} \\
Planner & case 1 & cases 2+3 & case 1 & cases 2+3 \\ \midrule
UniAD & 75.8 / 83.1 & \textbf{93.2} / 82.7 & 17.5 / 12.5 & \textbf{78.8} / 56.9 \\
VAD & \textbf{92.2} / 81.8 & \textbf{90.9} / 86.1 & 75.0 / 42.7 & \textbf{69.8} / 56.7 \\
GenAD & \textbf{89.8} / 85.7 & \textbf{91.0} / 87.8 & 63.8 / 48.3 & \textbf{77.0} / 65.8 \\
SparseDrive & \textbf{92.7} / 89.3 & \textbf{92.0} / 88.6 & 77.6 / 51.7 & \textbf{72.9} / 64.5 \\
\midrule
\multicolumn{5}{@{}l}{\newmonS{} $-$ \tokmonS{}, $95\%$ CIs} \\
UniAD & $-7.3$ {\tiny[$-14.2$, $16.8$]} & $+10.5$ {\tiny[$6.9$, $14.5$]} & $+5.0$ {\tiny[$-9.1$, $46.3$]} & $+21.9$ {\tiny[$12.9$, $33.9$]} \\
VAD & $+10.4$ {\tiny[$1.1$, $15.6$]} & $+4.7$ {\tiny[$2.2$, $7.5$]} & $+32.3$ {\tiny[$-11.5$, $54.3$]} & $+13.1$ {\tiny[$3.8$, $23.0$]} \\
GenAD & $+4.1$ {\tiny[$0.3$, $10.7$]} & $+3.1$ {\tiny[$1.0$, $5.6$]} & $+15.5$ {\tiny[$-5.9$, $33.3$]} & $+11.2$ {\tiny[$4.0$, $20.8$]} \\
SparseDrive & $+3.4$ {\tiny[$0.5$, $5.1$]} & $+3.4$ {\tiny[$1.3$, $5.7$]} & $+25.9$ {\tiny[$-5.6$, $45.9$]} & $+8.4$ {\tiny[$0.3$, $16.1$]} \\
\bottomrule
\end{tabular}}
\caption{\changeTwo{Stratified evaluation by failure cause on nuScenes (\%): \newmon{} / \tokmon{}. RM-T/RM-G: \tokmon{}/\newmon{}. Cases as in Tab.~\ref{tab:causes}. Per-case AUROC: collisions of one case against all non-collision samples. Recall@FPR10\%: recall within the case at the threshold where the overall false-positive rate is $10\%$. Bold: the paired difference (bottom) is significant. Intervals: scene bootstrap (Appendix~\ref{sec:bootstrap}).}}
\label{tab:stratified}
\end{table*}
}

\replaceTwo{See Fig.~D3 for FP predictions from RBD baseline.}{}

\subsection{\changeTwo{Precision--Recall Curves on NeuroNCAP}}
\label{sec:ncap-pr}
\begin{figure}[t]
\centering
\includegraphics[width=\linewidth]{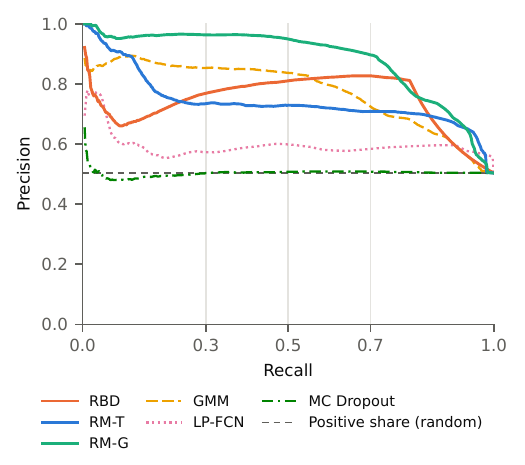}
\caption{\changeTwo{Precision--recall curves of the collision predictors on NeuroNCAP with UniAD; the methods are those of Tab.~\ref{tab:neuroncap}. The scores of the five cross-validation folds are merged, and samples without a score receive the lowest score. Precision is the expected precision with tied scores in random order, as for $\mathrm{Pr}_X$ (Sec.~\ref{sec:metrics}). AP (\%): RBD $72.25$, \tokmon{} $74.23$, \newmon{} $88.49$, GMM $77.69$, LP-FCN $59.58$, MC Dropout $50.46$. Dashed line: positive rate ($50.4\%$). RM-T/RM-G: \tokmon{}/\newmon{}.}}
\label{fig:pr-ncap}
\end{figure}

\changeTwo{Fig.~\ref{fig:pr-ncap} shows the precision--recall curves of the collision predictors on NeuroNCAP.
The non-zero scores of RBD cover the positives up to a recall of $79.6\%$, beyond which its precision falls linearly to the positive rate.
\newmon{} is more precise than RBD at almost every recall, except at about $77\%$--$82\%$ and above $98\%$.}

\subsection{\changeTwo{Closed-Loop Braking}}
\label{sec:brake-A}
\changeTwo{This appendix details the monitor-triggered braking of Tab.~\ref{tab:brake}.

\parsection{Implementation}
NeuroNCAP replays the logged ego motion for the first $p$ planning calls of a run ($p=3$ for side-0921 and $p=2$ for all other scene-sequence pairs) and executes the planner's output from call $p$ on.
From call $p$ on, the monitor scores every step from the same inputs as the cached datasets (Appendix~\ref{sec:cache}); re-scoring the saved inputs offline reproduces the online scores (largest difference $10^{-14}$).
A step without any tracked agent raises no alarm of \tokmon{} or \newmon{}.
Each scene is scored by the ensemble of the cross-validation fold that tests it (Tab.~\ref{tab:neuroncap}), with the threshold that maximizes Youden's J on the validation fold of that fold (Tab.~\ref{tab:brake-thr}).
From the first step whose score reaches the threshold, the planner outputs an all-zero trajectory, and NeuroNCAP's vehicle model applies its stopping controller (deceleration $1.5\,v$, at most $6\,\mathrm{m/s^2}$), as in NeuroNCAP's own baseline; the brake stays engaged to the end of the run.}

\begin{table}[t]
\centering
\footnotesize
\setlength\tabcolsep{1.0mm}
{\color{roundtwo}
\resizebox{\columnwidth}{!}{
\begin{tabular}{@{}clccc@{}}
\toprule
Fold & Validation scenes & \newmonS{} & \tokmonS{} & RBD \\ \midrule
0 & 0099, 0101, 0796 & 0.433 (90.5/7.3) & 0.559 (75.4/15.9) & 0.019 (87.9/6.1) \\
1 & 0106, 0923, 0966 & 0.601 (86.0/15.5) & 0.580 (75.6/22.0) & 0.645 (75.7/20.3) \\
2 & 0278, 0346, 0921 & 0.440 (96.5/6.1) & 0.524 (89.6/23.2) & 0.003 (77.3/3.7) \\
3 & 0103, 0110, 0331, 0783 & 0.761 (74.2/19.6) & 0.727 (30.6/7.3) & 0.371 (71.4/13.0) \\
4 & 0108 & 0.403 (44.4/14.2) & 0.391 (77.7/25.2) & 0.001 (85.8/45.0) \\
\bottomrule
\end{tabular}}}
\caption{\changeTwo{Braking thresholds of the five folds (Youden's J on the validation fold), with the true- and false-positive rates (\%) on the validation fold in brackets. The test scenes of fold $k$ are the validation scenes of fold $k-1$ (fold 0: fold 4).}}
\label{tab:brake-thr}
\end{table}

\subsection{Qualitative Examples on NeuroNCAP\replaceTwo{ Dataset}{}}
\label{sec:qual-ncap}

\changeTwo{Fig.~\ref{fig:quantitative-examples-ncap} (Sec.~\ref{sec:simulation}) shows a true positive (Fig.~\ref{fig:tp1}) and a false negative (Fig.~\ref{fig:fn1}) of \tokmon{} with UniAD.
On the left of Fig.~\ref{fig:fn1}, the BEV display reveals that the UniAD model fails to detect the bus (the huge grey rectangle without a blue box indicates the wrong detection result) in front of the ego vehicle.
Consequently, the motion token lacks information about the bus, leading \tokmon{} to misclassify the scenario.
This suggests that the accuracy of the perception modules in the AD model directly influences the predictive capability of \tokmon{}.

In the true positive example of Fig.~\ref{fig:tp1}, UniAD provides more accurate coordinate estimations for a nearby object (indicated by the dark red box), allowing \tokmon{} to correctly predict the collision.
Interestingly, the plan from UniAD is still going straight even if there is a detected road user coming from the side. 
This highlights the necessity of having collision prediction when deploying end-to-end planners for AD.}

\begin{figure}[htbp]
    \centering
    \includegraphics[width=1.0\linewidth]{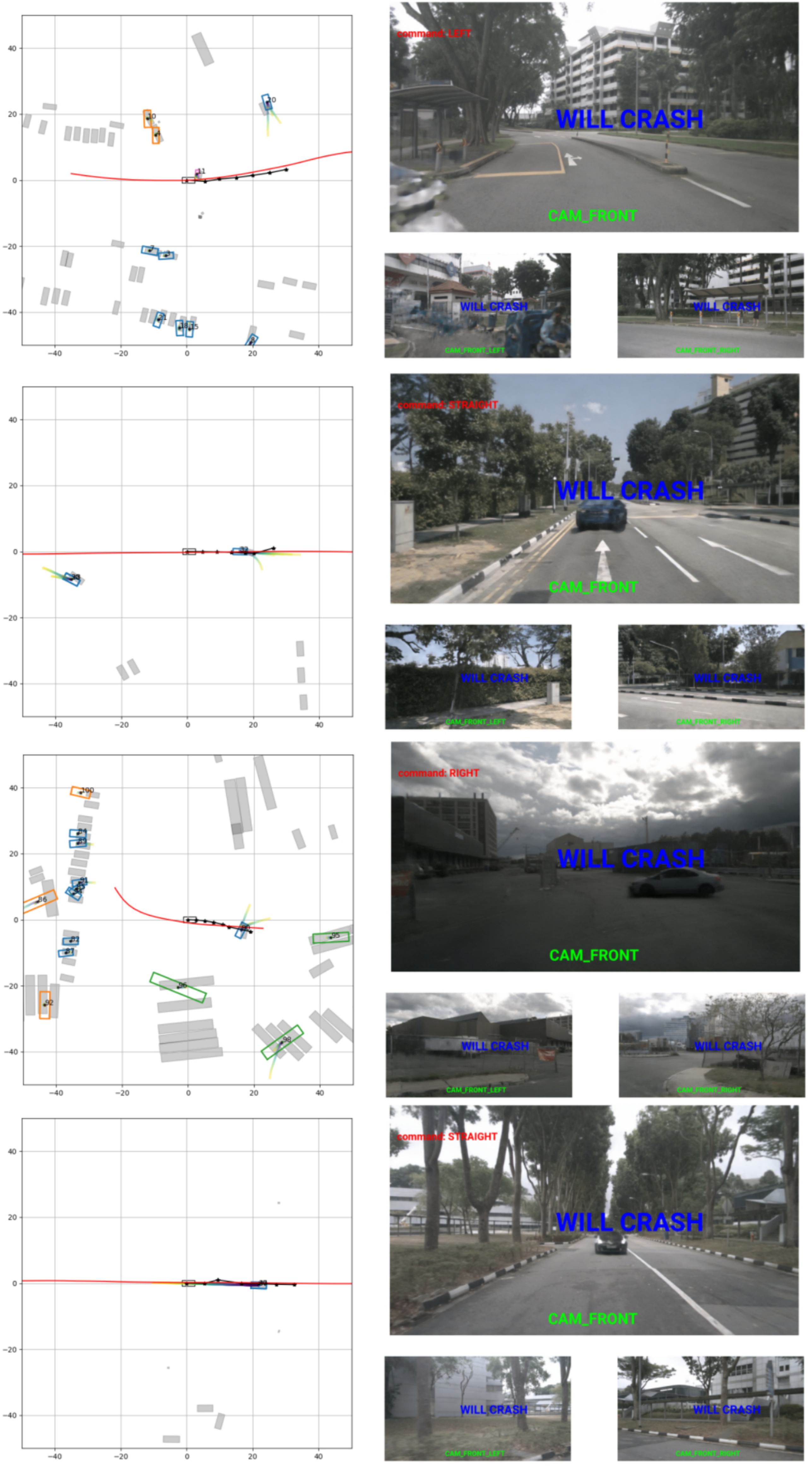}
    \caption{\replaceTwo{True Positive Examples on NeuroNCAP Simulation Dataset (Scenarios}{True positives of \tokmon{} on NeuroNCAP with UniAD (scenarios} from top to bottom: side, stationary, stationary, frontal)}
    \label{fig:fn-agg-ncap-tp}
\end{figure}

\begin{figure}[htbp]
    \centering
    \includegraphics[width=1.0\linewidth]{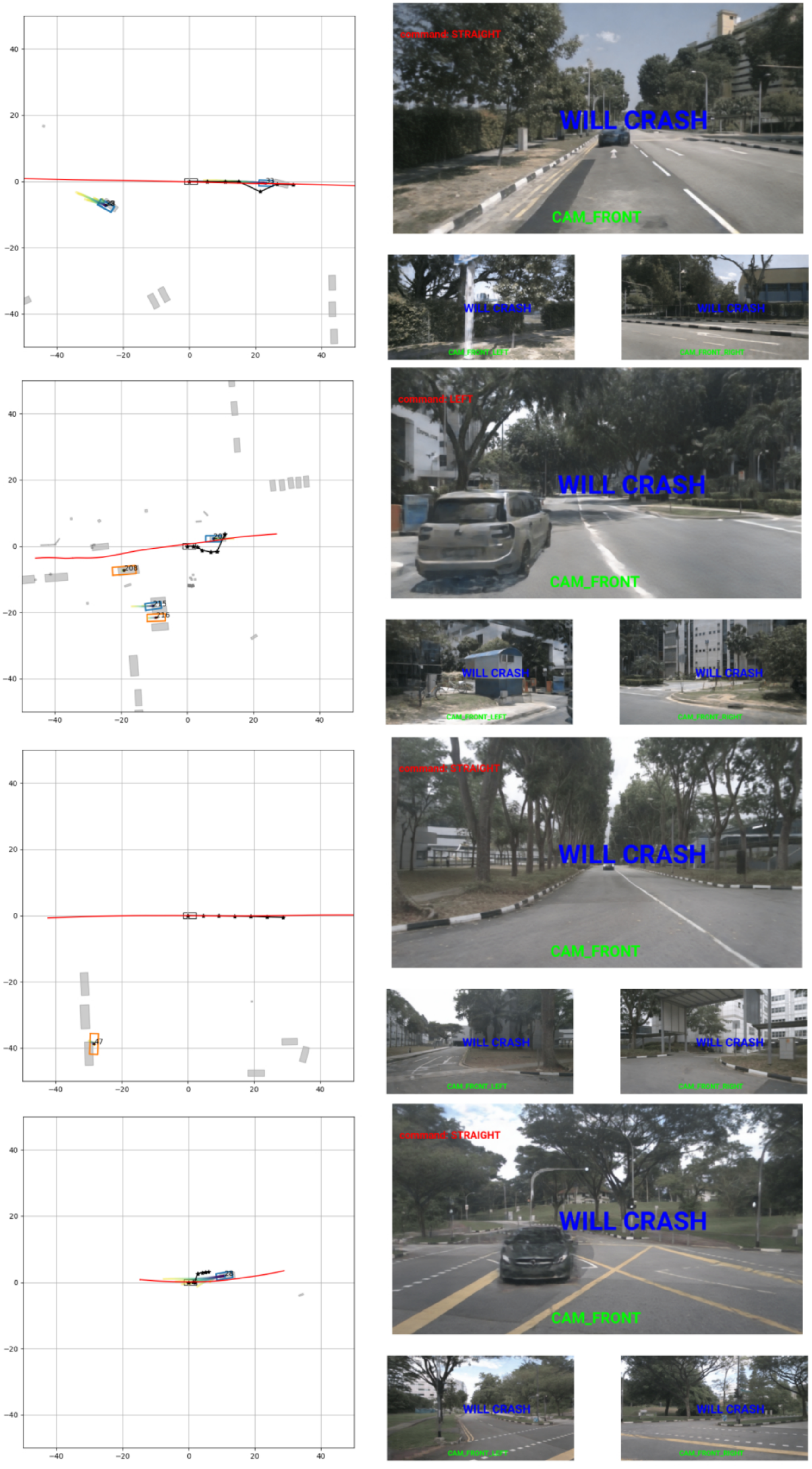}
    \caption{\replaceTwo{False Negative Examples on the NeuroNCAP Simulation Dataset (Scenarios}{False negatives of \tokmon{} on NeuroNCAP with UniAD (scenarios} from top to bottom: stationary, stationary, frontal, frontal)}
    \label{fig:fn-agg-ncap-fn}
\end{figure}

\replaceTwo{We demonstrate qualitative examples in the NeuroNCAP Simulation Dataset with UniAD as the end-to-end planner.}{}
The true positive examples in which \tokmon{} has succeeded in predicting collisions are shown in Fig.~\ref{fig:fn-agg-ncap-tp}, while false negative examples in which \tokmon{} fails to predict collisions are shown in Fig.~\ref{fig:fn-agg-ncap-fn}.

As seen in the first and third images of Fig.~\ref{fig:fn-agg-ncap-fn}, the AD model's predicted coordinates for objects near the ego-vehicle are inaccurate.
In the first image, for example, the ground truth distance to the ego-vehicle is farther than the estimated distance.
In the third image, the failure to detect the object ahead results in incorrectly classifying a collision sample as a non-collision.
\replaceTwo{Additionally, in the second sample, our model's incorrect prediction}{In the second sample, the incorrect prediction of \tokmon{}} may have been caused by the AD model's erroneous estimation, where a stationary object was misunderstood as moving by motion token.

\replaceTwo{In the true positive examples shown in Fig.~\ref{fig:fn-agg-ncap-tp}, UniAD provides more accurate coordinate estimations for a nearby object, allowing \tokmon{} to accurately identify the collision.}{}

\subsection{Distribution Shift Analysis}
\label{sec:tsne}
\replaceTwo{In Sec.~\ref{sec:result-A} and Sec.~\ref{sec:ab-A}, we note that the performance on the validation tokens is substantially better than on the test tokens.
During experiments in Sec.~\ref{sec:simulation}, we also note that the test performance is nearly perfect if we allow training tokens and test tokens from the same sequence, even if they belong to different scenario types.
This indicates that it is the distribution shift between different AD video sequences that hinders RiskMonitor from generalizing and thus causes a drop in the performance on test tokens.}{}

\begin{figure*}[htbp!]
    \centering

    \begin{subfigure}[b]{0.39\linewidth}
        \centering
        \includegraphics[trim={1.5cm 1.3cm 1.5cm 1.3cm}, clip, width=0.95\linewidth]{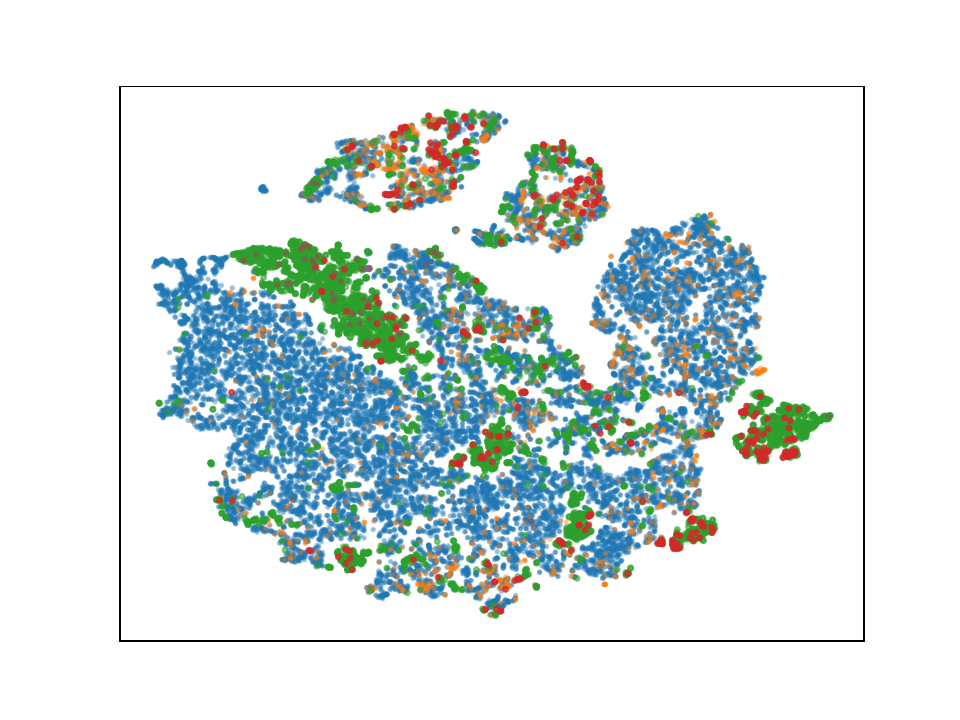}
        \caption{nuScenes}
        \label{fig:tsne-nuscense}
    \end{subfigure}
    \hfill
    \begin{subfigure}[b]{0.2\linewidth}
        \centering
        \includegraphics[width=0.95\linewidth]{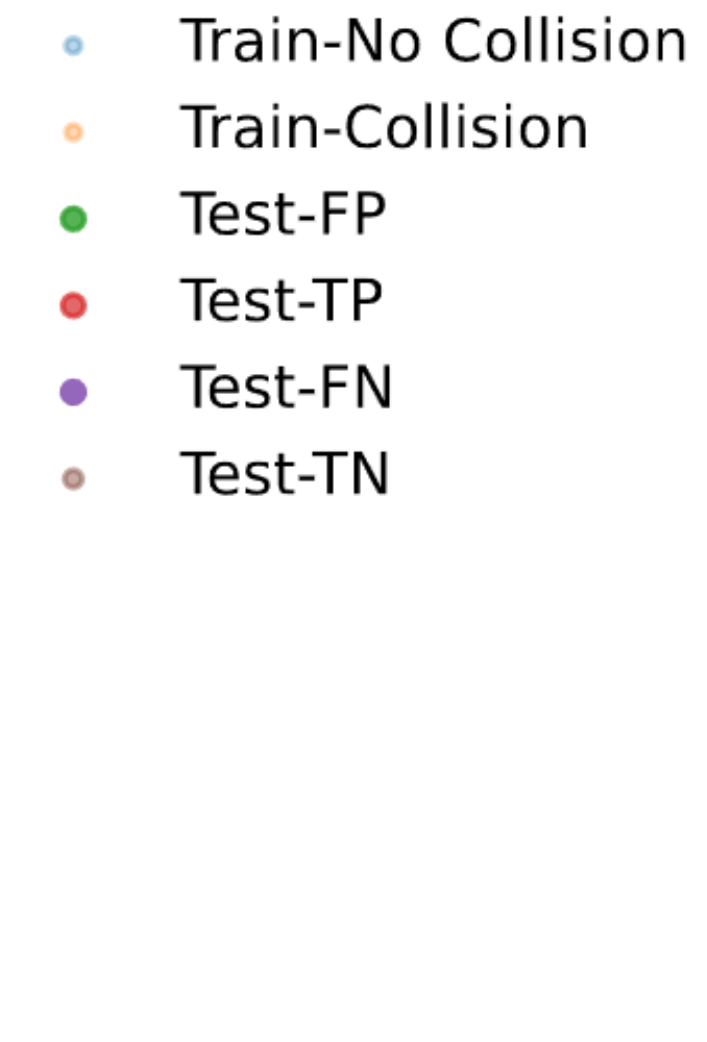}
        \label{fig:legend}
    \end{subfigure}
    \hfill
    \begin{subfigure}[b]{0.37\linewidth}
        \centering
        \includegraphics[trim={0.2cm 0.35cm 0.2cm 0cm}, clip, width=\linewidth]{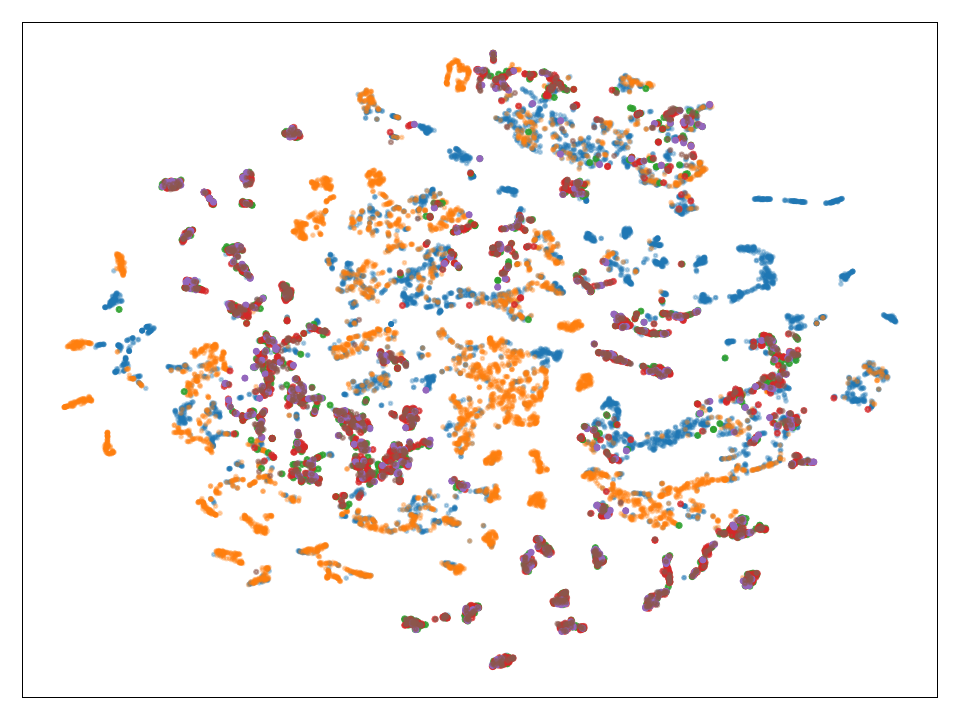}
        \caption{NeuroNCAP}
        \label{fig:tsne-ncap}
    \end{subfigure}
    
    \caption{2D t-SNE plots for plan tokens $h_\plan$ from UniAD on nuScenes and NeuroNCAP. 
    It is better to review in color. For the training tokens, we assign a color to distinguish collision and non-collision samples.
For the test tokens, we assign colors based on the test results, true positives (TP), false positives (FP), true negatives (TN), and false negatives (FN).}
    \label{fig:tsne}
\end{figure*}

\replaceTwo{In Fig.~\ref{fig:tsne}, we use the t-SNE algorithm to reduce the dimension of plan tokens and visualize them in 2D space. 
we visualize the planning token~$h_\plan$ from UniAD on nuScenes in both the training and validation split.}{The planner tokens carry sequence-specific structure (Sec.~\ref{sec:discussions}). Fig.~\ref{fig:tsne} visualizes it with t-SNE for the planning tokens~$h_\plan$ of UniAD on the training and validation splits of nuScenes.}
In Fig.~\ref{fig:tsne-nuscense}, \replaceTwo{it is notable that }{}(1) there are two clusters at the top, mixed with training and test samples, and one larger cluster at the bottom, mainly consisting of training samples; (2) there is a hole in the bottom cluster, which happens to be filled with test samples that are FP predictions; (3) there are more TP predictions in the top two clusters, where \replaceTwo{there seem to be less distribution shift between training and test samples}{the shift between training and test samples seems smaller}.
\replaceTwo{In Fig.~\ref{fig:clusters}, we further visualize the corresponding video sequence of these samples and find that the top-left cluster roughly corresponds to turning right in the video, the top-right cluster roughly corresponds to turning left, and the bottom cluster roughly corresponds to going straight.}{}
\replaceTwo{These findings reveal that the distribution shift between the plan and road user tokens that correspond to different sequences is significant.}{These findings indicate a large distribution shift between the tokens of different sequences.}
\replaceTwo{The distribution shift between nuScenes train and validation splits is the main reason for the performance drop of RiskMonitor on the test tokens.}{This structure is consistent with the lower performance of the monitors on unseen sequences.}
\replaceTwo{In mixup, setting~$\beta=3.0$ improves generalization since a convex combination of sample-target pairs has the potential to alleviate the gap. }{Mixup does not improve performance under this shift (Appendix~\ref{sec:train-detail}).}

In Fig.~\ref{fig:tsne-ncap}, we similarly visualize the plan tokens~$h_\plan$ on NeuroNCAP, which also exhibit distinct clusters corresponding to different scenario types and sequences.
The distribution appears more dispersed than in nuScenes\replaceTwo{, reflecting the greater variability and discreteness introduced by the simulation environments}{}. 
\replaceTwo{In addition, the overlap between training and test samples is relatively low, indicating a significant distribution shift between the test and training sets with different scenarios and sequences.}{The training and test samples overlap little, which indicates a large shift between scenarios and sequences.}
\replaceTwo{These distribution shifts become more pronounced when a particular scenario type or sequence has little resemblance to the training distribution. Consequently, RiskMonitor's performance declines for tokens lying far from learned regions of the feature space.}{}

\replaceTwo{Fig.~\ref{fig:clusters} visualizes the three front views in the training sequences from different clusters in Fig.~\ref{fig:tsne-nuscense}. Each cluster corresponds to turning left, turning right and going straight plan.}{}

\end{appendices}

\clearpage
\bibliography{sn-bibliography}%


\begin{thebibliography}{58}
\ifx \bisbn   \undefined \def \bisbn  #1{ISBN #1}\fi
\ifx \binits  \undefined \def \binits#1{#1}\fi
\ifx \bauthor  \undefined \def \bauthor#1{#1}\fi
\ifx \batitle  \undefined \def \batitle#1{#1}\fi
\ifx \bjtitle  \undefined \def \bjtitle#1{#1}\fi
\ifx \bvolume  \undefined \def \bvolume#1{\textbf{#1}}\fi
\ifx \byear  \undefined \def \byear#1{#1}\fi
\ifx \bissue  \undefined \def \bissue#1{#1}\fi
\ifx \bfpage  \undefined \def \bfpage#1{#1}\fi
\ifx \blpage  \undefined \def \blpage #1{#1}\fi
\ifx \burl  \undefined \def \burl#1{\textsf{#1}}\fi
\ifx \doiurl  \undefined \def \doiurl#1{\url{https://doi.org/#1}}\fi
\ifx \betal  \undefined \def \betal{\textit{et al.}}\fi
\ifx \binstitute  \undefined \def \binstitute#1{#1}\fi
\ifx \binstitutionaled  \undefined \def \binstitutionaled#1{#1}\fi
\ifx \bctitle  \undefined \def \bctitle#1{#1}\fi
\ifx \beditor  \undefined \def \beditor#1{#1}\fi
\ifx \bpublisher  \undefined \def \bpublisher#1{#1}\fi
\ifx \bbtitle  \undefined \def \bbtitle#1{#1}\fi
\ifx \bedition  \undefined \def \bedition#1{#1}\fi
\ifx \bseriesno  \undefined \def \bseriesno#1{#1}\fi
\ifx \blocation  \undefined \def \blocation#1{#1}\fi
\ifx \bsertitle  \undefined \def \bsertitle#1{#1}\fi
\ifx \bsnm \undefined \def \bsnm#1{#1}\fi
\ifx \bsuffix \undefined \def \bsuffix#1{#1}\fi
\ifx \bparticle \undefined \def \bparticle#1{#1}\fi
\ifx \barticle \undefined \def \barticle#1{#1}\fi
\bibcommenthead
\ifx \bconfdate \undefined \def \bconfdate #1{#1}\fi
\ifx \botherref \undefined \def \botherref #1{#1}\fi
\ifx \url \undefined \def \url#1{\textsf{#1}}\fi
\ifx \bchapter \undefined \def \bchapter#1{#1}\fi
\ifx \bbook \undefined \def \bbook#1{#1}\fi
\ifx \bcomment \undefined \def \bcomment#1{#1}\fi
\ifx \oauthor \undefined \def \oauthor#1{#1}\fi
\ifx \citeauthoryear \undefined \def \citeauthoryear#1{#1}\fi
\ifx \endbibitem  \undefined \def \endbibitem {}\fi
\ifx \bconflocation  \undefined \def \bconflocation#1{#1}\fi
\ifx \arxivurl  \undefined \def \arxivurl#1{\textsf{#1}}\fi
\csname PreBibitemsHook\endcsname

\bibitem[\protect\citeauthoryear{Akan and G{\"u}ney}{2022}]{akan2022stretchbev}
\begin{bchapter}
\bauthor{\bsnm{Akan}, \binits{A.K.}},
\bauthor{\bsnm{G{\"u}ney}, \binits{F.}}:
\bctitle{Stretchbev: Stretching future instance prediction spatially and
  temporally}.
In: \bbtitle{European Conference on Computer Vision},
pp. \bfpage{444}--\blpage{460}
(\byear{2022}).
\bcomment{Springer}
\end{bchapter}
\endbibitem

\bibitem[\protect\citeauthoryear{Altendorfer and
  Wilkmann}{2017}]{altendorfer_whatis}
\begin{botherref}
\oauthor{\bsnm{Altendorfer}, \binits{R.}},
\oauthor{\bsnm{Wilkmann}, \binits{C.}}:
What is the collision probability and how to compute it
(2017)
\doiurl{10.48550/arXiv.1711.07060}
\end{botherref}
\endbibitem

\bibitem[\protect\citeauthoryear{Bahari et~al.}{2021}]{bahari_vehicle_2021}
\begin{botherref}
\oauthor{\bsnm{Bahari}, \binits{M.}},
\oauthor{\bsnm{Saadatnejad}, \binits{S.}},
\oauthor{\bsnm{Rahimi}, \binits{A.}},
\oauthor{\bsnm{Shaverdikondori}, \binits{M.}},
\oauthor{\bsnm{Shahidzadeh}, \binits{A.-H.}},
\oauthor{\bsnm{Moosavi-Dezfooli}, \binits{S.-M.}},
\oauthor{\bsnm{Alahi}, \binits{A.}}:
Vehicle trajectory prediction works, but not everywhere.
{arXiv}.
Version Number: 2
(2021).
\doiurl{10.48550/ARXIV.2112.03909} .
\url{https://arxiv.org/abs/2112.03909}
Accessed 2025-07-04
\end{botherref}
\endbibitem

\bibitem[\protect\citeauthoryear{Caesar et~al.}{2020}]{caesar2020nuscenes}
\begin{bchapter}
\bauthor{\bsnm{Caesar}, \binits{H.}},
\bauthor{\bsnm{Bankiti}, \binits{V.}},
\bauthor{\bsnm{Lang}, \binits{A.H.}},
\bauthor{\bsnm{Vora}, \binits{S.}},
\bauthor{\bsnm{Liong}, \binits{V.E.}},
\bauthor{\bsnm{Xu}, \binits{Q.}},
\bauthor{\bsnm{Krishnan}, \binits{A.}},
\bauthor{\bsnm{Pan}, \binits{Y.}},
\bauthor{\bsnm{Baldan}, \binits{G.}},
\bauthor{\bsnm{Beijbom}, \binits{O.}}:
\bctitle{nuscenes: A multimodal dataset for autonomous driving}.
In: \bbtitle{Proceedings of the IEEE/CVF Conference on Computer Vision and
  Pattern Recognition},
pp. \bfpage{11621}--\blpage{11631}
(\byear{2020})
\end{bchapter}
\endbibitem

\bibitem[\protect\citeauthoryear{Chen et~al.}{2021}]{chen2021learning}
\begin{bchapter}
\bauthor{\bsnm{Chen}, \binits{D.}},
\bauthor{\bsnm{Koltun}, \binits{V.}},
\bauthor{\bsnm{Kr{\"a}henb{\"u}hl}, \binits{P.}}:
\bctitle{Learning to drive from a world on rails}.
In: \bbtitle{Proceedings of the IEEE/CVF International Conference on Computer
  Vision},
pp. \bfpage{15590}--\blpage{15599}
(\byear{2021})
\end{bchapter}
\endbibitem

\bibitem[\protect\citeauthoryear{Codevilla et~al.}{2018}]{codevilla2018end}
\begin{bchapter}
\bauthor{\bsnm{Codevilla}, \binits{F.}},
\bauthor{\bsnm{M{\"u}ller}, \binits{M.}},
\bauthor{\bsnm{L{\'o}pez}, \binits{A.}},
\bauthor{\bsnm{Koltun}, \binits{V.}},
\bauthor{\bsnm{Dosovitskiy}, \binits{A.}}:
\bctitle{End-to-end driving via conditional imitation learning}.
In: \bbtitle{2018 IEEE International Conference on Robotics and Automation
  (ICRA)},
pp. \bfpage{4693}--\blpage{4700}
(\byear{2018}).
\bcomment{IEEE}
\end{bchapter}
\endbibitem

\bibitem[\protect\citeauthoryear{Cai et~al.}{2020}]{cai2020probabilistic}
\begin{barticle}
\bauthor{\bsnm{Cai}, \binits{P.}},
\bauthor{\bsnm{Wang}, \binits{S.}},
\bauthor{\bsnm{Sun}, \binits{Y.}},
\bauthor{\bsnm{Liu}, \binits{M.}}:
\batitle{Probabilistic end-to-end vehicle navigation in complex dynamic
  environments with multimodal sensor fusion}.
\bjtitle{IEEE Robotics and Automation Letters}
\bvolume{5}(\bissue{3}),
\bfpage{4218}--\blpage{4224}
(\byear{2020})
\end{barticle}
\endbibitem

\bibitem[\protect\citeauthoryear{Chen et~al.}{2024}]{chen_eccv24_ppad}
\begin{bchapter}
\bauthor{\bsnm{Chen}, \binits{Z.}},
\bauthor{\bsnm{Ye}, \binits{M.}},
\bauthor{\bsnm{Xu}, \binits{S.}},
\bauthor{\bsnm{Cao}, \binits{T.}},
\bauthor{\bsnm{Chen}, \binits{Q.}}:
\bctitle{Ppad: Iterative interactions of prediction and planning for end-to-end
  autonomous driving}.
In: \bbtitle{Proceedings of the European Conference on Computer Vision
  ({ECCV})}
(\byear{2024})
\end{bchapter}
\endbibitem

\bibitem[\protect\citeauthoryear{Dong et~al.}{2026}]{dong2026to}
\begin{botherref}
\oauthor{\bsnm{Dong}, \binits{H.}},
\oauthor{\bsnm{Liu}, \binits{M.}},
\oauthor{\bsnm{Liang}, \binits{J.}},
\oauthor{\bsnm{Chatzi}, \binits{E.}},
\oauthor{\bsnm{Fink}, \binits{O.}}:
To trust or not to trust your vision-language model's prediction.
Transactions on Machine Learning Research
(2026)
\end{botherref}
\endbibitem

\bibitem[\protect\citeauthoryear{Dosovitskiy
  et~al.}{2017}]{dosovitskiy2017carla}
\begin{bchapter}
\bauthor{\bsnm{Dosovitskiy}, \binits{A.}},
\bauthor{\bsnm{Ros}, \binits{G.}},
\bauthor{\bsnm{Codevilla}, \binits{F.}},
\bauthor{\bsnm{Lopez}, \binits{A.}},
\bauthor{\bsnm{Koltun}, \binits{V.}}:
\bctitle{Carla: An open urban driving simulator}.
In: \bbtitle{Conference on Robot Learning},
pp. \bfpage{1}--\blpage{16}
(\byear{2017}).
\bcomment{PMLR}
\end{bchapter}
\endbibitem

\bibitem[\protect\citeauthoryear{Deo et~al.}{2022}]{deo2022multimodal}
\begin{bchapter}
\bauthor{\bsnm{Deo}, \binits{N.}},
\bauthor{\bsnm{Wolff}, \binits{E.}},
\bauthor{\bsnm{Beijbom}, \binits{O.}}:
\bctitle{Multimodal trajectory prediction conditioned on lane-graph
  traversals}.
In: \bbtitle{Conference on Robot Learning},
pp. \bfpage{203}--\blpage{212}
(\byear{2022}).
\bcomment{PMLR}
\end{bchapter}
\endbibitem

\bibitem[\protect\citeauthoryear{Ferreira et~al.}{2025}]{ferreira2025safety}
\begin{barticle}
\bauthor{\bsnm{Ferreira}, \binits{R.S.}},
\bauthor{\bsnm{Gu{\'e}rin}, \binits{J.}},
\bauthor{\bsnm{Delmas}, \binits{K.}},
\bauthor{\bsnm{Guiochet}, \binits{J.}},
\bauthor{\bsnm{Waeselynck}, \binits{H.}}:
\batitle{Safety monitoring of machine learning perception functions: A survey}.
\bjtitle{Computational Intelligence}
\bvolume{41}(\bissue{2}),
\bfpage{70032}
(\byear{2025})
\doiurl{10.1111/coin.70032}
\end{barticle}
\endbibitem

\bibitem[\protect\citeauthoryear{Gal and Ghahramani}{2016}]{gal2016dropout}
\begin{bchapter}
\bauthor{\bsnm{Gal}, \binits{Y.}},
\bauthor{\bsnm{Ghahramani}, \binits{Z.}}:
\bctitle{Dropout as a bayesian approximation: Representing model uncertainty in
  deep learning}.
In: \bbtitle{International Conference on Machine Learning},
pp. \bfpage{1050}--\blpage{1059}
(\byear{2016}).
\bcomment{PMLR}
\end{bchapter}
\endbibitem

\bibitem[\protect\citeauthoryear{Guo et~al.}{2017}]{guo2017calibration}
\begin{bchapter}
\bauthor{\bsnm{Guo}, \binits{C.}},
\bauthor{\bsnm{Pleiss}, \binits{G.}},
\bauthor{\bsnm{Sun}, \binits{Y.}},
\bauthor{\bsnm{Weinberger}, \binits{K.Q.}}:
\bctitle{On calibration of modern neural networks}.
In: \bbtitle{International Conference on Machine Learning},
pp. \bfpage{1321}--\blpage{1330}
(\byear{2017}).
\bcomment{PMLR}
\end{bchapter}
\endbibitem

\bibitem[\protect\citeauthoryear{Houenou et~al.}{2014}]{houenou_risk}
\begin{barticle}
\bauthor{\bsnm{Houenou}, \binits{A.}},
\bauthor{\bsnm{Bonnifait}, \binits{P.}},
\bauthor{\bsnm{Cherfaoui}, \binits{V.}}:
\batitle{Risk assessment for collision avoidance systems}.
\bjtitle{2014 17th IEEE International Conference on Intelligent Transportation
  Systems, ITSC 2014}
(\byear{2014})
\doiurl{10.1109/ITSC.2014.6957721}
\end{barticle}
\endbibitem

\bibitem[\protect\citeauthoryear{Hu et~al.}{2022}]{hu2022st}
\begin{bchapter}
\bauthor{\bsnm{Hu}, \binits{S.}},
\bauthor{\bsnm{Chen}, \binits{L.}},
\bauthor{\bsnm{Wu}, \binits{P.}},
\bauthor{\bsnm{Li}, \binits{H.}},
\bauthor{\bsnm{Yan}, \binits{J.}},
\bauthor{\bsnm{Tao}, \binits{D.}}:
\bctitle{St-p3: End-to-end vision-based autonomous driving via spatial-temporal
  feature learning}.
In: \bbtitle{European Conference on Computer Vision},
pp. \bfpage{533}--\blpage{549}
(\byear{2022}).
\bcomment{Springer}
\end{bchapter}
\endbibitem

\bibitem[\protect\citeauthoryear{Hu et~al.}{2021}]{hu2021safe}
\begin{bchapter}
\bauthor{\bsnm{Hu}, \binits{P.}},
\bauthor{\bsnm{Huang}, \binits{A.}},
\bauthor{\bsnm{Dolan}, \binits{J.}},
\bauthor{\bsnm{Held}, \binits{D.}},
\bauthor{\bsnm{Ramanan}, \binits{D.}}:
\bctitle{Safe local motion planning with self-supervised freespace
  forecasting}.
In: \bbtitle{Proceedings of the IEEE/CVF Conference on Computer Vision and
  Pattern Recognition},
pp. \bfpage{12732}--\blpage{12741}
(\byear{2021})
\end{bchapter}
\endbibitem

\bibitem[\protect\citeauthoryear{Huang et~al.}{2021}]{9511299}
\begin{barticle}
\bauthor{\bsnm{Huang}, \binits{C.}},
\bauthor{\bsnm{Hang}, \binits{P.}},
\bauthor{\bsnm{Hu}, \binits{Z.}},
\bauthor{\bsnm{Lv}, \binits{C.}}:
\batitle{Collision-probability-aware human-machine cooperative planning for
  safe automated driving}.
\bjtitle{IEEE Transactions on Vehicular Technology}
\bvolume{70}(\bissue{10}),
\bfpage{9752}--\blpage{9763}
(\byear{2021})
\doiurl{10.1109/TVT.2021.3102251}
\end{barticle}
\endbibitem

\bibitem[\protect\citeauthoryear{Hu et~al.}{2021}]{hu2021fiery}
\begin{bchapter}
\bauthor{\bsnm{Hu}, \binits{A.}},
\bauthor{\bsnm{Murez}, \binits{Z.}},
\bauthor{\bsnm{Mohan}, \binits{N.}},
\bauthor{\bsnm{Dudas}, \binits{S.}},
\bauthor{\bsnm{Hawke}, \binits{J.}},
\bauthor{\bsnm{Badrinarayanan}, \binits{V.}},
\bauthor{\bsnm{Cipolla}, \binits{R.}},
\bauthor{\bsnm{Kendall}, \binits{A.}}:
\bctitle{Fiery: Future instance prediction in bird's-eye view from surround
  monocular cameras}.
In: \bbtitle{Proceedings of the IEEE/CVF International Conference on Computer
  Vision},
pp. \bfpage{15273}--\blpage{15282}
(\byear{2021})
\end{bchapter}
\endbibitem

\bibitem[\protect\citeauthoryear{Hu et~al.}{2023}]{hu2023planning}
\begin{bchapter}
\bauthor{\bsnm{Hu}, \binits{Y.}},
\bauthor{\bsnm{Yang}, \binits{J.}},
\bauthor{\bsnm{Chen}, \binits{L.}},
\bauthor{\bsnm{Li}, \binits{K.}},
\bauthor{\bsnm{Sima}, \binits{C.}},
\bauthor{\bsnm{Zhu}, \binits{X.}},
\bauthor{\bsnm{Chai}, \binits{S.}},
\bauthor{\bsnm{Du}, \binits{S.}},
\bauthor{\bsnm{Lin}, \binits{T.}},
\bauthor{\bsnm{Wang}, \binits{W.}}, \betal:
\bctitle{Planning-oriented autonomous driving}.
In: \bbtitle{Proceedings of the IEEE/CVF Conference on Computer Vision and
  Pattern Recognition},
pp. \bfpage{17853}--\blpage{17862}
(\byear{2023})
\end{bchapter}
\endbibitem

\bibitem[\protect\citeauthoryear{Hu et~al.}{2018}]{hu2018gmm}
\begin{bchapter}
\bauthor{\bsnm{Hu}, \binits{Y.}},
\bauthor{\bsnm{Zhan}, \binits{W.}},
\bauthor{\bsnm{Tomizuka}, \binits{M.}}:
\bctitle{Probabilistic prediction of vehicle semantic intention and motion}.
In: \bbtitle{{IEEE} Intelligent Vehicles Symposium (IV)},
pp. \bfpage{307}--\blpage{313}
(\byear{2018})
\end{bchapter}
\endbibitem

\bibitem[\protect\citeauthoryear{Jiang et~al.}{2023}]{jiang2023vad}
\begin{bchapter}
\bauthor{\bsnm{Jiang}, \binits{B.}},
\bauthor{\bsnm{Chen}, \binits{S.}},
\bauthor{\bsnm{Xu}, \binits{Q.}},
\bauthor{\bsnm{Liao}, \binits{B.}},
\bauthor{\bsnm{Chen}, \binits{J.}},
\bauthor{\bsnm{Zhou}, \binits{H.}},
\bauthor{\bsnm{Zhang}, \binits{Q.}},
\bauthor{\bsnm{Liu}, \binits{W.}},
\bauthor{\bsnm{Huang}, \binits{C.}},
\bauthor{\bsnm{Wang}, \binits{X.}}:
\bctitle{Vad: Vectorized scene representation for efficient autonomous
  driving}.
In: \bbtitle{Proceedings of the IEEE/CVF International Conference on Computer
  Vision},
pp. \bfpage{8340}--\blpage{8350}
(\byear{2023})
\end{bchapter}
\endbibitem

\bibitem[\protect\citeauthoryear{Jia et~al.}{2024}]{jia2024bench2drive}
\begin{barticle}
\bauthor{\bsnm{Jia}, \binits{X.}},
\bauthor{\bsnm{Yang}, \binits{Z.}},
\bauthor{\bsnm{Li}, \binits{Q.}},
\bauthor{\bsnm{Zhang}, \binits{Z.}},
\bauthor{\bsnm{Yan}, \binits{J.}}:
\batitle{Bench2drive: Towards multi-ability benchmarking of closed-loop
  end-to-end autonomous driving}.
\bjtitle{Advances in Neural Information Processing Systems}
\bvolume{37},
\bfpage{819}--\blpage{844}
(\byear{2024})
\end{barticle}
\endbibitem

\bibitem[\protect\citeauthoryear{Kirchhof
  et~al.}{2024}]{kirchhof2024pretrained}
\begin{botherref}
\oauthor{\bsnm{Kirchhof}, \binits{M.}},
\oauthor{\bsnm{Collier}, \binits{M.}},
\oauthor{\bsnm{Oh}, \binits{S.J.}},
\oauthor{\bsnm{Kasneci}, \binits{E.}}:
Pretrained visual uncertainties.
arXiv preprint arXiv:2402.16569
(2024)
\end{botherref}
\endbibitem

\bibitem[\protect\citeauthoryear{Khurana
  et~al.}{2022}]{khurana2022differentiable}
\begin{bchapter}
\bauthor{\bsnm{Khurana}, \binits{T.}},
\bauthor{\bsnm{Hu}, \binits{P.}},
\bauthor{\bsnm{Dave}, \binits{A.}},
\bauthor{\bsnm{Ziglar}, \binits{J.}},
\bauthor{\bsnm{Held}, \binits{D.}},
\bauthor{\bsnm{Ramanan}, \binits{D.}}:
\bctitle{Differentiable raycasting for self-supervised occupancy forecasting}.
In: \bbtitle{European Conference on Computer Vision},
pp. \bfpage{353}--\blpage{369}
(\byear{2022}).
\bcomment{Springer}
\end{bchapter}
\endbibitem

\bibitem[\protect\citeauthoryear{Kirchhof et~al.}{2023}]{kirchhof2023url}
\begin{barticle}
\bauthor{\bsnm{Kirchhof}, \binits{M.}},
\bauthor{\bsnm{Mucs{\'a}nyi}, \binits{B.}},
\bauthor{\bsnm{Oh}, \binits{S.J.}},
\bauthor{\bsnm{Kasneci}, \binits{D.E.}}:
\batitle{{URL}: A representation learning benchmark for transferable
  uncertainty estimates}.
\bjtitle{Advances in Neural Information Processing Systems}
\bvolume{36},
\bfpage{13956}--\blpage{13980}
(\byear{2023})
\end{barticle}
\endbibitem

\bibitem[\protect\citeauthoryear{Koopman and
  Wagner}{2016}]{koopman2016challenges}
\begin{barticle}
\bauthor{\bsnm{Koopman}, \binits{P.}},
\bauthor{\bsnm{Wagner}, \binits{M.}}:
\batitle{Challenges in autonomous vehicle testing and validation}.
\bjtitle{SAE International Journal of Transportation Safety}
\bvolume{4}(\bissue{1}),
\bfpage{15}--\blpage{24}
(\byear{2016})
\end{barticle}
\endbibitem

\bibitem[\protect\citeauthoryear{Lafage et~al.}{2025}]{lafage2025hierarchical}
\begin{bchapter}
\bauthor{\bsnm{Lafage}, \binits{A.}},
\bauthor{\bsnm{Barbier}, \binits{M.}},
\bauthor{\bsnm{Franchi}, \binits{G.}},
\bauthor{\bsnm{Filliat}, \binits{D.}}:
\bctitle{Hierarchical light transformer ensembles for multimodal trajectory
  forecasting}.
In: \bbtitle{Proceedings of the IEEE/CVF Winter Conference on Applications of
  Computer Vision (WACV)},
pp. \bfpage{1682}--\blpage{1691}
(\byear{2025})
\end{bchapter}
\endbibitem

\bibitem[\protect\citeauthoryear{Liu et~al.}{2026}]{Liu_2026_CVPR}
\begin{bchapter}
\bauthor{\bsnm{Liu}, \binits{M.}},
\bauthor{\bsnm{Dong}, \binits{H.}},
\bauthor{\bsnm{Fink}, \binits{O.}},
\bauthor{\bsnm{Trapp}, \binits{M.}}:
\bctitle{Adaptive confidence regularization for multimodal failure detection}.
In: \bbtitle{Proceedings of the IEEE/CVF Conference on Computer Vision and
  Pattern Recognition (CVPR)},
pp. \bfpage{15850}--\blpage{15859}
(\byear{2026})
\end{bchapter}
\endbibitem

\bibitem[\protect\citeauthoryear{Lin et~al.}{2017}]{lin2017focal}
\begin{bchapter}
\bauthor{\bsnm{Lin}, \binits{T.-Y.}},
\bauthor{\bsnm{Goyal}, \binits{P.}},
\bauthor{\bsnm{Girshick}, \binits{R.}},
\bauthor{\bsnm{He}, \binits{K.}},
\bauthor{\bsnm{Doll{\'a}r}, \binits{P.}}:
\bctitle{Focal loss for dense object detection}.
In: \bbtitle{Proceedings of the IEEE International Conference on Computer
  Vision},
pp. \bfpage{2980}--\blpage{2988}
(\byear{2017})
\end{bchapter}
\endbibitem

\bibitem[\protect\citeauthoryear{Lambert et~al.}{2009}]{lambert_mccollision}
\begin{bchapter}
\bauthor{\bsnm{Lambert}, \binits{A.}},
\bauthor{\bsnm{Gruyer}, \binits{D.}},
\bauthor{\bsnm{Saint~Pierre}, \binits{G.}}:
\bctitle{A fast monte carlo algorithm for collision probability estimation},
pp. \bfpage{406}--\blpage{411}
(\byear{2009}).
\doiurl{10.1109/ICARCV.2008.4795553}
\end{bchapter}
\endbibitem

\bibitem[\protect\citeauthoryear{Lakshminarayanan
  et~al.}{2017}]{lakshminarayanan2017simple}
\begin{botherref}
\oauthor{\bsnm{Lakshminarayanan}, \binits{B.}},
\oauthor{\bsnm{Pritzel}, \binits{A.}},
\oauthor{\bsnm{Blundell}, \binits{C.}}:
Simple and scalable predictive uncertainty estimation using deep ensembles.
Advances in neural information processing systems
\textbf{30}
(2017)
\end{botherref}
\endbibitem

\bibitem[\protect\citeauthoryear{Ljungbergh
  et~al.}{2024}]{ljungbergh2024neuroncap}
\begin{bchapter}
\bauthor{\bsnm{Ljungbergh}, \binits{W.}},
\bauthor{\bsnm{Tonderski}, \binits{A.}},
\bauthor{\bsnm{Johnander}, \binits{J.}},
\bauthor{\bsnm{Caesar}, \binits{H.}},
\bauthor{\bsnm{{\AA}str{\"o}m}, \binits{K.}},
\bauthor{\bsnm{Felsberg}, \binits{M.}},
\bauthor{\bsnm{Petersson}, \binits{C.}}:
\bctitle{{NeuroNCAP}: Photorealistic closed-loop safety testing for autonomous
  driving}.
In: \bbtitle{European Conference on Computer Vision},
pp. \bfpage{161}--\blpage{177}
(\byear{2024}).
\bcomment{Springer}
\end{bchapter}
\endbibitem

\bibitem[\protect\citeauthoryear{Li
  et~al.}{2022}]{li2022bevformerlearningbirdseyeviewrepresentation}
\begin{botherref}
\oauthor{\bsnm{Li}, \binits{Z.}},
\oauthor{\bsnm{Wang}, \binits{W.}},
\oauthor{\bsnm{Li}, \binits{H.}},
\oauthor{\bsnm{Xie}, \binits{E.}},
\oauthor{\bsnm{Sima}, \binits{C.}},
\oauthor{\bsnm{Lu}, \binits{T.}},
\oauthor{\bsnm{Yu}, \binits{Q.}},
\oauthor{\bsnm{Dai}, \binits{J.}}:
BEVFormer: Learning Bird's-Eye-View Representation from Multi-Camera Images via
  Spatiotemporal Transformers
(2022).
\url{https://arxiv.org/abs/2203.17270}
\end{botherref}
\endbibitem

\bibitem[\protect\citeauthoryear{Malinin et~al.}{2021}]{malinin_shifts_2021}
\begin{botherref}
\oauthor{\bsnm{Malinin}, \binits{A.}},
\oauthor{\bsnm{Band}, \binits{N.}},
\oauthor{\bsnm{{Ganshin}}},
\oauthor{\bsnm{{Alexander}}},
\oauthor{\bsnm{Chesnokov}, \binits{G.}},
\oauthor{\bsnm{Gal}, \binits{Y.}},
\oauthor{\bsnm{Gales}, \binits{M.J.F.}},
\oauthor{\bsnm{Noskov}, \binits{A.}},
\oauthor{\bsnm{Ploskonosov}, \binits{A.}},
\oauthor{\bsnm{Prokhorenkova}, \binits{L.}},
\oauthor{\bsnm{Provilkov}, \binits{I.}},
\oauthor{\bsnm{Raina}, \binits{V.}},
\oauthor{\bsnm{Raina}, \binits{V.}},
\oauthor{\bsnm{{Roginskiy}}},
\oauthor{\bsnm{{Denis}}},
\oauthor{\bsnm{Shmatova}, \binits{M.}},
\oauthor{\bsnm{Tigas}, \binits{P.}},
\oauthor{\bsnm{Yangel}, \binits{B.}}:
Shifts: A Dataset of Real Distributional Shift Across Multiple Large-Scale
  Tasks.
{arXiv}.
Version Number: 3
(2021).
\doiurl{10.48550/ARXIV.2107.07455} .
\url{https://arxiv.org/abs/2107.07455}
Accessed 2025-07-04
\end{botherref}
\endbibitem

\bibitem[\protect\citeauthoryear{Meinhardt
  et~al.}{2022}]{meinhardt2022trackformer}
\begin{bchapter}
\bauthor{\bsnm{Meinhardt}, \binits{T.}},
\bauthor{\bsnm{Kirillov}, \binits{A.}},
\bauthor{\bsnm{Leal-Taixe}, \binits{L.}},
\bauthor{\bsnm{Feichtenhofer}, \binits{C.}}:
\bctitle{Trackformer: Multi-object tracking with transformers}.
In: \bbtitle{Proceedings of the IEEE/CVF Conference on Computer Vision and
  Pattern Recognition},
pp. \bfpage{8844}--\blpage{8854}
(\byear{2022})
\end{bchapter}
\endbibitem

\bibitem[\protect\citeauthoryear{Neumeier et~al.}{2024}]{inproceedings}
\begin{bchapter}
\bauthor{\bsnm{Neumeier}, \binits{M.}},
\bauthor{\bsnm{Dorn}, \binits{S.}},
\bauthor{\bsnm{Botsch}, \binits{M.}},
\bauthor{\bsnm{Utschick}, \binits{W.}}:
\bctitle{Reliable trajectory prediction and uncertainty quantification with
  conditioned diffusion models},
pp. \bfpage{3461}--\blpage{3470}
(\byear{2024}).
\doiurl{10.1109/CVPRW63382.2024.00350}
\end{bchapter}
\endbibitem

\bibitem[\protect\citeauthoryear{Nordlund and
  Gustafsson}{2008}]{nordlund2008conflict}
\begin{botherref}
\oauthor{\bsnm{Nordlund}, \binits{P.-J.}},
\oauthor{\bsnm{Gustafsson}, \binits{F.}}:
Probabilistic Conflict Detection for Piecewise Straight Paths.
Submitted to Automatica
(2008)
\end{botherref}
\endbibitem

\bibitem[\protect\citeauthoryear{Prakash et~al.}{2021}]{prakash2021multi}
\begin{bchapter}
\bauthor{\bsnm{Prakash}, \binits{A.}},
\bauthor{\bsnm{Chitta}, \binits{K.}},
\bauthor{\bsnm{Geiger}, \binits{A.}}:
\bctitle{Multi-modal fusion transformer for end-to-end autonomous driving}.
In: \bbtitle{Proceedings of the IEEE/CVF Conference on Computer Vision and
  Pattern Recognition},
pp. \bfpage{7077}--\blpage{7087}
(\byear{2021})
\end{bchapter}
\endbibitem

\bibitem[\protect\citeauthoryear{Philipp and
  Goehring}{2019}]{philipp2019analytic}
\begin{bchapter}
\bauthor{\bsnm{Philipp}, \binits{A.}},
\bauthor{\bsnm{Goehring}, \binits{D.}}:
\bctitle{Analytic collision risk calculation for autonomous vehicle
  navigation}.
In: \bbtitle{2019 International Conference on Robotics and Automation (ICRA)},
pp. \bfpage{1744}--\blpage{1750}
(\byear{2019}).
\bcomment{IEEE}
\end{bchapter}
\endbibitem

\bibitem[\protect\citeauthoryear{Pek et~al.}{2020}]{pek2020using}
\begin{barticle}
\bauthor{\bsnm{Pek}, \binits{C.}},
\bauthor{\bsnm{Manzinger}, \binits{S.}},
\bauthor{\bsnm{Koschi}, \binits{M.}},
\bauthor{\bsnm{Althoff}, \binits{M.}}:
\batitle{Using online verification to prevent autonomous vehicles from causing
  accidents}.
\bjtitle{Nature Machine Intelligence}
\bvolume{2}(\bissue{9}),
\bfpage{518}--\blpage{528}
(\byear{2020})
\doiurl{10.1038/s42256-020-0225-y}
\end{barticle}
\endbibitem

\bibitem[\protect\citeauthoryear{Sun et~al.}{2025}]{sun2025sparsedrive}
\begin{bchapter}
\bauthor{\bsnm{Sun}, \binits{W.}},
\bauthor{\bsnm{Lin}, \binits{X.}},
\bauthor{\bsnm{Shi}, \binits{Y.}},
\bauthor{\bsnm{Zhang}, \binits{C.}},
\bauthor{\bsnm{Wu}, \binits{H.}},
\bauthor{\bsnm{Zheng}, \binits{S.}}:
\bctitle{Sparsedrive: End-to-end autonomous driving via sparse scene
  representation}.
In: \bbtitle{2025 IEEE International Conference on Robotics and Automation
  (ICRA)},
pp. \bfpage{8795}--\blpage{8801}
(\byear{2025}).
\bcomment{IEEE}
\end{bchapter}
\endbibitem

\bibitem[\protect\citeauthoryear{Toit and Burdick}{2011}]{toit_probabilistic}
\begin{barticle}
\bauthor{\bsnm{Toit}, \binits{N.}},
\bauthor{\bsnm{Burdick}, \binits{J.}}:
\batitle{Probabilistic collision checking with chance constraints}.
\bjtitle{Robotics, IEEE Transactions on}
\bvolume{27},
\bfpage{809}--\blpage{815}
(\byear{2011})
\doiurl{10.1109/TRO.2011.2116190}
\end{barticle}
\endbibitem

\bibitem[\protect\citeauthoryear{Thulasidasan
  et~al.}{2019}]{thulasidasan2019mixup}
\begin{botherref}
\oauthor{\bsnm{Thulasidasan}, \binits{S.}},
\oauthor{\bsnm{Chennupati}, \binits{G.}},
\oauthor{\bsnm{Bilmes}, \binits{J.A.}},
\oauthor{\bsnm{Bhattacharya}, \binits{T.}},
\oauthor{\bsnm{Michalak}, \binits{S.}}:
On mixup training: Improved calibration and predictive uncertainty for deep
  neural networks.
Advances in neural information processing systems
\textbf{32}
(2019)
\end{botherref}
\endbibitem

\bibitem[\protect\citeauthoryear{Tonderski et~al.}{2024}]{tonderski2024neurad}
\begin{bchapter}
\bauthor{\bsnm{Tonderski}, \binits{A.}},
\bauthor{\bsnm{Lindstr{\"o}m}, \binits{C.}},
\bauthor{\bsnm{Hess}, \binits{G.}},
\bauthor{\bsnm{Ljungbergh}, \binits{W.}},
\bauthor{\bsnm{Svensson}, \binits{L.}},
\bauthor{\bsnm{Petersson}, \binits{C.}}:
\bctitle{Neurad: Neural rendering for autonomous driving}.
In: \bbtitle{Proceedings of the IEEE/CVF Conference on Computer Vision and
  Pattern Recognition},
pp. \bfpage{14895}--\blpage{14904}
(\byear{2024})
\end{bchapter}
\endbibitem

\bibitem[\protect\citeauthoryear{Vaswani et~al.}{2017}]{vaswani17}
\begin{bchapter}
\bauthor{\bsnm{Vaswani}, \binits{A.}},
\bauthor{\bsnm{Shazeer}, \binits{N.}},
\bauthor{\bsnm{Parmar}, \binits{N.}},
\bauthor{\bsnm{Uszkoreit}, \binits{J.}},
\bauthor{\bsnm{Jones}, \binits{L.}},
\bauthor{\bsnm{Gomez}, \binits{A.N.}},
\bauthor{\bsnm{Kaiser}},
\bauthor{\bsnm{Polosukhin}, \binits{I.}}:
\bctitle{Attention is all you need}.
In: \bbtitle{{NIPS'17}: Proceedings of the 31st International Conference on
  Neural Information Processing Systems},
pp. \bfpage{6000}--\blpage{6010}
(\byear{2017})
\end{bchapter}
\endbibitem

\bibitem[\protect\citeauthoryear{Wu et~al.}{2023}]{wu2023policy}
\begin{bchapter}
\bauthor{\bsnm{Wu}, \binits{P.}},
\bauthor{\bsnm{Chen}, \binits{L.}},
\bauthor{\bsnm{Li}, \binits{H.}},
\bauthor{\bsnm{Jia}, \binits{X.}},
\bauthor{\bsnm{Yan}, \binits{J.}},
\bauthor{\bsnm{Qiao}, \binits{Y.}}:
\bctitle{Policy pre-training for autonomous driving via self-supervised
  geometric modeling}.
In: \bbtitle{11th International Conference on Learning Representations, ICLR
  2023}
(\byear{2023})
\end{bchapter}
\endbibitem

\bibitem[\protect\citeauthoryear{Westny et~al.}{2021}]{westny2021vehicle}
\begin{bchapter}
\bauthor{\bsnm{Westny}, \binits{T.}},
\bauthor{\bsnm{Frisk}, \binits{E.}},
\bauthor{\bsnm{Olofsson}, \binits{B.}}:
\bctitle{Vehicle behavior prediction and generalization using imbalanced
  learning techniques}.
In: \bbtitle{2021 IEEE International Intelligent Transportation Systems
  Conference (ITSC)},
pp. \bfpage{2003}--\blpage{2010}
(\byear{2021}).
\bcomment{IEEE}
\end{bchapter}
\endbibitem

\bibitem[\protect\citeauthoryear{Wiederer et~al.}{2023}]{wiederer2023joint}
\begin{bchapter}
\bauthor{\bsnm{Wiederer}, \binits{J.}},
\bauthor{\bsnm{Schmidt}, \binits{J.}},
\bauthor{\bsnm{Kressel}, \binits{U.}},
\bauthor{\bsnm{Dietmayer}, \binits{K.}},
\bauthor{\bsnm{Belagiannis}, \binits{V.}}:
\bctitle{Joint out-of-distribution detection and uncertainty estimation for
  trajectory prediction}.
In: \bbtitle{2023 IEEE/RSJ International Conference on Intelligent Robots and
  Systems (IROS)},
pp. \bfpage{5487}--\blpage{5494}
(\byear{2023}).
\bcomment{IEEE}
\end{bchapter}
\endbibitem

\bibitem[\protect\citeauthoryear{Wen et~al.}{2020}]{wenbatchensemble}
\begin{bchapter}
\bauthor{\bsnm{Wen}, \binits{Y.}},
\bauthor{\bsnm{Tran}, \binits{D.}},
\bauthor{\bsnm{Ba}, \binits{J.}}:
\bctitle{Batchensemble: an alternative approach to efficient ensemble and
  lifelong learning}.
In: \bbtitle{International Conference on Learning Representations}
(\byear{2020})
\end{bchapter}
\endbibitem

\bibitem[\protect\citeauthoryear{Wang et~al.}{2025}]{wangredoubt}
\begin{bchapter}
\bauthor{\bsnm{Wang}, \binits{S.}},
\bauthor{\bsnm{Zhou}, \binits{Q.}},
\bauthor{\bsnm{Wu}, \binits{K.}},
\bauthor{\bsnm{Wu}, \binits{D.}},
\bauthor{\bsnm{Lee}, \binits{W.-B.}},
\bauthor{\bsnm{Wang}, \binits{J.}}:
\bctitle{Redoubt: Duo safety validation for autonomous vehicle motion
  planning}.
In: \bbtitle{The Thirty-ninth Annual Conference on Neural Information
  Processing Systems}
(\byear{2025})
\end{bchapter}
\endbibitem

\bibitem[\protect\citeauthoryear{Yu et~al.}{2024}]{yu2024discretization}
\begin{bchapter}
\bauthor{\bsnm{Yu}, \binits{X.}},
\bauthor{\bsnm{Franchi}, \binits{G.}},
\bauthor{\bsnm{Gu}, \binits{J.}},
\bauthor{\bsnm{Aldea}, \binits{E.}}:
\bctitle{Discretization-induced dirichlet posterior for robust uncertainty
  quantification on regression}.
In: \bbtitle{Proceedings of the AAAI Conference on Artificial Intelligence},
vol. \bseriesno{38},
pp. \bfpage{6835}--\blpage{6843}
(\byear{2024})
\end{bchapter}
\endbibitem

\bibitem[\protect\citeauthoryear{Yoo and Kweon}{2019}]{yoo2019learning}
\begin{bchapter}
\bauthor{\bsnm{Yoo}, \binits{D.}},
\bauthor{\bsnm{Kweon}, \binits{I.S.}}:
\bctitle{Learning loss for active learning}.
In: \bbtitle{Proceedings of the IEEE/CVF Conference on Computer Vision and
  Pattern Recognition},
pp. \bfpage{93}--\blpage{102}
(\byear{2019})
\end{bchapter}
\endbibitem

\bibitem[\protect\citeauthoryear{Youden}{1950}]{youden1950index}
\begin{barticle}
\bauthor{\bsnm{Youden}, \binits{W.J.}}:
\batitle{Index for rating diagnostic tests}.
\bjtitle{Cancer}
\bvolume{3}(\bissue{1}),
\bfpage{32}--\blpage{35}
(\byear{1950})
\doiurl{10.1002/1097-0142(1950)3:1<32::AID-CNCR2820030106>3.0.CO;2-3}
\end{barticle}
\endbibitem

\bibitem[\protect\citeauthoryear{Yang et~al.}{2024}]{10598864}
\begin{bchapter}
\bauthor{\bsnm{Yang}, \binits{X.}},
\bauthor{\bsnm{Zhang}, \binits{G.}},
\bauthor{\bsnm{Gao}, \binits{F.}},
\bauthor{\bsnm{Huang}, \binits{H.}}:
\bctitle{Risk-aware defensive motion planning for distributed connected
  autonomous vehicles}.
In: \bbtitle{2024 IEEE Transportation Electrification Conference and Expo
  (ITEC)},
pp. \bfpage{1}--\blpage{6}
(\byear{2024}).
\doiurl{10.1109/ITEC60657.2024.10598864}
\end{bchapter}
\endbibitem

\bibitem[\protect\citeauthoryear{Zhang et~al.}{2017}]{zhang2017mixup}
\begin{botherref}
\oauthor{\bsnm{Zhang}, \binits{H.}},
\oauthor{\bsnm{Cisse}, \binits{M.}},
\oauthor{\bsnm{Dauphin}, \binits{Y.N.}},
\oauthor{\bsnm{Lopez-Paz}, \binits{D.}}:
mixup: Beyond empirical risk minimization.
arXiv preprint arXiv:1710.09412
(2017)
\end{botherref}
\endbibitem

\bibitem[\protect\citeauthoryear{Zhou et~al.}{2023}]{zhou2023interaction}
\begin{botherref}
\oauthor{\bsnm{Zhou}, \binits{J.}},
\oauthor{\bsnm{Olofsson}, \binits{B.}},
\oauthor{\bsnm{Frisk}, \binits{E.}}:
Interaction-aware motion planning for autonomous vehicles with multi-modal
  obstacle uncertainties using model predictive control.
{IEEE} Transactions on Intelligent Vehicles
\textbf{9}(1)
(2023)
\end{botherref}
\endbibitem

\bibitem[\protect\citeauthoryear{Zheng et~al.}{2024}]{zheng2024genad}
\begin{bchapter}
\bauthor{\bsnm{Zheng}, \binits{W.}},
\bauthor{\bsnm{Song}, \binits{R.}},
\bauthor{\bsnm{Guo}, \binits{X.}},
\bauthor{\bsnm{Zhang}, \binits{C.}},
\bauthor{\bsnm{Chen}, \binits{L.}}:
\bctitle{Genad: Generative end-to-end autonomous driving}.
In: \bbtitle{European Conference on Computer Vision},
pp. \bfpage{87}--\blpage{104}
(\byear{2024}).
\bcomment{Springer}
\end{bchapter}
\endbibitem

\end{thebibliography}

\end{document}